\documentclass{article}

\usepackage[utf8]{inputenc}
\usepackage{amssymb}
\usepackage{amsmath}
\usepackage{comment}
\usepackage{amsthm}
\usepackage{todonotes}
\allowdisplaybreaks[4]
\usepackage{geometry}
\usepackage{multicol}
\usepackage{enumerate}
\usepackage{enumitem}

\usepackage{algorithm}  
\usepackage{algorithmicx}  
\usepackage{algpseudocode}

\floatname{algorithm}{Algorithm}

\usepackage{appendix}
\usepackage{bm}
\usepackage{authblk}

\definecolor{purple}{rgb}{0.8,0.0,0.8}

\usepackage{booktabs}
\usepackage{array}
\usepackage{verbatim}
\usepackage[colorlinks, citecolor=blue]{hyperref}
\usepackage{marvosym}

\numberwithin{equation}{section}

\usepackage{prettyref}
\newcommand{\pref}[1]{\prettyref{#1}}

\usepackage{natbib}

\theoremstyle{plain}
\newtheorem{theorem}{Theorem}[section]
\newtheorem{proposition}[theorem]{Proposition}
\newtheorem{lemma}[theorem]{Lemma}
\newtheorem{corollary}[theorem]{Corollary}
\theoremstyle{definition}
\newtheorem{definition}[theorem]{Definition}
\newtheorem{assumption}[theorem]{Assumption}
\theoremstyle{remark}
\newtheorem{remark}[theorem]{Remark}

\begin{document}

\title{Off-Policy Fitted Q-Evaluation with Differentiable Function Approximators: Z-Estimation and Inference Theory}
\author[1]{Ruiqi Zhang}
\author[2]{Xuezhou Zhang}
\author[2]{Chengzhuo Ni}
\author[2]{Mengdi Wang}
\affil[1]{School of Mathematical Science, Peking University \authorcr Email: 1800010777@pku.edu.cn}
\affil[2]{Department of Electrical and Computer Engineering, Princeton University \authorcr Email: \{xz7392, chengzhuo.ni, mengdiw\}@princeton.edu}
\date{}
\maketitle

\begin{abstract}
Off-Policy Evaluation (OPE) serves as one of the cornerstones in Reinforcement Learning (RL). Fitted Q Evaluation (FQE) with various function approximators, especially deep neural networks, has gained practical success. While statistical analysis has proved FQE to be minimax-optimal with tabular, linear and several nonparametric function families, its practical performance with more general function approximator is less theoretically understood. We focus on FQE with general \textit{differentiable function approximators}, making our theory applicable to neural function approximations. We approach this problem using the Z-estimation theory and establish the following results: The FQE estimation error is asymptotically normal with explicit variance determined jointly by the tangent space of the function class at the ground truth, the reward structure, and the distribution shift due to off-policy learning; The finite-sample FQE error bound is dominated by the same variance term, and it can also be bounded by function class-dependent divergence, which measures how the off-policy distribution shift intertwines with the function approximator. In addition, we study bootstrapping FQE estimators for error distribution inference and estimating confidence intervals, accompanied by a Cramer-Rao lower bound that matches our upper bounds. The Z-estimation analysis provides a generalizable theoretical framework for studying off-policy estimation in RL and provides sharp statistical theory for FQE with differentiable function approximators.
\end{abstract}

\section{Introduction}
Off-Policy Evaluation (OPE) studies the problem where one aims to estimate the expected cumulative rewards of a target policy in a Markov decision process, by using only off-policy data, e.g. data generated from some possibly unknown behavior policy different from the one to be evaluated. OPE plays a central role in sample-efficient reinforcement learning, both in online RL where it helps off-policy algorithms achieve superior sample efficiency through data reuse, and in offline RL such as medical applications where online experimentation becomes prohibited. 

Among various approaches to OPE (more discussed in \pref{sec:related_work}), we focus on a specific family of algorithms known as Fitted Q-Evaluation (FQE), that performs iterative regressions to estimate the Q function of the target policy \citep{munos2008finite,le2019batch}. Q or value function approximation is widely used in RL practice in conjunction with neural networks, and has been a key component in many empirically successful RL algorithms, such as the Deep-Q-Network (DQN) and its variants \citep{mnih2013playing}, and Actor-Critic and its variants \citep{mnih2016asynchronous}.
For neural networks, their differentiability is arguably a most important feature that contributes to its generalizability, computation efficiency, and versatility. While most statistical analyses for OPE or FQE focus on specific tabular, parametric (linear, finite) or non-parametric (kernel, Hölder) function classes, we wish to establish theories that are flexible enough to capture most commonly used differential function approximators, including neural networks. 

Statistical theory of OPE is nontrivial due to the distribution shift from off-policy data to the target policy's unknown state-action occupancy measures. When a function approximator is employed, the effect of distribution shift becomes more subtle. 

 In this paper we adopt a substantially more powerful theoretical tool, i.e. the Z-estimation theory, to analyze statistical properties of FQE with general differentiable function class. Roughly speaking, we can view the FQE estimator as the root to a system of equations (i.e. Karush–Kuhn–Tucker conditions for least-square regression), whose parameters come from an empirical process. We study the statistical properties of such root and for more information about the background of Z-estimation theory please see \pref{sec:z} and subsequently in Appendix \ref{sec:prep}.
 
 
 Contributions of this paper are three folds:
\vspace{-5pt}\looseness=-1
\begin{enumerate}[leftmargin=*,itemsep=0pt]
    \item We show that the FQE estimation error is asymptotically normal and give its variance $\sigma^2$ in a closed form. The variance $\sigma^2$ is determined jointly by the tangent plane of the function class $\mathcal{F}$ at the ground truth, the reward and the distribution shift.
    \item We established an $O(\sigma K^{-1/2})$ high-probability finite-sample error bound for FQE. This $\sigma$ can be further relaxed to a restricted $\chi^2_{\mathcal{F}}$-divergence between the behavior policy's and the target policy's occupancy measures. For any two probability measure $p_1$ and $p_2$, we define 
    \begin{equation}\label{definition_chi_square}
        \chi^2_{\mathcal{F}} (p_1,p_2) := \sup_{g \in \mathcal{F}} \frac{\left[\mathbb{E}_{p_1} g(x)\right]^2}{\mathbb{E}_{p_2}\left[g^2(x)\right]} - 1.
    \end{equation}
    It is a function class-dependent divergence between two distributions, and it measures the {\it partial distribution shift with respect to $\mathcal{F}$}.
    \item For statistical inference, we consider bootstrapping the FQE estimator and show that the bootstrap estimators are distributionally consistent. Lastly we provide a Cramer-Rao lower bound that matches the error upper bounds, showing that differentiable FQE is statistical-optimal. 
\end{enumerate}
These results generalize prior minimax-optimal OPE theory for tabular and linear MDP. They apply to a substantially broader family of common function approximators. See \pref{sec:special_cases} for a detailed discussion.

\section{Related Work}\label{sec:related_work}

\paragraph{Off Policy Evaluation} Off policy evaluation(OPE) is a fundamental problem in batch RL. One classic family of OPE methods estimates the value function in an iterative fashion, including Fitted Q Evaluation(FQE) \citep{munos2008finite,le2019batch} and its variant Lasso FQE \citep{hao2021sparse}. Another family of methods uses importance sampling to address the distribution mismatch and get unbiased estimation for policy value\citep{precup2000eligibility}. Vanilla Importance sampling is known to suffer from large variance that in the worst case can scale exponentially with horizon length\citep{yin2020asymptotically,jiang2016doubly}. This drawback was improved by several variance reduction techniques, including doubly robust estimation \citep{jiang2016doubly} and marginal importance ratio estimation \citep{xie2019towards}. Numerous empirical studies have shown their effectiveness for variance reduction and compare their strength and weakness\citep{thomas2016data,li2015toward}. For tabular MDP with full data coverage, \citep{yin2020asymptotically} showed that marginal importance sampling(MIS) estimator is asymptotic optimal with mean square error bound matching the Cramer Rao lower bound in \citep{jiang2016doubly}.

\paragraph{OPE with Linear Function Approximation}   A variety of literature focused on linear function approximator under Bellman completeness, i.e, the Bellman operator maps to state-action value functions that are linear combination of given features \citep{bootstrap,duan2020minimax,wang2019optimism}. \citep{duan2020minimax} provided a minimax lower bound for FQE with linear function approximation, and showed that it matches the upper bound. They also revealed that the dominating term of this minimax lower bound mainly depends on a $\chi^2$-divergence term, which measures the distribution mismatch, {\it in the feature space}, between the data distribution and the occupancy measure of the policy to be evaluated. \citep{bootstrap} further combines bootstrapping with linear FQE and show that the bootstrap estimator is asymptotically efficient. \citep{min2021variance} further provided a tighter upper bound with smaller instance-dependent constants.

\paragraph{OPE with Nonparametric Approximation}  Many efforts have studied the use of nonparametric or seminonparametric function approximators. \citep{kallus2020double} proposed a Double Reinforcement Learning(DRL) estimator using properly estimated $Q$-functions and marginal density ratios, and proved DRL estimator matches a semiparametric efficiency limit for OPE. \citep{uehara2020minimax} proved 
another two estimators based on MIS achieve this efficiency limit. \citep{duan2021optimal} studied on-policy evaluation, and provided non-asymptotic bound on estimation error of kernel Least Square Temporal Difference(LSTD) estimator, and proved that $L^2(\xi^\pi)$-norm of estimation error is $O\left(N^{-\frac{\alpha}{2\alpha + 1}}\right),$ where $\xi^\pi$ is the stationary distribution of underlying transition kernel and $\alpha$ is the decay rate of eigenvalues of kernel functions. Moreover, \citep{shi2021statistical,uehara2021finite} proposed their estimators for $Q$-functions derived the $L^2$-norm convergence rate. \citep{chen2022wellposedness} extended this result in a more general setting under weaker condition. They expressed $Q$-function estimation as non-parametric instrumental variables estimation problem, and proved that 2SLS estimator achieves the minimax optimal convergence rate in both sup and $L^2$-norm, which matches optimal rates in non-parametric regression. 

\paragraph{Other Works with Function Approximation} Function approximation has many applications in different branches of RL, such as Approximate Policy Iteration(API), Fitted Q Iteration(FQI) or Policy Optimization. \citep{cai2020provably,jin2019provably,zhou2021provably} focused on provable policy optimization with linear function approximation.  \citep{munos2008finite} studied Fitted Value Iteration and provided an error bound dependent on the metric entropy. \citep{farahmand2016regularized} focused on both policy evaluation and policy optimization, and provided error rate approximated a rather general function class using metric entropy term. For FQI problem, \citep{chen2019information} provided the sample complexity approximated by general but finite function class, while \citep{le2019batch} proved another upper bound with finite pseudo-dimension function class and cncentrability condition. Recently, FQI with more general function class have been studied. \citep{long2020neuralnetwork} focused on kernel methods based on Reproducing kernel Hilbert spaces and two-layer neural networks based on Barron Space. \citep{fan2020theoretical} considered deep ReLU networks and provided error rate. \citep{nguyentang2021sample} improved this rate under offline setting using smoothness measure in Besov Space. 

\begin{table*}[htb!]
    \centering
    \begin{tabular}{m{2cm}<{\centering} m{2cm}<{\centering}  m{2cm}<{\centering} m{2cm}<{\centering} m{2.5cm}<{\centering} m{5cm}<{\centering}}
        \toprule
        Work & Method & Parametric? &Function Class &Key Assumption & Result\\
        \hline
        \citep{yin2020asymptotically}  & MIS & Yes & Tabular & Concentrability & $\left|v_{\pi} - \widehat{v}_{\pi}\right| \leq \frac{C_{\mu,\bar{\mu}}}{\sqrt{N}} + o(\frac{1}{\sqrt{N}}),$ Meet Cramer-Rao Lower Bound and Locally Minimax\\
        \hline
        \citep{uehara2020minimax}  & MWL / MQL & No & General & Concentrability and Stronger Realizability & $\left|\widehat{v}_{\pi} - v_{\pi}\right| \leq \frac{C}{\sqrt{N}} + \varepsilon_{\text{approx}},$ Semi-parametric Asymptotic Lower Bound \\
        \hline
        \citep{duan2020minimax}  & FQE & Yes & Linear & Completeness & $\left|\widehat{v}^{\pi}-v^{\pi}\right| \leq \frac{CH^2}{\sqrt{N}} \left[\sum_{h=1}^H \sqrt{1 + \chi^2_{\mathcal{Q}}(\mu_h,\bar{\mu})}\right] + O(\frac{1}{N}),$ Minimax Optimal\\
        \hline
        \citep{bootstrap}  & FQE & Yes & Linear & Completeness & Asymptotic Normality, Cramer-Rao Lower Bound and Distributional Consistency\\
        \hline
        \citep{kallus2020double}  & DRL & No & General & Concentrability and Proper rate of Nuisance Estimator & $\left|\widehat{v}_{\pi} - v_{\pi}\right| \leq \sqrt{\frac{2 \log (14 / \delta) \operatorname{Effbd}\left(\mathcal{M}_{2}\right)}{K}} + O(\frac{1}{K}),$ Semi-Efficiency\\
        \hline
        \citep{duan2021optimal}  & LSTD & No & Kernel & Uniformly Bounded Kernel Functions / Eigenfunctions & $\left\|\widehat{v}_{\pi}-v_{\pi}\right\|_{\xi^{\pi}}^{2} \leq c_{1} R^{2}\left\{\delta^{2}+\frac{\lambda_{n}}{1-\gamma}\right\},$ Minimax Optimal on Sample Size and Effective Horizon\\
        \hline
        \textbf{Our Work}  & FQE & Yes & General and Differentiable & Completeness & $\left|\widehat{v}^{\pi}-v^{\pi}\right| \leq \frac{CH^2}{\sqrt{N}} \left[\sum_{h=1}^H \sqrt{1 + \chi^2_{\mathcal{G}_h}(\mu,\bar{\mu})}\right] + O(\frac{1}{N}),$ Asymptotic Normality, Cramer-Rao Lower Bound and Distributional Consistency\\
        \bottomrule
    \end{tabular}
    \caption{Comparison on Different Function Approximators of OPE}
	\label{table_function_approximators} 
\end{table*}

\paragraph{Comparison of OPE Theories} 
We briefly compare existing OPE theories to ours. We emphasize that while many existing works focus on specific linear function or intrinsically non-parametric function approximators, our result applies to almost any class of differentiable, compact function approximators. To our knowledge, we establish the most general parametric function class with $O(\frac{1}{\sqrt{K}})$ convergence rate, with precise variance characterization. Further, we do not require the strong concentrability condition which requires uniformly bounded density ratio. 


\citep{yin2020asymptotically} used marginal importance sampling to set up asymptotic efficiency in batch tabular RL. They assume stronger concentrability(Discussion after Assumption 2.2 and Assumption 2.3). These two assumptions imply $\frac{\mu(s, a)}{\bar{\mu}(s, a)} \leq C$ in \citep{chen2019information}, where $\mu$ and $\bar{\mu}$ denote occupancy distribution of state-action pairs generated by target and behavior policy respectively. Their main result (Theorem 3.1) is about $\mathbb{E}_{\bar{\mu}}\left[\widehat{v}_{\pi} - v_{\pi}\right],$ but by standard concentration techniques we can soon get a high probability bound. Roughly speaking, this is $\left|\widehat{v}_{\pi} - v_{\pi}\right| \leq \frac{C_{\mu,\bar{\mu}}}{\sqrt{N}},$ where $C_{\mu,\bar{\mu}}$ is a constant similar to a $\chi^2$-divergence which captures distribution mismatch.

Our work is mostly related to \citep{bootstrap} and \citep{duan2020minimax}. They considered linear function approximation with possibly infinite state-action space. Under policy completeness, \citep{duan2020minimax} showed that $\left|\widehat{v}^{\pi}-v^{\pi}\right| \leq \frac{C}{\sqrt{N}} \left[\sum_{h=1}^H (H-h+1) \sqrt{1 + \chi^2_{\mathcal{Q}}(\mu_h,\bar{\mu})}\right] + O(\frac{1}{N}),$ where $\mu_h$ is marginal distribution of $(s_h,a_h)$ generated by target policy, $\mathcal{Q}$ is linear function class spanned by feature map $\phi(s,a),$ and is also the space where all $Q$-functions lie in. This $\chi^2$-divergence is a special case of ours. \citep{bootstrap} established asymptotic results under the same setting, including asymptotic normality(equivalent to asymptotic upper bound), and asymptotic lower bound. The asymptotic variance meets Remark 3.2 in \citep{yin2020asymptotically} in tabular case. Moreover they consider standard bootstrap, while we consider one of its alternatives to get a more general asymptotic confidence interval.

\citep{uehara2020minimax} assumes stronger realizability not only for $Q$-functions, but for density ratio as well. Although two realizability are not necessary for their finite upper bound, they still need certain assumption stronger than realizability for $Q$-functions only. Their concentrability condition(Assumption 2) is the same as that in \citep{chen2019information}. They proved the sample complexity of MWL / MQL estimators and provided semi-parametric efficiency. Another example of non-parametric OPE method is in \citep{kallus2020double}, which matches the semi-parametric lower bound as well. Assumption 1 in \citep{kallus2020double} assumes a full data coverage and bounded density ratio, which is stronger than concentrability in \citep{chen2019information}. They further assume proper rates of estimators of $Q$-functions and density ratios, but they do not focus on estimators of both nuisance. Under these assumptions, they get an $O(\frac{1}{\sqrt{K}})$ error rate, and the dominating term depends on effective bound of MDP $\operatorname{Effbd}\left(\mathcal{M}_{2}\right)$ which captures the distribution mismatch(See equation (4) in their paper). \citep{duan2021optimal} considered online policy evaluation, hence neither concentratability nor completeness is needed. They consider a uniformly bounded kernel function class, which is intrinsically non-parametric when there are infinite eigenvalues for kernel operator. $R$ in their result captures the structural property of the projected fixed point(In case of no approximation error, this is just policy value). In linear kernel setting, this reduces into linear LSTD estimation and has an error rate of order $O(\frac{1}{\sqrt{N}}).$


\section{Preliminaries}

\paragraph{Markov Decision Process} We consider the finite-horizon time-homogeneous Markov Decision Process denoted by $\mathcal{MDP}\left(\mathcal{S},\mathcal{A},\mathcal{P},\pi,r,\xi,H\right)$, where $\mathcal{S}$ and $\mathcal{A}$ are the state space and the action space and their joint space is denoted by $\mathcal{X} = \mathcal{S} \times \mathcal{A}.$ Denote $r \in \mathbb{R}^{\mathcal{X}}$ as the reward function, where $r(s,a)\in[0,1]$ for all $s\in \mathcal{S}$ and $a \in \mathcal{A}$. The transition probability is determined by $p\left(\cdot \mid s,a\right),$ and policy $\pi(\cdot \mid s)$ gives the probability distribution on $\mathcal{A}$ conditional on current state $s$. Throughout our paper we use $\pi$ to denote the target policy to evaluate. $\xi$ is the initial distribution and $H$ is horizon length. We denote $\boldsymbol{\tau} = (s_1,a_1,s_2,a_2,...,s_H,a_H,s_{H+1})$ as a random trajectory in the data. The state-action value function (Q function) is $Q_h(s,a) := \mathbb{E}^{\pi} \left[\sum_{h^{\prime} = h}^H r(s_{h^{\prime}},a_{h^{\prime}}) \mid s_h = s, a_h = a\right],$ for $h \in [H]$, where $\mathbb{E}^{\pi}$ denotes expectation over random trajectories generated by the target policy $\pi.$ The target value function is $V_h(s) := \mathbb{E}^{\pi} \left[\sum_{h^{\prime} = h}^H r(s_{h^{\prime}},a_{h^{\prime}}) \mid s_h = s\right].$

\paragraph{Off-Policy Policy Evaluation (OPE)} Our goal is to estimate the cumulative reward of target policy $\pi$ from an initial distribution $\xi(\cdot),$ which is a scalar value defined as $v_{\pi} := \mathbb{E}^{\pi} \left[\sum_{h=1}^H r(s_{h},a_{h}) \mid s_1 \sim \xi(\cdot)\right].$ We focus on the off-policy learning problem, where logged experiences were pre-collected according to some (possibly unknown) behavior policy and no more interaction with the MDP environment is allowed. Specifically, suppose we have $K$ episodes of data $\left\{\boldsymbol{\tau}_k\right\}_{k=1}^K,$ which are i.i.d sampled using the behavior policy $\Bar{\pi}.$ We sometimes use $\mathcal{D}=\left\{\left(s_{n}, a_{n}, r_{n}\right)\right\}_{n \in[N]} = \left\{\left(s_{h}^k, a_{h}^k, r_{h}^k\right)\right\}_{h \in[H], k \in [K]}$ to denote our batch samples where $N = KH$ is the total number of state transitions.

We denote $\mu$ as the state-action occupation measure generated by policy $\pi$ from the initial distribution. Also we denote $\Bar{\mu}$ as the population distribution, i.e., the state-action measure generated by behaviour policy $\Bar{\pi}.$ 

\section{Fitted Q Evaluation (FQE) with Differential Function Approximation}
\label{sec:fqe}

While there exist a variety of OPE algorithms, fitted Q evaluation is most popular in practice. Popularity of such value-based methods is related to their compatibility with deep learning, where training and function fitting is most convenient when minimizing squared losses. 
\subsection{FQE Algorithm}
Fitted Q Evaluation exploits iterative regression to approximate Q functions and eventually estimates target policy value by integrating estimator of Q function \citep{le2019batch, fonteneau2013batch}. 

Let $\mathcal{F}$ be a class of function approximators. FQE framework can be summarized as follows: We let $\widehat{Q}_{H+1} = 0$ and for $h = H,H-1,...,1,$ we iteratively solve
\begin{equation}\label{FQE}
    \widehat{Q}_h = \mathop{\arg \min}_{f \in \mathcal{F}} \left\{ \frac{1}{N} \sum_{n=1}^N \left[f(s_n,a_n) - y_n\right]^2 + \lambda \rho(f)\right\}.
\end{equation}
where \begin{small}$y_n = r(s_n,a_n) +  \int_{\mathcal{A}} \widehat{Q}_{h+1}(s_{n+1},a) \pi(a \mid s_{n+1}) d a,$\end{small} $\lambda > 0$ and $\rho(\cdot)$ is a proper regularizer. 
The full algorithm is given below.
\begin{algorithm}[htb!]
\caption{Framework for Fitted Q Evaluation}
\label{alg1}
	\begin{algorithmic}[1] 
		\Require Target policy $\pi$, initial distribution $\xi,$ dataset $\mathcal{D}=\{(s^{k}_h,a^{k}_h,s^{k}_{h+1},r^{k}_h)\}_{h\in[H],k\in[K]}$, function class $\mathcal{F}.$
		\State \textbf{Initialize } $\widehat{Q}_{H+1}(s,a) = 0,\forall (s,a)\in\mathcal{X}.$
		\For{$h=H,H-1,\ldots,1$}  
		\State Solve \eqref{FQE}
		\EndFor
		\State \textbf{Return} $\widehat{v}_\pi = \int_{\mathcal{S} \times \mathcal{A}} \widehat{Q}_1(s,a) \pi(a|s) \xi(s) \mathrm{d} a \mathrm{d} s.$
	\end{algorithmic}
\end{algorithm}

\subsection{FQE with a Differentiable Function Class}
Let there be a feature map $\phi(\cdot): \mathcal{S} \times \mathcal{A} \to \Psi \subset \mathbb{R}^m,$ where $\Psi$ is the space of state-action feature vectors. We consider function approximators that take $\phi(s,a)$ as input and output an estimated value. For simplicity, previous theory often confined $\mathcal{F}$ to some linear function class \citep{bootstrap, duan2020minimax}. Note this contains the tabular case by letting $\phi(s,a)$'s be one-hot features.

We consider {\it almost arbitrarily} parametrized $\mathcal{F}$ with mild smoothness condition, i.e., 
$$
\mathcal{F}:=\left\{f_{\theta}(\phi(\cdot)): \mathcal{S} \times \mathcal{A} \rightarrow \mathbb{R}, f_{\theta}: \mathbb{R}^{m} \rightarrow \mathbb{R}, \theta\in\Theta \right\},
$$
where $\Theta$ is the parameter space.
We make no distinction between $f_{\theta}(\phi)$ and $f(\theta,\phi).$ Without loss of generosity, we assume $\boldsymbol{0} \in \Theta$ and $f(\boldsymbol{0},\phi(s,a)) = 0 $ for any $(s,a) \in \mathcal{S} \times \mathcal{A}.$ We use $\nabla_{\theta} f(\theta,\phi)$ to denote its partial derivative with respect to $\theta,$ and use $\nabla_{\theta }^2 f(\theta,\phi)$ to denote Hessian matrix. By parametrization the recursive minimization in FQE can be turned into least square optimization in $\theta$. We assume the regularization on function $f$ is actually on parameter $\theta \in \Theta,$ hence we make no distinction between $\rho(f)$ and $\rho(\theta).$ We denote $\widehat{Q}_h = f(\widehat{\theta}_h,\phi)$ and we use $\phi_n$ for short of $\phi(s_n,a_n)$, then \eqref{FQE} is equivalent to
\begin{equation}\label{definition_theta_h}
    \widehat{\theta}_h = \mathop{\arg \min}_{\theta \in \Theta} \left\{\frac{1}{2N} \sum_{n=1}^N \left[f(\theta,\phi_n) - y_n(\widehat{\theta}_{h+1})\right]^2 + \lambda \rho(\theta)\right\},
\end{equation}
where \begin{footnotesize} $y_n(\theta^\prime) = r(s_n,a_n) +  \int_{\mathcal{A}} f(\theta^\prime, \phi(s_{n+1},a)) \pi(a \mid s_{n+1}) \mathrm{d} a.$ \end{footnotesize} Then, the final FQE estimator becomes 
$
    \widehat{v}_{\pi} = \int_{\mathcal{S} \times \mathcal{A}} f(\widehat{\theta}_1,\phi(s,a)) \pi(a \mid s) \xi(s) d a d s.
$
We denote by $\theta^*=(\theta^*_1,\ldots,\theta^*_H)$ the ground truth parameter, which are solutions to \eqref{definition_theta_h} when there is no regularizer and the empirical sum is replaced with expectation over the population distribution. 
\section{Assumptions}

In this section, we summarize main assumptions for statistical theory of differentiable FQE.

\begin{assumption}[Compactness]\label{assumption_compact} \it
    $\Theta \subset \mathbb{R}^d$ and $\Psi \subset \mathbb{R}^m$ are compact. We denote $\operatorname{int}(\Theta)$ as its interior. We assume $\widehat{\theta}_h \in \operatorname{int}(\Theta)$ for $h \in [H].$
\end{assumption}

Compactness of state-action space is a natural assumption and was also assumed in \citep{bootstrap,duan2020minimax,hao2021sparse} and \citep{yang2020bridging}. Without loss of generality, we assume that the ground truth belongs to the interior of $\Theta$, i.e.,  $\left\{\theta: \left\|\theta - \theta_h^*\right\|_2 \leq 1\right\} \subset \operatorname{int}(\Theta)$ for all $h$.


\begin{assumption}[Differentiability]\label{assumption_differentiability} \it
    For any $f\in\mathcal{F}$, $f(\theta,\phi)$ is third-time continuously differentiable in $\Theta \times \Psi$ with respect to $\theta$ and $\phi.$ The regularizer function $\rho(\theta)$ is differentiable with bounded gradient in $\Theta.$ 
\end{assumption}

This assumption requires only sufficient smoothness of $f$. It does not require $f$ to take any specific parametric form or belong to certain kernel space.  

\begin{assumption}[Policy Completeness]\label{assumption_completeness}
\it
    For any function $f,$ we define the operator $\mathcal{P}: \mathbb{R}^{\mathcal{X}} \to \mathbb{R}^{\mathcal{X}},$ such that for any $(s,a) \in \mathcal{X},$
    \begin{equation}\label{definition_P}
        \left(\mathcal{P} f\right)(s,a) = \mathbb{E}^{\pi}\left[r(s,a) + \int_{\mathcal{A}} f(s^{\prime},a^{\prime}) \pi(a^{\prime} \mid s^{\prime}) \mathrm{d} a^{\prime} \Bigg\vert s,a\right].
    \end{equation}
We assume $r\in \mathcal{F}$ and for any $f \in \mathcal{F}, $ we have $\mathcal{P}f \in \mathcal{F}.$ 
\end{assumption}


\paragraph{Remark on policy completeness assumption}
This assumption requires sufficient expressiveness of $\mathcal{F}$. It implies the realizability for $Q$ functions, i.e., the groundtruth $\theta^*$ fully recovers the Q functions $f(\theta_h^*,\phi) = Q_h(s,a).$ 
It is a rather crucial assumption for RL with function approximation, and commonly used in OPE literatures \citep{bootstrap, duan2020minimax, le2019batch,hao2021sparse,fan2020theoretical}. \citep{chen2019information} conjectured the realizability alone is not enough for sample-efficient offline RL. Later \citep{wang2020statistical} verified that only assuming realizability without policy completeness can lead to exponential sample complexity, unless one assumes a strong concentrability condition such that the distribution shift be uniformly bounded across state-action space ($\sup_{s,a} \frac{\mu(s,a)}{\bar \mu(s,a)} < C$) \citep{munos2008finite,farahmand2016regularized,le2019batch}. The concentrability condition is rather restrictive and requires that the target and behavior policies be extremely close. However, for a simple linear Gaussian system, even if  $\mu,\bar\mu$ are two close Gaussian distributions,  a small mean difference easily leads to $\sup_{s,a} \frac{\mu(s,a)}{\bar \mu(s,a)} = \infty.$ An exception is the work of \citep{uehara2020minimax}, which proposed a minimax approach for OPE requiring only realizability, albeit on both the Q function and the density function $\mu$ using two function classes. However, minimax optimization is computationally harder to implement than least-square regression. 

Therefore to handle practical off-policy problems and nontrivial distribution shift, we choose to make the policy completeness assumption. 
Note that even if the completeness fails to hold and there exists a nonzero approximation error, our results still apply to bounding the statistical error of $\hat \theta$ and the approximation error can be handled by classic results on approximate value iteration \citep{szepesvari2005finite}. 

In what follows, we will explicate the dependency of error upper bound on the distribution shift and $\mathcal{F}$. We will show that FQE can be efficient without strong concentrability.
 
 

\paragraph{Notation} 
Let $\theta_1,\theta_2,...,\theta_H \in \mathbb{R}^d$ be variables and define $\theta^{\top} = (\theta_1^{\top},\theta_2^{\top},...,\theta_H^{\top}) \in \mathbb{R}^{Hd}$ as the variable stacked by $\theta_h \ h \in [H].$ Denote $\theta^{*\top} = (\theta_1^{*\top},\theta_2^{*\top},...,\theta_H^{*\top}) \in \mathbb{R}^{Hd},$ and denote the FQE estimator as $\widehat{\theta}^{\top} = (\widehat{\theta}_1^{\top}, \widehat{\theta}_2^{\top},...,\widehat{\theta}_H^{\top})$. For any matrix $E\in\mathbb{R}^{d_1\times d_2}$ (including scalars and vectors as special cases), we define $\frac{\partial}{\partial \theta} E_{\theta} = \nabla_\theta E_\theta = (\nabla_\theta^1 E_\theta, \nabla_\theta^2 E_\theta, \ldots, \nabla_\theta^m E_\theta)\in\mathbb{R}^{d_1\times md_2}.$  We denote $\mathbb{E}_{\boldsymbol{\tau}}$ or $\mathbb{E}$ as the expectation over the population distribution generated by behavior policy, and $\mathbb{E}^{\pi}$ as the expectation over target policy. For any matrix $E,$ we denote $\left\|E\right\|$ as its operator norm, i.e. its maximal singular value. We use $\phi_h$ and $r_h$ for short of $\phi(s_h,a_h)$ and $r(s_h,a_h).$

\section{$Z$-Estimation Theory for Differentiable FQE Estimators}\label{sec:z}


In this section, we study statistical properties of the differentiable FQE estimator. The challenge with such general function approximation is the lack of analytical expressions for ground true and estimated parameters. 
We adopt the Z-Estimator theory as a central tool and all the proof will be deferred to Appendix.

\subsection{FQE as a $Z$ Estimator}
First we show that the FQE can be written as a Z-Estimator, which means the estimator takes the form of the root of some systems. From the optimality condition of \eqref{definition_theta_h} for interior solutions, we know 
    $$
    \nabla_{\theta}\left\{\frac{1}{2N} \sum_{n=1}^N \left[f(\widehat{\theta}_h,\phi_n) - y_n(\widehat{\theta}_{h+1})\right]^2 + \lambda \rho(\widehat{\theta})\right\} = 0.
    $$
For any sample path $\boldsymbol{\tau} = (s_1,a_1,...,s_H,a_H)$ and any $\theta = (\theta_1^{\top},\theta_2^{\top},...,\theta_H^{\top})^{\top} \in \Theta^H$ and $\theta_{H+1} = \boldsymbol{0},$ we define the $z_{(h)}$ function as
\begin{equation*}
    z_{(h)}(\theta,\boldsymbol{\tau}) = \sum_{j=1}^H \bigg(f(\theta_h,\phi_j) - r_j - y_j(\theta_{h+1})\bigg) \cdot \nabla_{\theta}^{\top} f(\theta_h,\phi_j),
\end{equation*}
where \begin{footnotesize}$y_j(\theta) := \int_{\mathcal{A}} f(\theta,\phi(s_{j+1},a^{\prime})) \pi(a^{\prime}\mid s_{j+1}) \mathrm{d} a^{\prime}.$ \end{footnotesize} Denote 
$$z(\theta,\boldsymbol{\tau})^{\top} = (z_{(1)}^{\top}(\theta,\boldsymbol{\tau}),  z_{(2)}^{\top}(\theta,\boldsymbol{\tau}), ..., z_{(H)}^{\top}(\theta,\boldsymbol{\tau})) \in \mathbb{R}^{Hd}$$
and its expectation $Z(\theta) := \mathbb{E}_{\boldsymbol{\tau}} \left\{z(\theta,\boldsymbol{\tau})\right\}.$ The groundtruth $\theta^*$ is the root of the expected $Z$ function, i.e., 
\begin{small}
\begin{equation}
    Z(\theta^*) = 0.
\end{equation}
\end{small}
Define the empirical $Z$ function as $Z_K\left(\theta\right) := \frac{1}{K} \sum_{k=1}^K z(\theta,\boldsymbol{\tau}_k)$ and denote \begin{small} $R(\theta) := (\nabla_{\theta} \rho(\theta_1), \nabla_{\theta} \rho(\theta_2), \ldots, \nabla_{\theta} \rho(\theta_H))^{\top}.$\end{small} Then we have
\begin{small}
\begin{equation}\label{Z-Estimator}
    Z_K(\widehat{\theta}) + \lambda R(\widehat{\theta})= \boldsymbol{0}.
\end{equation}
\end{small}
Therefore $\widehat{\theta}$ is the root of above equation and thus is a Z-Estimator. We use $\widehat \theta$ interchangeably with $\widehat{\theta}_K = (\widehat{\theta}_{K,1}^{\top},...,\widehat{\theta}_{K,H}^{\top})^{\top}$ to explicate its dependence on $K$. 

Define the $Z$ function class as
\begin{equation*}
    \mathcal{Z}:=\bigcup_{h=1}^{H} \bigcup_{i=1}^{d}\left\{z_{h, i}(\theta, \cdot): \theta \in \Theta^{H} \subset \mathbb{R}^{H d}\right\},
\end{equation*}
where $z_{h,i}$ denotes the $i$-th entey of $z_{(h)}(\theta,\boldsymbol{\tau}).$ 
The complexity of $\mathcal{Z}$, which comes from the complexity of $\mathcal{F}$, determines the statistical efficiency of FQE. 
We analyze the estimation error via Tarlor expansions, concentration inequalities and bounding the complexity of $\mathcal{Z}$ using its bracketing integral \citep{kosorok}.

\subsection{Asymptotic Normality and Variance}

\begin{theorem}[Asymptotic Normality]\label{thm_normality}
Under assumption \ref{assumption_compact}, \ref{assumption_differentiability} and \ref{assumption_completeness}, if $\theta^*$ is the unique root of $Z(\theta)$ in $\Theta^H$ and the Jacobian matrix of $Z(\theta)$ at $\theta^*$ is non-singular. Then, when $K \to \infty$ and $\lambda = o(K^{-1/2}),$ we have
    $$
    \sqrt{K} \left(\widehat{v}_{\pi} - v_{\pi}\right) \stackrel{d}{\longrightarrow} \mathcal{N}(0,\sigma^2).
    $$
    The asymptotic variance is given by
    \begin{equation}\label{variance_expression}
    \sigma^2 = \frac{1}{H} \sum_{h_1,h_2 = 1}^H \nu_{h_1}^{\top} \Sigma^{-1}_{h_1} \Omega_{h_1,h_2} \Sigma^{-1}_{h_2} \nu_{h_2}.
    \end{equation}
    where for $h,i,j \in [H],$
    \begin{align*}
        \Sigma_{h}&=\mathbb{E}\left[\frac{1}{H} \sum_{j=1}^{H}\left(\nabla_{\theta_h} f\left(\theta_{h}^*, \phi_j\right)\right)^{\top}\left(\nabla_{\theta_h} f\left(\theta_{h}^*, \phi_j\right)\right)\right];\quad & \nu_h^{\top} &= \mathbb{E}^{\pi}\left[\nabla_{\theta_h} f\left(\theta_{h}^*, \phi\left(s_{h}, a_{h}\right)\right) \mid s_{1} \sim \xi(\cdot) \right]; \\
        \Omega_{i,j} &= \mathbb{E} \left[\frac{1}{H} \sum_{h = 1}^H \left(\nabla_{\theta_i}^{\top} f \left(\theta_i^*,\phi_h\right) \right) \left(\nabla_{\theta_j} f \left(\theta_j^*,\phi_h\right) \right) \varepsilon_{i,h} \varepsilon_{j,h}\right]; \quad & \varepsilon_{j,h} &=f\left(\theta_{j}^*, \phi_h\right) - r_h - \mathbb{E}^{\pi} \left[f(\theta_{j+1}^*,\phi_{h+1}) \bigg| s_{h+1}\right].
    \end{align*}
\end{theorem}

\begin{remark}
{}In linear case, this expression for variance is exactly the same as in \citep{bootstrap}. In that case, all $\Sigma_h$ become to the dataset's covariance matrix, $\nu_h$ becomes the state feature expectation $\phi(s_h,a_h)$ under target policy. They become both independent of $\theta_h^*.$ In tabular case with time-inhomogeneous MDP and one-hot feature, all $\Sigma_h$ become diagonal and the asymptotic variance matches the result in Remark 3.2 in \citep{yin2020asymptotically}.
\end{remark}

\paragraph{Proof of Theorem \ref{thm_normality}} The central tool for the proof is Z-Estimator Master Theorem \citep{kosorok}. To use this theorem, we need to 
verify the function class comprising all entries of $z(\xi,\boldsymbol{\tau})$ indexed by $\theta \in \Theta^H$ is both Glivenko-Cantelli and Donsker. Glivenko-Cantelli and Donsker are properties of a function class that measure its complexity. Z-Estimator theory mainly tells us that, with a function class not too complex, asymptotic normality holds. For our differentiable $\mathcal{F}$, we will use a bracket integral argument to bound its complexity. A gentle introduction to these tools can be found in Appendix \pref{sec:prep}. Asymptotic normality implies the following corollary, which implies that the convergence rate of $|\widehat{v}_{\pi} - v_{\pi}|$ is $O(\frac{1}{\sqrt{K}}).$ 
\begin{corollary}\label{coro_prohorov}
    For any $\delta > 0,$ there exists a constant $B(\delta) > 0$ such that 
    $$
    \sup _{K \in \mathbb{N}} \mathbb{P}\left(\left\|\sqrt{K}\left(\widehat{\theta}_{K}-\theta^{*}\right)\right\|_{2}>B(\delta)\right) \leq \delta.
    $$
\end{corollary}

\subsection{Finite Sample Error Upper Bound}

Next we will show finite-sample error bound for FQE. In addition to the $O(\frac{1}{\sqrt{K}})$ rate, we will show that the leading order term in FQE error largely depends on the variance $\sigma^2$. We also provide a reward-free error bound that depends on a function class-dependent divergence, which measures the partial mismatch between $\mu$ and $\bar \mu$ with respect to the $\mathcal{F}$ space. Our results strictly generalize the minmax-optimal error bounds for linear FQE \citep{bootstrap,duan2020minimax,On_PG}. 

\begin{assumption}[Data Coverage]\label{assumption_data_coverage}
    Let $C_2$ be a positive constant. We assume for any $(s,a) \in \mathcal{S} \times \mathcal{A}$ and $h \in [H],$
    $$
    \nabla_{\theta} f(\theta_h^*,\phi(s,a)) \Sigma_h^{-1} \nabla_{\theta}^{\top} f(\theta_h^*,\phi(s,a)) \leq C_2 d.
    $$
\end{assumption}

To interpret this assumption, consider the case where $f$ is linear. In this case, the assumption reduces to $\phi(s,a)^{\top} \Sigma^{-1}\phi(s,a) \leq C_2 d$. It holds as long as the data has a non-singular covariance $\Sigma$. Even if $\sup_{s,a} \frac{\mu(s,a)}{\bar\mu(s,a)}=\infty$, data can still be well-conditioned and cover every dimension of $\nabla_{\theta} f(\theta^*,\phi)$, thus 
satisfying Assumption \ref{assumption_data_coverage}. Such coverage can be achieved with as small as $O(Hd)$ sample transitions, while full data coverage requires $\Omega(SA)$ samples.  It is a much weaker assumption than assuming $\sup_{s,a} \frac{\mu(s,a)}{\bar\mu(s,a)}<C$ or assuming $\min_{s,a}\mu(s,a) > c $.


For $h \in [H],$ we define
\begin{equation}\label{definition_G_h}
    \mathcal{G}_h := \left\{\left(\nabla_{\theta} f\left(\theta_{h}^*, \phi(\cdot)\right)\right)\mu: \mu \in \mathbb{R}^d\right\}.
\end{equation}

We denote $\alpha = (\alpha_1,\alpha_2,...,\alpha_d)$ as a multi-index, and the order of $\alpha$ is $|\alpha|=\alpha_{1}+\alpha_{2}+\cdots+\alpha_{n}.$ We denote $\partial^{\alpha}_{\theta} f(\theta,\phi) = \partial_{\theta_1}^{\alpha_{1}} \partial_{\theta_2}^{\alpha_{2}} \cdots \partial_{\theta_d}^{\alpha_{d}} f=\frac{\partial^{|\alpha|} f}{\partial \theta_{1}^{\alpha_{1}} \partial \theta_{2}^{\alpha_{2}} \cdots \partial \theta_{d}^{\alpha_{d}}},$ and for $l = 1,2,3,$ we define
\begin{equation*}
    \kappa_{l} := \sup_{\theta \in \Theta, \phi \in \Psi} \sup_{\left|\alpha\right| = l} \left| \partial^{\alpha}_{\theta} f(\theta,\phi) \right| < \infty.
\end{equation*}
Our main result on finite sample upper bound is given below.
\begin{theorem}[Finite Sample Upper Bound]\label{thm_finite_sample_square}
Under the assumptions of Theorem \ref{thm_normality} and Assumption \ref{assumption_data_coverage}, let $\lambda = 0$, for every fixed $K \geq \max\left\{\frac{2(C_2 d + 1)^2}{C_2 d}\log\left(\frac{12dH}{\delta}\right),B(\delta/3)^2\right\},$ and every $\delta > 0,$ with probability at least $1 - \delta,$ we have
\begin{itemize}
    \item[(i)] Variance-aware error bound:
    \begin{align}\label{finite_bound2}
    \left|\widehat{v}_{\pi} - v_{\pi} \right| \leq \sqrt{\frac{2 \log (6/ \delta) \sigma^2}{K}} +\frac{1}{K}\left[\frac{2}{3} \ln \left(\frac{6}{\delta}\right) \sqrt{C_{2} d} \sum_{h=1}^{H}(H-h+1) \sqrt{\nu_{h}^{\top} \Sigma_{h}^{-1} \nu_{h}} + C\right],
\end{align}
\item[(ii)] Reward-free error bound:
\begin{align}\label{finite_bound1}
    \left|\widehat{v}_{\pi} - v_{\pi} \right| \leq \left[\sum_{h=1}^{H}(H-h+1) \sqrt{1+\chi_{\mathcal{G}_{h}}^{2}(\mu, \bar{\mu})}\right] \cdot \left[\sqrt{\frac{\log (12/ \delta)}{2 K H}} + \frac{4 \ln(12dH/\delta)}{3K} \sqrt{C_{2} d H}\right] + \frac{C}{K}.
\end{align}
\end{itemize}
where
\begin{align*}
    C = B(\delta / 3)\left[H^{2 \frac{1}{2}} d \kappa_{1}^{2} B_{0}+2 H^{\frac{3}{2}} d \kappa_{2} B_{0}+4 d^{\frac{3}{2}} \kappa_{1} \kappa_{2} B_{0}\right] + B(\delta / 3)^{2}\left[B_{0} H^{3} d\left[\kappa_{3} H+3 \kappa_{1} \kappa_{2}+2 \sqrt{d} \kappa_{1} \kappa_{3}\right]\right]
\end{align*}
is a constant, where $B_{0}:=\max _{h \in[H]} \sqrt{\nu_{h}^{\top} \Sigma_{h}^{-2} \nu_{h}}$, $D$ is a constant dependent on $H$ and $d$ only and $\chi^2_{\mathcal{G}_h}$ is restricted $\chi^2$-divergence defined as \eqref{definition_chi_square}. 
\end{theorem}

\paragraph{Proof} We decompose $\widehat{v}_{\pi} - v_{\pi}$ into a sum of a first order term and higher order terms by using Taylor expansion. We bound the first order term with Freedman's Inequality (lemma \ref{ine_freedman}). The higher order term is much more complex than that of linear case. We use upper bound for supremum norm of empirical process to bound them conditional on the event $\left\{\left\|\widehat{\theta}_K - \theta^*\right\| \leq \frac{B(\delta)}{\sqrt{K}}\right\}$ for any fixed $K.$ Exact characterization of $B(\delta)$ is not possible without more specific assumptions on $\mathcal{F}$, which is an open problem in empirical process theory and beyond our scope.

\paragraph{Remarks}
The error bound (i) is variance-aware and the tightest in the sense that $\sigma^2$ equals to the asymptotic variance given by Theorem \ref{thm_normality}. The error bound (ii) is reward-free in the sense that it does not involve the reward function $r$ at all. It is a worst-case error bound that determined solely by the distribution shift of off-policy learning, measured by $\sqrt{1+\chi_{\mathcal{G}_{h}}^{2}(\mu, \bar{\mu})}$. In this linear function case, $1+\chi_{\mathcal{G}_{h}}^{2}(\mu, \bar{\mu})$ can be bounded by the relative condition number between the two distributions' covariance matrices. Even without function approximation, if $\mu,\bar\mu$ are two Gaussians with the same variance and different means, we have $\sup_{s,a} \frac{\mu(s,a)}{\bar\mu(s,a)}=\infty$, but their chi-square divergence is finite. 



\subsection{Special Cases}\label{sec:special_cases}

\paragraph{Positivity Condition} First we derive a sharper finite sample upper bound under stronger assumption for data coverage. 
\begin{assumption}[Positivity]\label{assumption_positivity}
    We further assume for any $(s,a),(s^{\prime},a^{\prime}) \in \mathcal{S} \times \mathcal{A}$ and $h \in [H],$
    $$
    \nabla_{\theta} f(\theta_h^*,\phi(s,a)) \Sigma_h^{-1} \nabla_{\theta}^{\top} f(\theta_h^*,\phi(s^{\prime},a^{\prime})) \geq 0,
    $$
\end{assumption}
Under positivity condition, we have with probability at least $1-\delta,$
\begin{align*}
    \left|\widehat{v}_{\pi} - v_{\pi}\right| \leq \left[\sum_{h_{1}=1}^{H} \sum_{h_{2}=1}^{H}\left(H-h_{1}+1\right)\left(H-h_{2}+1\right) \sqrt{\nu_{h_{1}}^{\top} \Sigma_{h_{1}}^{-1} \nu_{h_{1}}} \sqrt{\nu_{h_{2}}^{\top} \Sigma_{h_{2}}^{-1} \nu_{h_{2}} } \sigma_{h_{1}, h_{2}} \right]^{\frac{1}{2}} \sqrt{\frac{\log (12 / \delta)}{2 H K}} + O(\frac{1}{K}).
\end{align*}
where $\sigma_{h_{1}, h_{2}}:=\left\|\Sigma_{h_{1}}^{-\frac{1}{2}} \Sigma_{h_{1}, h_{2}} \Sigma_{h_{2}}^{-\frac{1}{2}}\right\|_{2}$ and
\begin{equation*}
    \Sigma_{h_{1}, h_{2}}:=\mathbb{E}\left[\frac{1}{H} \sum_{j=1}^{H} \nabla_{\theta}^{\top} f\left(\theta_{h_{1}}^*, \phi_j\right) \nabla_{\theta} f\left(\theta_{h_{2}}^*, \phi_j\right)\right].
\end{equation*}
Since $\sigma_{h_1,h_2} \leq 1,$ this bound is sharper than \eqref{finite_bound1}. When all covariance matrices are zero, all cross terms vanish and this bound is much better than \eqref{finite_bound1}.

Next we show that our results match the best known error bounds for FQE in the tabular case and in the case of linear function approximation.

\paragraph{Tabular Case} In tabular case, there are finite states and finite actions in the MDP we consider, and we can always represent all $Q$-functions in tables. We denote
\begin{equation}\label{definition_tilde_mu}
    \Tilde{\mu} := \frac{2}{H(H+1)} \mathbb{E}^{\pi} \left[\sum_{h=1}^H (H-h+1)\textbf{1}(s_h = s,a_h = a)\right],
\end{equation}
Then Corollary 1 in \citep{duan2020minimax} provided the upper bound in tabular case as 
\begin{equation*}
    \left|\widehat{v}_{\pi} - v\right|
    \leq C H^2 \sqrt{1 + \chi^2(\tilde{\mu},\bar{\mu})} \sqrt{\frac{\log (12 / \delta)}{2 K H}} + O(\frac{1}{K}),
\end{equation*}
where $C$ is a constant and $\chi^2$ is standard $\chi^2$ divergence. We give a outline sketch in the Appendix. Further \citep{duan2020minimax} proved the upper bound in time inhomogeneous MDP matches the upper bound in Theorem 3.1 in \citep{yin2020asymptotically}. The same result can be derived as a special case of our theorem. 

\paragraph{Linear Case} 
In case of linear function class, since all $\Sigma_h$ are the same, we can bound the dominant error term in another sharper way. In linear case, $\theta_h^*$ and $\phi(s,a)$ will have the same dimension, and all $\mathcal{G}_h$ will be $\mathcal{G} = \left\{\mu^{\top}\phi(\cdot) \mid \mu \in \mathbb{R}^d\right\}.$ Under the assumption $\phi^{\top}(s,a)\Sigma^{-1}\phi(s^{\prime},a^{\prime}) \geq 0,$ thee upper bound can be improved to
\begin{equation*}
    \left|\widehat{v}_{\pi} - v\right|
    \leq \frac{H(H+1)}{2} \sqrt{\frac{\log (12 / \delta)}{2 K H}} \sqrt{1 + \chi^2_{\mathcal{G}}(\Tilde{\mu},\bar{\mu})} + O(\frac{1}{K}).
\end{equation*}
This matches the minimax lower bound in \citep{duan2020minimax} and shows that in linear case this upper bound is nearly optimal.

\section{Bootstrapping FQE and Distributional Consistency}

FQE provides a point estimator. In order to quantify its uncertainty in practice, there is need to infer the estimator's distribution and to compute confidence intervals. 

\subsection{Bootstrapping General FQE Estimator}
We consider using bootstrap for statstical inference of FQE  \citep{efron1982jackknife}. By bootstrapping the FQE algorithm, we can subsample the data and get a series of bootstrapping FQE estimators. These bootstrap estimators allow us to estimate the error distribution and perform downstream inference tasks such as confidence intervals, variance estimation \citep{bootstrap}. We will bootstrap by episodes, instead of by transitions as in some previous research \citep{kostrikov2020statistical}. It was shown in \citep{bootstrap} that bootstrapping by transitions, which are dependent, for FQE might lead to inconsistency. 

Denote $W^\circ = (W_1,W_2,...,W_K)^{\top}$ as the bootstrap weights sampled according to certain distribution. 
In bootstrapping general FQE, we use $\widehat{Q}_h^\circ$ to denote bootstrapping estimator of Q functions. We let $\widehat{Q}^\circ_{H+1} (s,a) = 0$ and \eqref{FQE} is turned into
\begin{equation}\label{FQE_Bootstrapping}
    \widehat{Q}_h^\circ = \mathop{\arg \min}_{f \in \mathcal{F}} \bigg\{\frac{1}{2N} \sum_{k=1}^K W_k \sum_{h=1}^H \left[f(s_h^k,a_h^k) - y_h^{k\circ}\right]^2 + \lambda \rho(f)\bigg\}.
\end{equation}
where \begin{footnotesize} $y_{h}^{k\circ} = r\left(s_{h}^k, a_{h}^k\right)+\int_{\mathcal{A}} \widehat{Q}_{h+1}^\circ\left(s_{h+1}^k, a\right) \pi\left(a \mid s_{h+1}^k\right) d a.$ \end{footnotesize} Similar to \pref{sec:fqe}, under assumption of parameteric function space, we turn this problem into a minimization in parameter space and we denote $\widehat{Q}_h^\circ = f(\widehat{\theta}_h^\circ,\phi).$ Then \eqref{FQE_Bootstrapping} can be turned into
\begin{equation}\label{definition_theta_h_circ}
    \widehat{\theta}_h^\circ = \mathop{\arg \min}_{\theta \in \Theta} \bigg\{\sum_{k=1}^K \frac{W_k}{2N} \sum_{h=1}^H \left[f\left(\theta,\phi_h^k\right) - y_h^{k\circ}\left(\widehat{\theta}_{h+1}^{k\circ}\right)\right]^2 + \lambda \rho(\theta)\bigg\}.
\end{equation}
where \begin{scriptsize} $y_h^{k}\left(\widehat{\theta}_{h+1}^{\circ}\right) := \int_{\mathcal{A}} f\left(\widehat{\theta}_{h+1}^{\circ}, \phi\left(s_{h+1}^k, a^{\prime}\right)\right) \pi\left(a^{\prime} \mid s_{h+1}^k\right) \mathrm{d} a^{\prime}.$ \end{scriptsize} Then the general bootstrapping FQE estimator can be written as
\begin{align}\label{bootstrapping_FQE}
    \widehat{v}_{\pi}^\circ &= \int_{\mathcal{S} \times \mathcal{A}} f(\widehat{\theta}_1^\circ,\phi(s,a)) \pi(a \mid s) \xi(s) \mathrm{d} a \mathrm{d} s.
\end{align}
For $\theta \in \Theta^H,$ we denote $Z_K^\circ(\theta) := \frac{1}{K} \sum_{k=1}^K W_k z\left(\theta,\boldsymbol{\tau}_k\right).$ From the perspective of Z-Estimator, if we assume $\widehat{\theta}_h^\circ \in \operatorname{int}(\Theta)$ and write $\widehat{\theta}^\circ_{K,h} = \widehat{\theta}^\circ_{h}$ to explicate its dependency on $K,$ then we have when $\widehat{\theta}_K^\circ = (\widehat{\theta}_{K,1}^{\circ \top},...,\widehat{\theta}_{K,H}^{\circ \top})^{\top},$
\begin{equation*}
    Z_K^\circ \left(\widehat{\theta}_K^\circ\right) + \lambda R(\widehat{\theta}_K^\circ) = 0,
\end{equation*}
which implies $\widehat{\theta}_K^\circ$ is a bootstrapping Z-estimator. Next we consider standard bootstrap and one of its simple alternatives. Both bootstrapping FQE estimator have nice asymptotic properties and can be exploited to estimate variance and confidence interval.

\paragraph{Vanilla Bootstrap} When $W^\circ$ follows multinomial distribution with probability $(1/K,1/K,...,1/K),$ \eqref{bootstrapping_FQE} leads to vanilla bootstrapping FQE estimator. In other words, it amounts to resampling $K$ episodes $\boldsymbol{\tau}_1^{\circ},\boldsymbol{\tau}_2^{\circ},...,\boldsymbol{\tau}_K^\circ$ independently with replacement from the data $\{\boldsymbol{\tau}_k\}_{k=1}^{K},$ and solving \eqref{FQE} iteratively using resampled data. \citep{bootstrap} used vanilla bootstrapping strategy to construct asymptotic confidence interval of policy value, and proved that bootstrapping error has the same asymptotic distribution with standard FQE error.

\paragraph{Multiplier Bootstrap} Suppose $U$ is a non-constant non-negative random variable with finite mean $m > 0$ and variance $0 < \eta^2 < \infty,$ and $U$ satisfies
\begin{equation}\label{random_vector_U}
    \int_{0}^{\infty} \sqrt{P(|U|>x)} d x < \infty.
\end{equation}
This is slightly stronger than assuming a bounded second order moment, but is weaker than boundedness of $2+\varepsilon$ order moment for any $\varepsilon > 0.$ We let $u_1,u_2,...,u_K,..$ are an infinite series of i.i.d. samples from the distribution of $U$ and independent of all trajectories and $\bar{u}_K = \frac{1}{K}\sum_{k=1}^K u_k.$ Setting $W_k = u_k/\bar{u}_K$ in \eqref{FQE_Bootstrapping} and \eqref{bootstrapping_FQE} gives multiplier bootstrapping FQE estimator. For example, when $U$ follows standard exponential distribution, $W^\circ$ is called Dirichlet weights and the resulting multiplier bootstrap is Bayesian bootstrap\citep{rubin1981bayesian}.

\begin{algorithm}[htb!]
\caption{Bootstrapping General FQE}
\label{alg1}
	\begin{algorithmic}[1] 
		\Require Target policy $\pi$, initial state distribution $\xi,$ dataset $\mathcal{D}=\{(s^{k}_h,a^{k}_h,s^{k}_{h+1},r^{k}_h)\}_{h\in[H],k\in[K]}$. If multiplier bootstrap used, input random vector $U$ which satisfied \eqref{random_vector_U}.
		\If{Vanilla bootstrap} 
		\State Sample $(W_1,...,W_K) \sim $ multinomial distribution with probability $(1/K,...,1/K).$
	    \ElsIf{Multiplier Bootstrap} 
	    \State Sample $u_k \sim U, W_k = \frac{K u_k}{\sum_{k=1}^K u_k}.$
	    \EndIf
		\State \textbf{Initialize } $\widehat{\theta}_{H+1}^\circ = 0.s$
		\For{$h=H,H-1,\ldots,1$}  
		\State Solve \eqref{definition_theta_h_circ}
		\EndFor
		\State\textbf{Return} $\widehat{v}_{\pi}^{\circ}=\int_{\mathcal{S} \times \mathcal{A}} f\left(\widehat{\theta}_{1}^{\circ}, \phi(s, a)\right) \pi(a \mid s) \xi(s) \mathrm{d} a \mathrm{d} s.$
	\end{algorithmic}
\end{algorithm}

\subsection{Distributional Consistency}
We show that bootstrapping FQE with differential function approximator is distributionally consistent, i.e. the limit distribution of bootstrapping error $\widehat{\theta}^{\circ}_K - \widehat{\theta}_K$ conditional on batch data is that of standard FQE error $\widehat{\theta}_K - \theta^*$ established in Theorem \ref{thm_normality} or a multiplication of it.

\begin{theorem}[Distributional Consistency]\label{thm_consistency}
    Under the assumptions of Theorem \ref{thm_normality}, when $K \to \infty$ and $\lambda = o(K^{-1/2})$, conditional on $\left\{\boldsymbol{\tau}_k\right\}_{k=1}^K$, we have
    $$
    \sqrt{K} \left(\widehat{v}_{\pi}^{\circ} - \widehat{v}_{\pi}\right) \stackrel{d}{\longrightarrow} \mathcal{N}(0,k_0\sigma^2).
    $$
    for the same $\sigma$ as defined in Eq.~\eqref{variance_expression}. For vanilla bootstrapping FQE estimator, $k_0 = 1.$ For multiplier bootstrapping FQE estimator, $ k_0 = \frac{\eta^2}{m^2}.$
\end{theorem}

This distribution consistency certifies that we may use the bootstrapping FQE to approximate the estimation error distributions and inference tasks. For example, we may estimate the confidence interval (CI) of the FQE by taking quantiles; see Algorithm \ref{alg1}.

Suppose our batch data $\mathcal{D}$ is generated from a probability space $\left(\mathcal{X}^H, \Sigma\left(\mathcal{X}^H\right), \mathbb{P}_{\mathcal{D}}\right)$, and the bootstrap weight $W^{\circ}$ is from an independent probability space $\left(\mathcal{W}, \Omega, \mathbb{P}_{W}\right)$. Their joint probability measure is $\mathbb{P}_{\mathcal{D} W^{\circ}}$. Let $\mathbb{P}_{W^{\circ} \mid \mathcal{D}}$ denote the conditional distribution once the data $\mathcal{D}$ is given. Next corollary shows consistency of the CI estimate. 
\begin{corollary}\label{coro_CI}
    Denote the lower $\delta$-th quantile of bootstrapping error distribution $$q_{\delta}^{\pi}=\inf \left\{t: \mathbb{P}_{W^{\circ} \mid \mathcal{D}} \left(\widehat{v}_{\pi}^{\circ}-\widehat{v}_{\pi} \leq t\right) \geq \delta\right\} .$$ 
    We construct the $1-\delta$ confidence interval of the policy value by: $\operatorname{CI}(\delta)=\left[\widehat{v}_{\pi} - \frac{1}{\sqrt{k_0}} q_{1 - \delta / 2}^{\pi}, \widehat{v}_{\pi}- \frac{1}{\sqrt{k_0}} q_{\delta / 2}^{\pi}\right] .$ Then we have when $K \to \infty,$
    \begin{small}
    $$
    \mathbb{P}_{\mathcal{D} W^{\circ}}\left(v_{\pi} \in \mathrm{CI}(\delta)\right) \rightarrow 1-\delta.$$
    \end{small}
\end{corollary}
This gives a convenient and provable way to construct confidence interval for OPE. 

\begin{algorithm}[htb!]
\caption{Bootstrapping Confidence Interval}
\label{alg1}
	\begin{algorithmic}[1] 
		\Require Target policy $\pi$, initial state distribution $\xi,$ dataset $\mathcal{D}=\{(s^{k}_h,a^{k}_h,s^{k}_{h+1},r^{k}_h)\}_{h\in[H],k\in[K]},$ confidence level $\delta,$ number of bootstrap $B.$ If multiplier bootstrap used, input random vector $U$ which satisfied \eqref{random_vector_U}.
		\State Compute standard FQE estimator $\widehat{v}(\mathcal{D}).$
		\For{$b = 1,2,\ldots,B$}  
		\State Sample a bootstrapping weight $W_b^\circ$ according to multinomial distribution or random variable $U.$
		\State Compute $\widehat{v}\left(\mathcal{D};W_b^\circ\right).$
		\EndFor
		\State Compute errors $\varepsilon_b := \widehat{v}\left(\mathcal{D};W_b^\circ\right) - \widehat{v}(\mathcal{D})$ for $b \in [B].$
		\State Compute $\delta/2$ and $1 - \delta/2$ empirical quantile of $\left\{\varepsilon_1,...,\varepsilon_B\right\},$ denoted as $\widehat{q}^{\pi}_{\delta/2}$ and $\widehat{q}^{\pi}_{1 - \delta/2}.$
		\State \textbf{Return} $\left[\widehat{v}\left(\mathcal{D}\right) - \widehat{q}^{\pi}_{1 - \delta/2}, \widehat{v}\left(\mathcal{D}\right) - \widehat{q}^{\pi}_{\delta/2}\right].$
	\end{algorithmic}
\end{algorithm}

\section{Information-Theoretic Lower Bound}
An estimator is called asymptotic efficient if its variance is minimal among all unbiased estimators. Cramer-Rao lower bound is a theoretical lower bound for variance of all unbiased estimators. We will show that the general FQE estimator can achieve Cramer-Rao lower bound for variance and is therefore asymptotically efficient.

\begin{theorem}[Cramer Rao Lower Bound]\label{thm_CR}
    Under assumptions of Theorem \ref{thm_normality}, the variance of any unbiased estimator of $v_{\pi}$ is lower bounded by $\sigma^2$ in \eqref{variance_expression}.
\end{theorem}
This lower bound generalize several previous results. To our knowledge, \citep{JiangL15} established the first Cramer-Rao lower bound for off policy evaluation in tabular MDP. \citep{bootstrap} established the lower bound and proved asymptotic efficiency for using linear function approximation. Our proof is based on influence function. We first compute the influence function of $v_{\pi}$ and then show the expectation of squared influence function equals to $\sigma^2.$

\section{Summary and Future Work}

This paper studies statistical properties of fitted Q evaluation using compact, differentiable function approximators. It establishes a set of statistical results including asymptotic normality, finite-sample error bounds, characterization of distribution shift and Cramer-Rao lower bound. Differentiability is a key assumption in our analysis, but it is not difficult to generalize our result to almost everywhere differentiable approximators such as ReLU neural networks. Besides our results for general function approximators, we want to emphasize that Z estimator may be a useful tool for analysis in other off-policy RL problems. For less smooth approximators such as Holder functions, M estimators instead of Z Estimator can provide a more powerful analysis.
\newpage

\bibliography{arxiv}
\bibliographystyle{plainnat}

\appendix
\onecolumn

\section{Proof of Asymptotic Normality and Distributional Consistency}
\subsection{Proof of Theorem \ref{thm_normality} and Theorem \ref{thm_consistency}}
Since Theorem \ref{thm_normality} and Theorem \ref{thm_consistency} are proven by the same technique, we will prove them simultaneously.

\subsubsection{Preliminaries on Glivenko Cantelli and Donsker Function Class}\label{sec:prep}
Before we present our main result, we prove some basic properties of the following function class. Below $z_{h,i}(\theta,\cdot)$ is the $i-$th entry of $z_{(h)}(\theta,\cdot)$ for $i \in [d].$
\begin{equation}\label{definition_Z}
    \mathcal{Z} := \bigcup_{h=1}^{H} \bigcup_{i=1}^{d} \left\{z_{h,i}\left(\theta,\cdot\right): \theta \in \Theta^H \subset \mathbb{R}^{Hd} \right\}.
\end{equation}
We remark that each function in $\mathcal{Z}$ is a map from $( \mathcal{S} \times \mathcal{A} )^H$ to $\mathbb{R}$ and is indexed by $\theta \in \Theta^H.$ We present the standard definition of $\mathbb{P}$-Glivenko Cantelli and $\mathbb{P}$-Donsker in asymptotic statistics as follows. In the definition and following proof, we denote $\mathbb{P}$ as a random probability measure and $\mathbb{P}_n$ as its empirical measure. We further denote $\mathbb{G}_n(\cdot) := \sqrt{n} \left(\mathbb{P}_n(\cdot) - \mathbb{P}(\cdot)\right)$ as empirical process.

\begin{definition}[P-Glivenko-Cantelli and P-Donsker\citep{van}]
    Suppose $\mathcal{U}$ is a measurable function class. We denote $l^{\infty}(\mathcal{U})$ as the set of all bounded functions on $\mathcal{U}.$ We call $\mathcal{U}$ is ($\mathbb{P}$-)Glivenko-Cantelli (or equivalently, strong Glivenko-Cantelli), if 
\begin{equation}\label{definition_GC}
    \left\|\mathbb{P}_n u - \mathbb{P} u\right\|_{\mathcal{U}} := \sup_{u \in \mathcal{U}} \left|\mathbb{P}_n u - \mathbb{P} u\right| \stackrel{a.s.}{\longrightarrow} 0;
\end{equation}
 We call a $\mathcal{U}$ of measurable functions is ($\mathbb{P}$-)Donsker, if the sequence of processes $\left\{\mathbb{G}_n u: u \in \mathcal{U}\right\}$ converges in distribution to a tight limit process $G$ in $l^{\infty}(\mathcal{U}).$ By converging in distribution in $l^{\infty}\left(\mathcal{U}\right)$, we mean for every bounded, continuous function $g: l^{\infty}(\mathcal{U}) \to \mathbb{R},$ it always holds that $\mathbb{E} [g(\mathbb{G}_n u)] \to \mathbb{E}[g(G)]$ when $n$ tends to infinity. When there is only one probability measure $\mathbb{P},$ we omit the prefix $\mathbb{P}$- in front of Glivenko-Cantelli or Donsker. We define a class $\mathcal{U}$ of vector-valued functions $u: x \to \mathbb{R}^k$ to be Glivenko-Cantelli or Donsker if the union of the $k$ coordinate classes is Glivenko-Cantelli or Donsker.
\end{definition}

Glivenko-Cantelli and Donsker function classes are difficult to verify by definition. A usual sufficient condition involves the concept of bracketing number and bracketing integral.

\begin{definition}[Bracketing Number and Bracketng Integral]
    For a function $u \in \mathcal{U},$ we define its $L_r(\mathbb{P})$-norm as $\left\| u \right\|_{\mathbb{P},r} = (\mathbb{P} \left|u \right|^r)^{1/r}.$ Given two functions $l$ and $h$ with finite $L_r(\mathbb{P})$-norm (need not to be in $\mathcal{U}$), we define the bracket $[l,h]$ as all functions $f$ with $l \leq f \leq h$ in the whole domain. 
    An $\varepsilon$-bracket in $L_r(\mathbb{P})$ is a bracket $[l,h]$ with $\mathbb{P}|h-l|^r < \varepsilon^r.$ We define the bracketing number $N_{[]}(\varepsilon,\mathcal{U},L_r(\mathbb{P}))$ as the minimal number of $\varepsilon$- brackets needed to cover $\mathcal{U},$ i.e. each function $u \in \mathcal{U}$ is contained in a $\varepsilon$- bracket. The speed of growth of bracketing number are described using the bracketing integral.
    \begin{equation} \label{definition_bracketing_integral}
        J_{[]}\left(\delta,\mathcal{U},L_r(\mathbb{P})\right)  = \int_0^{\delta} \sqrt{\log N_{[]}\left(\varepsilon,\mathcal{U},L_r(\mathbb{P})\right)} d \varepsilon.
    \end{equation}
\end{definition}

Next two propositions give us sufficient conditions for a measurable function class to be Glivenko-Cantelli or Donsker.

\begin{proposition}[Glivenko-Cantelli \citep{empirical_process}] \label{GC_thm}
    Every function class $\mathcal{U}$ of measurable functions such that for any $\varepsilon > 0,$ 
    $$N_{[]}(\varepsilon,\mathcal{U},L_1(\mathbb{P})) < \infty$$ 
    is $\mathbb{P}$-Glivenko-Cantelli. 
\end{proposition}

\begin{proposition}[Donsker\citep{empirical_process}] \label{Donsker_thm}
    Every function class $\mathcal{U}$ of measurable functions with
    $$
    J_{[]}\left(1,\mathcal{U},L_2(\mathbb{P})\right) < \infty
    $$
    is $\mathbb{P}$-Donsker.
\end{proposition}

It is obvious that Glivenko Cantellli and Donsker function class has the following relation.

\begin{proposition}\label{GC_and_Donsker}
    Every Donsker function class is Glivenko Cantelli.
\end{proposition}

\subsubsection{Z-Estimator Master Theorem}
Our central tool will be the following Z-Estimator Master Theorem. It provides sufficient condition for asymptotic normality of Z-estimators and bootstrapping Z-estimators. We use this lemma to prove both asymptotic normality and distributional consistency.

\begin{lemma}[Z-Estimator Master Theorem\citep{kosorok}]\label{Z_Estimator_thm}
Let $u(\theta,\boldsymbol{\tau}): \mathcal{X}^H \rightarrow \mathbb{R}^{Hd}$ be a Borel-measurable function with $\boldsymbol{\tau} \in \mathcal{X}^H.$ Let 
$$U(\theta)=\mathbb{E}_{\boldsymbol{\tau}} u(\theta,\boldsymbol{\tau}), \ U_{K}(\theta) = \frac{1}{K}\sum_{k=1}^K u(\theta,\boldsymbol{\tau}_k), \ U_{K}^\circ(\theta) = \frac{1}{K}\sum_{k=1}^K W_k u(\theta,\boldsymbol{\tau}_k),$$
where $W^\circ = \left(W_1,W_2,...,W_K\right)^{\top}$ is either vanilla bootstrapping weights or multiplier bootstrapping weights. When $W^\circ$ is vanilla bootstrapping weights, $W^\circ$ follows multinomial distribution with parameter $\left(1/K,1/K,...,1/K\right).$ When $W^\circ$ is multiplier bootstrapping weights, we have $W_k = \frac{K u_k}{\bar{u}_K},$ where $u_k$ are i.i.d samples from distribution of random vector $U$ which satisfies \eqref{random_vector_U}.

Let $\Theta \subset \mathbb{R}^{d}$ be compact, and assume $\theta^* \in \operatorname{int}(\Theta^H)$ satisfies $U\left(\theta^*\right)=0 .$ Assume the following:
\begin{itemize}
    \item (i) For any sequence $\left\{\theta^l\right\} \in \operatorname{int} \Theta^H, U\left(\theta^l\right) \rightarrow 0$ implies $\left\|\theta^l-\theta^*\right\| \rightarrow 0$;
    \item (ii) The class $\mathcal{U} = \left\{u(\theta,\boldsymbol{\tau}): \theta \in \Theta^H \right\}$ is P-Glivenko-Cantelli;
    \item (iii) For some $\delta>0$, the class $\mathcal{U}_{\delta}:=\left\{u(\theta,\boldsymbol{\tau}): \theta \in \Theta^H ,\left\|\theta-\theta^*\right\| \leq \delta\right\}$ is $P$-Donsker 
    \item (iv) $\mathbb{E} \left\|u(\theta ,\boldsymbol{\tau}) -u(\theta^*,\boldsymbol{\tau})\right\|^{2} \rightarrow 0$ as $\left\|\theta-\theta^*\right\| \rightarrow 0.$
    \item (v) $\mathbb{E} \left\|u(\theta^*,\boldsymbol{\tau})\right\|^{2}<\infty$, and $U(\theta)$ is differentiable at $\theta^*$ with non-singular Jacobian matrix $V(\theta^*).$
    \item (vi) $U_K(\widehat{\theta}_K) = o(K^{-1/2})$ and $U_K^\circ(\widehat{\theta}_K^{\circ}) = o(K^{-1/2}).$ 
\end{itemize}
Then, we have
$$
\sqrt{K}\left(\widehat{\theta}_K-\theta^*\right) \stackrel{d}{\rightarrow} N\left(0, V(\theta^*)^{-1} \mathbb{E}\left[u(\theta^*,\boldsymbol{\tau}) u(\theta^*,\boldsymbol{\tau})^{\top}\right]\left[V(\theta^*)^{-1}\right]^{\top}\right)
$$
and
$$
\sqrt{K}\left(\widehat{\theta}_K^{\circ}-\widehat{\theta}_K\right) \stackrel{d}{\rightarrow} N\left(0, k_0 V(\theta^*)^{-1} \mathbb{E}\left[u(\theta^*,\boldsymbol{\tau}) u(\theta^*,\boldsymbol{\tau})^{\top}\right]\left[V(\theta^*)^{-1}\right]^{\top}\right).
$$
conditionally on $\boldsymbol{\tau}_1,...,\boldsymbol{\tau}_K.$ When we use vanilla bootstrap, $k_0 = 1;$ when we use multiplier bootstrap, $k_0 = \frac{\eta^2}{m^2},$ where $m$ and $\eta^2$ are population mean and variance of distribution of bootstrapping weights.
\end{lemma}

\subsubsection{Completion of the Proof of Asymptotic Normality}
All we need to do is to verify that function class $\mathcal{Z}$ satisfies all conditions in Z-Estimator Master Theorem. From compactness of $\Theta$ and the uniqueness of root of $Z(\theta)$ in the assumption of Theorem \ref{thm_normality}, (i) and (v) is verified. By continuity of $z(\theta,\boldsymbol{\tau})$ with respect to $\theta,$ (iv) holds. (vi) holds by letting $\lambda = o(K^{-1/2}).$ 

Next we will prove (ii) and (iii). Since every Donsker function class is Glivenko-Cantelli, we only need to prove (iii) for any $\delta > 0.$ Then (ii) can be proven by letting $\delta = \operatorname{diam}(\Theta).$ Since $f(\theta,\phi)$ is third times continuously differentiable in a compact set $\Theta \times \Psi,$ $f(\theta,\phi)$ has continuous derivatives up to third order. 
We denote $\alpha = (\alpha_1,\alpha_2,...,\alpha_d)$ as a multi-index. 
We define the order of $\alpha$ or the degree of $\alpha$ as
\begin{equation*}
    |\alpha|=\alpha_{1}+\alpha_{2}+\cdots+\alpha_{n}
\end{equation*}
and
\begin{equation}\label{notation1}
    \partial^{\alpha}_{\theta} f(\theta,\phi) = \partial_{\theta_1}^{\alpha_{1}} \partial_{\theta_2}^{\alpha_{2}} \cdots \partial_{\theta_d}^{\alpha_{d}} f=\frac{\partial^{|\alpha|} f}{\partial \theta_{1}^{\alpha_{1}} \partial \theta_{2}^{\alpha_{2}} \cdots \partial \theta_{d}^{\alpha_{d}}}
\end{equation}
We denote for $l = 1,2,3,$
\begin{equation}
    \kappa_{l} := \sup_{\theta \in \Theta, \phi \in [0,1]^m} \sup_{\left|\alpha\right| = l} \left| \partial^{\alpha}_{\theta} f(\theta,\phi) \right| < \infty.
\end{equation}
Then
\begin{equation*}
    \sup_{\theta \in \Theta, \phi \in [0,1]^m} \left\|\nabla_{\theta} f(\theta,\phi)\right\|_2 \leq \sqrt{d} \kappa_1,
\end{equation*}
and
\begin{equation*}
    \sup_{\theta \in \Theta, \phi \in [0,1]^m} \left\| \nabla_{\theta}^2 f(\theta,\phi)\right\|_2 \leq \sup_{\theta \in \Theta, \phi \in [0,1]^m} \left\| \nabla_{\theta}^2 f(\theta,\phi)\right\|_{\operatorname{F}} \leq d \kappa_2.
\end{equation*}
where $\left\|\cdot\right\|_{\operatorname{F}}$ is Frobenius norm. This implies $\left\|f(\theta,\phi) - f(\theta^{\prime},\phi)\right\|_2 \leq \sqrt{d} \kappa_1 \left\|\theta - \theta^{\prime}\right\|_2$ and $\left\|\nabla_{\theta}f(\theta,\phi) - \nabla_{\theta} f(\theta^{\prime},\phi)\right\|_2 \leq d \kappa_2 \left\|\theta - \theta^{\prime}\right\|_2$ for every $\theta,\theta^{\prime} \in \Theta$ and $\phi \in \Psi.$ In the following proof, we denote $\theta^i = (\theta_1^{i \top},\theta_2^{i \top},...,\theta_H^{i \top})^{\top} \in \Theta^H$ for $i = 1,2$ and $\theta_{H+1}^i = 0.$ We use $\boldsymbol{\tau}$ to represent a random trajectory. We define
$$
\varepsilon_{j,h} := f\left(\theta_{j}^*, \phi(s_{h}, a_{h})\right) - r\left(s_{h}, a_{h}\right) - \int_{\mathcal{A}} f\left(\theta_{j+1}^*, \phi(s_{h+1}, a^{\prime})\right) \pi\left(a^{\prime} \mid s_{h+1}\right) d a^{\prime},
$$
then since for any $(s,a) \in \mathcal{S} \times \mathcal{A},$ $r(s,a) \in [0,1],$ we have $Q_h(s,a) = f(\theta_h^*,\phi(s,a)) \in [0,H-h+1]$ and $\left|\varepsilon_{j,h}\right| \leq H-h+1$ for $h,j \in [H].$ Further for $i = 1,2$ we define
$$
Y_{j, h}^{i}:= f\left(\theta_{j}^{i}, \phi(s_{h}, a_{h})\right) - r\left(s_{h}, a_{h}\right) - \int_{\mathcal{A}} f\left(\theta_{j+1}^{i}, \phi(s_{h+1}, a)\right) \pi\left(a \mid s_{h+1}\right) d a
$$
Therefore, we have
\begin{align*}
&\left|z\left(\theta^{1}, \boldsymbol{\tau}\right)-z\left(\theta^{2}, \boldsymbol{\tau}\right)\right|^2 \\
\leq & \sum_{j=1}^{H}\left|\sum_{h=1}^{H} Y_{j, h}^{1} \cdot \nabla_{\theta} f\left(\theta_{j}^{1}, \phi(s_{h}, a_{h})\right)-Y_{j, h}^{2} \cdot \nabla_{\theta} f\left(\theta_{j}^{2}, \phi(s_{h}, a_{h})\right)\right|^2 \\
\leq & 2\sum_{j=1}^{H}\left|\sum_{h=1}^{H} Y_{j, h}^{1}\left[\nabla_{\theta} f\left(\theta_{j}^{1}, \phi(s_{h}, a_{h})\right)-\nabla_{\theta} f\left(\theta_{j}^{2}, \phi(s_{h}, a_{h})\right)\right]\right|^2 + 2 \sum_{j=1}^{H}\left|\sum_{h=1}^{H}\left(Y_{j, h}^{1}-Y_{j, h}^{2}\right) \nabla_{\theta} f\left(\theta_{j}^{2}, \phi(s_{h}, a_{h})\right)\right|^2 \\
\leq & 2 d^2 \kappa_2^2 \sum_{j=1}^{H} \left(\sum_{h=1}^{H} \left|Y^1_{j,h}\right| \cdot \left|\theta^1_j - \theta^2_j \right|\right)^2 + 2 d \kappa_1^2 \sum_{j=1}^{H} \left(\sum_{h=1}^{H} \left|Y^1_{j,h} - Y^2_{j,h}\right| \right)^2.
\end{align*}
Next we bound $Y_{j,h}^i.$ Since
\begin{align*}
    \left|Y_{j,h}^1\right|
    & \leq \left|\varepsilon_{j,h}\right| + \left|\varepsilon_{j,h} - Y^1_{j,h}\right| \\
    & \leq (H - h + 1) + \kappa_1 \sqrt{d} \left[\left|\theta_j^* - \theta_j^1\right| + \left|\theta_{j+1}^* - \theta_{j+1}^1\right| \right] \leq H + 2 \kappa_1 \sqrt{d} \delta.
\end{align*}
and 
\begin{equation*}
    \left|Y^1_{j,h} - Y^2_{j,h}\right| \leq \kappa_1 \sqrt{d} \left[ \left|\theta_j^1 - \theta_j^2\right| + \left|\theta_{j+1}^1 - \theta_{j+1}^2\right| \right],
\end{equation*}
we have
\begin{align*}
&\left|z\left(\theta^{1}, \boldsymbol{\tau}\right)-z\left(\theta^{2}, \boldsymbol{\tau}\right)\right|^2 \\
\leq & 2 d^2 \kappa_2^2 \sum_{j=1}^H \left[\left(H^2 + 2\kappa_1\delta H\right)^2 \left(\sum_{h=1}^H\left|\theta^1_j - \theta^2_j \right|^2\right)\right] + 2 \kappa_1^4 H^2 \left(\sum_{j=1}^H \left|\theta_j^1 - \theta_j^2\right| + \left|\theta_{j+1}^1 - \theta_{j+1}^2\right| \right)^2\\
\leq & \left[2 \kappa_2^2 \left(H^2 + 2\kappa_1\sqrt{d} \delta H\right)^2 H + 8 d^2 \kappa_1^4 H^3\right] \left\|\theta^1 - \theta^2\right\|_2^2.
\end{align*}
We denote $C_1 > 0$ as 
where
\begin{equation}\label{definition_C}
    C_1^2 = 2 d^2 \kappa_2^2 \left(H^2 + 2\kappa_1\sqrt{d}\delta H\right)^2 H + 8 d^2 \kappa_1^4 H^3.
\end{equation}
Then $\left|z\left(\theta^{1}, \boldsymbol{\tau}\right)-z\left(\theta^{2}, \boldsymbol{\tau}\right)\right| \leq C_1 \left\|\theta^1 - \theta^2\right\|_2.$ Therefore, for every $\left\|\theta - \theta^* \right\| \leq \delta,$ we consider the high-dimensional bracket 
    $$[z(\theta,\boldsymbol{\tau}) - C_1 \sqrt{Hd} \boldsymbol{\varepsilon}, z(\theta,\boldsymbol{\tau}) + C_1 \sqrt{Hd} \boldsymbol{\varepsilon} ],$$ 
    where $\boldsymbol{\varepsilon}$ is the $Hd$-dimension vector with every entry being $\varepsilon.$ Since the brackets we consider are one-dimensional, these brackets have $L_2(\mathbb{P})$-size of $2 C_1 \sqrt{Hd} \varepsilon.$ If $\left\|\theta^1 - \theta^2\right\|_{\infty} \leq \varepsilon,$ then $\left|z\left(\theta^{1}, \boldsymbol{\tau}\right)-z\left(\theta^{2}, \boldsymbol{\tau}\right)\right| \leq \sqrt{Hd} C_1 \varepsilon,$ which implies $z(\theta^2,\boldsymbol{\tau}) \in [z(\theta^1,\boldsymbol{\tau}) - C_1 \sqrt{Hd} \boldsymbol{\varepsilon}, z(\theta^1,\boldsymbol{\tau}) + C_1 \sqrt{Hd} \boldsymbol{\varepsilon}],$ and every dimension of $z(\theta^2,\boldsymbol{\tau})$ is contained in the corresponding one-dimensional bracket. And since every dimension of $\theta^1 - \theta^2$ spans within a distance no more than $2\delta,$ we can partition every dimension of $\left\|\theta - \theta^* \right\| \leq \delta$ into grids with meshwidth being $2\varepsilon.$ By dividing like this, we will get no more than $(\frac{\delta}{\varepsilon})^{Hd}$ hypercubes. For every $\left\|\theta - \theta^* \right\| \leq \delta,$ there exists one hypercube whose center $\theta^c$ satisfies $\left\|\theta - \theta^c\right\|_{\infty} \leq \varepsilon.$ Then we have $z(\theta,\boldsymbol{\tau}) \in [z(\theta^c,\boldsymbol{\tau}) - C_1 \sqrt{Hd} \boldsymbol{\varepsilon}, z(\theta^c,\boldsymbol{\tau}) + C_1 \sqrt{Hd} \boldsymbol{\varepsilon}],$ and every component of $z(\theta,\boldsymbol{\tau})$ is contained in the corresponding one-dimensional bracket. Therefore, every function of $\mathcal{Z}_{\delta}$ is contained in a bracket. Summing over all brackets, we have
    \begin{equation*}
        N_{[]}\left(2 C_1 \sqrt{Hd} \varepsilon, \mathcal{Z}_{\delta}, L_r(\mathbb{P})\right) \leq Hd \left(\frac{\delta}{\varepsilon}\right)^{Hd}.
    \end{equation*}
    This implies 
    \begin{equation}\label{bracketing_num_result}
        N_{[]}\left(\varepsilon, \mathcal{Z}_{\delta}, L_r(\mathbb{P})\right) \leq Hd \left(\frac{2 C_1 \sqrt{Hd} \delta}{\varepsilon}\right)^{Hd}.
    \end{equation}
    Because that the bracketing number is a decreasing function of $\varepsilon,$ whether the bracketing integral converges or not depends on the asymptotic behavior of bracketing number for $\varepsilon \to 0.$ From the estimation above, $\sqrt{\log N_{[]} (\varepsilon,\mathcal{Z}_{\delta},L_2(\mathbb{P}))}$ is of order $O(\sqrt{\log(\frac{1}{\varepsilon})}),$ the integral of which converges. By Glibenko-Cantelli's Theorem and Donsker's Theorem, we know $\mathcal{Z}$ is both Glivenko Cantelli and Donsker. In this way, we verify all conditions of Z-Estimator Master Theorem and we have
        \begin{equation}\label{finite_normal_convergence1}
        \sqrt{K}\left(\widehat{\theta}_K - \theta^*\right) \stackrel{d}{\longrightarrow} N\left(0,\frac{1}{H} \Sigma^{-1}_* \Omega \Sigma^{-\top}_*\right);
    \end{equation}
    and conditioned on $\boldsymbol{\tau}_1,\boldsymbol{\tau}_2,...,\boldsymbol{\tau}_K,$
    \begin{equation}\label{finite_normal_convergence_bootstrapping}
        \sqrt{K}\left(\widehat{\theta}^{\circ}_K - \widehat{\theta}_K\right) \stackrel{d}{\longrightarrow} N\left(0,\frac{k_0}{H} \Sigma^{-1}_* \Omega \Sigma^{-\top}_*\right);
    \end{equation}
    where 
    \begin{equation}\label{definiton_Sigma_and_Omega}
        \Sigma_* = \frac{1}{H} \left.\frac{\partial}{\partial \xi} Z\left(\xi\right)\right|_{\xi = \theta^*} \text{ and }
        \Omega = \frac{1}{H} \left.\mathbb{E}_{\boldsymbol{\tau}} \left[z(\xi,\boldsymbol{\tau}) z(\xi,\boldsymbol{\tau})^{\top}\right]\right|_{\xi = \theta^*}.
    \end{equation}
    When we adopt vanilla bootstrap strategy, $k_0 = 1,$ when we adopt multiplier bootstrap, $k_0 = \frac{\eta^2}{m^2}.$ Finally we compute the covariance matrix in the asymptotic variance. Below, we write out $\Sigma_*$ as a partitioned matrix:
\begin{equation*}\label{expression_Sigma}
    \Sigma_* = \begin{pmatrix}
    \Sigma_{1} & -A_1 & 0 & \ldots & 0 \\
    0 & \Sigma_2 & -A_2 & \ldots & 0 \\
    0 & 0 & \Sigma_3 & \ldots & 0\\
    \vdots & \vdots & \vdots & \vdots & \vdots \\
    0 & 0 & 0 & \ldots & \Sigma_H
    \end{pmatrix}
\end{equation*}
where for $h \in [H],$
\begin{align}
    &\Sigma_h
    = \frac{1}{H} \mathbb{E}_{\boldsymbol{\tau}} \left\{\sum_{h^{\prime}=1}^{H} \left(f(\theta_h^*,\phi(s_{h^{\prime}},a_{h^{\prime}})) -  r(s_{h^{\prime}},a_{h^{\prime}}) - \int_{\mathcal{A}} f(\theta_{h+1}^*,\phi(s_{h^{\prime}+1},a)) \pi(a \mid s_{h^{\prime}+1}) d a\right)\nabla_{\theta \theta}^2 f(\theta_h^*,\phi(s_{h^{\prime}},a_{h^{\prime}}))\right. \notag \\
    & \hspace{10ex}+\left.\sum_{h^{\prime}=1}^{H} \bigg(\nabla_{\theta} f(\theta_h^*,\phi(s_{h^{\prime}},a_{h^{\prime}}))\bigg)^{\top} \bigg(\nabla_{\theta} f(\theta_h^*,\phi(s_{h^{\prime}},a_{h^{\prime}}))\bigg)\right\}. \notag
\end{align}
Since 
\begin{equation*}
    \mathbb{E}_{\boldsymbol{\tau}} \left\{ f(\theta_h^*,\phi(s_{h^{\prime}},a_{h^{\prime}})) - r(s_{h^{\prime}},a_{h^{\prime}}) - \int_{\mathcal{A}} f(\theta_{h+1}^*,\phi(s_{h^{\prime}+1},a)) \pi(a \mid s_{h^{\prime}+1}) d a \bigg| s_{h^{\prime}}, a_{h^{\prime}}\right\} = 0,
\end{equation*}
by taking conditional expectation we find the first term in the expectation vanishes and 
\begin{equation}\label{appendix_definition_Sigma_h}
    \Sigma_h
    = \frac{1}{H} \mathbb{E}_{\boldsymbol{\tau}} \left\{\sum_{h^{\prime}=1}^{H} \bigg(\nabla_{\theta} f(\theta_h^*,\phi(s_{h^{\prime}},a_{h^{\prime}}))\bigg)^{\top} \bigg(\nabla_{\theta} f(\theta_h^*,\phi(s_{h^{\prime}},a_{h^{\prime}}))\bigg)\right\}.
\end{equation}
For $h \in [H-1],$
\begin{align*}\label{expression_A}
    A_h 
    &= - \frac{1}{H}\nabla_{\beta} \mathbb{E}_{\boldsymbol{\tau}} \left\{ \sum_{h^{\prime}=1}^H \left(f(\alpha, \phi(s_{h^{\prime}},a_{h^{\prime}})) -  r(s_{h^{\prime}},a_{h^{\prime}}) - \int_{\mathcal{A}} f(\beta,\phi(s_{h^{\prime}+1},a)) \pi(a \mid s_{h^{\prime}+1}) d a\right)\nabla_{\theta} f(\alpha,\phi(s_{h^{\prime}},a_{h^{\prime}})) \right\}\bigg|_{\alpha = \theta_h^*, \beta = \theta_{h+1}^*}.
\end{align*} 
Here the gradient means partial derivatives with respect to $\beta$ and take $\alpha = \theta_h^*, \beta = \theta_{h+1}^*.$ Therefore, the inverse of $\Sigma_*$ can be expressed as
\begin{equation*}
    \Sigma^{-1}_* = \begin{pmatrix}
    \Sigma^{-1}_1 & \Sigma^{-1}_1 A_1 \Sigma^{-1}_2 & \Sigma^{-1}_1 A_1 \Sigma^{-1}_2 A_2 \Sigma^{-1}_3 & \ldots & \Sigma^{-1}_1 A_1 \Sigma^{-1}_2 ... A_{H-1}\Sigma^{-1}_H \\
    0 & \Sigma^{-1}_2 & \Sigma^{-1}_2 A_2 \Sigma^{-1}_3 & \ldots & \Sigma^{-1}_2 A_2 \Sigma^{-1}_3 ... A_{H-1} \Sigma^{-1}_H \\
    0 & 0 & \Sigma^{-1}_3 & \ldots & \Sigma^{-1}_3 A_3 \Sigma^{-1}_4 ... A_{H-1} \Sigma^{-1}_H \\
    \vdots & \vdots & \vdots & \ddots & \vdots \\
    0 & 0 & 0 & \ldots & \Sigma^{-1}_H \\
    \end{pmatrix}.
\end{equation*}
Moreover we express $\Omega$ as $(\Omega_{i,j})_{i,j \in [H]}$ where $\Omega_{i,j} \in \mathbb{R}^{d \times d}$ is
\begin{equation*}
    \Omega_{i,j}=\frac{1}{H} \sum_{h_{1}=1}^{H} \sum_{h_{2}=1}^{H} \mathbb{E}\left[\bigg(\nabla_{\theta} f\left(\theta_{i}^*, \phi(s_{h_{1}}, a_{h_{1}})\right)\bigg)^{\top}\bigg(\nabla_{\theta} f\left(\theta_{j}^*, \phi(s_{h_{2}}, a_{h_{2}})\right)\bigg) \varepsilon_{i,h_1} \varepsilon_{j,h_2}\right] \in \mathbb{R}^{d \times d}
\end{equation*}
where $\varepsilon_{j,h}$ is defined in Theorem \ref{thm_normality}. We take the conditional expectation to find that the cross terms vanish. Therefore the expression of $\Omega_{ij}$ can be simplified as
\begin{equation}
    \Omega_{i,j} = \mathbb{E} \left[\frac{1}{H} \sum_{h = 1}^H  \bigg(\nabla_{\theta} f \left(\theta_i^*,\phi(s_{h},a_{h})\right) \bigg)^{\top} \bigg(\nabla_{\theta} f \left(\theta_j^*,\phi(s_{h},a_{h})\right) \bigg) \varepsilon_{i,h} \varepsilon_{j,h}\right] \in \mathbb{R}^{d \times d}.
\end{equation} 
Therefore, we have the following convergence(the second one is conditional on the original data).
\begin{align}\label{convergence_theta}
    \sqrt{K}\left(\widehat{\theta}_1 - \theta_1\right) & \stackrel{d}{\longrightarrow} N\left(0,\frac{1}{H} \sum_{h_1=1}^H \sum_{h_2 = 1}^H \left(\prod_{i=1}^{h_1-1} \Sigma_i^{-1} A_i \right)\Sigma^{-1}_{h_1} \Omega_{h_1,h_2} \Sigma^{-1}_{h_2} \left(\prod_{j=1}^{h_2-1} \Sigma_j^{-1} A_j \right)^{\top} \right); \notag \\
    \sqrt{K}\left(\widehat{\theta}_1^{\circ} - \widehat{\theta}_1\right) & \stackrel{d}{\longrightarrow} N\left(0,k_0 \frac{1}{H} \sum_{h_1=1}^H \sum_{h_2 = 1}^H \left(\prod_{i=1}^{h_1-1} \Sigma_i^{-1} A_i \right)\Sigma^{-1}_{h_1} \Omega_{h_1,h_2} \Sigma^{-1}_{h_2} \left(\prod_{j=1}^{h_2-1} \Sigma_j^{-1} A_j \right)^{\top} \right).
\end{align}
where $k_0$ is defined in Theorem \ref{thm_consistency}. Finally we use multivariate delta method and eventually get
\begin{equation}\label{normal_result}
    \sqrt{K} \left(\widehat{v}_{\pi} - v_{\pi}\right) \stackrel{d}{\longrightarrow} N(0,\sigma^2);
\end{equation}
and conditional on $\boldsymbol{\tau}_1,...,\boldsymbol{\tau}_K,$
\begin{equation*}
    \sqrt{K} \left(\widehat{v}_{\pi}^{\circ} - \widehat{v}_{\pi}\right) \stackrel{d}{\longrightarrow} N(0,k_0 \sigma^2);
\end{equation*}
where
\begin{align*}
    \sigma^2 
    :&= \sum_{h_1=1}^H \sum_{h_2 = 1}^H \left(\int_{\mathcal{S} \times \mathcal{A}} \nabla_{\theta} f(\theta_1^*,\phi(s,a)) \pi(a \mid s) \xi(s) d a d s\right) \\
    & \hspace{10ex} \cdot \left(\prod_{i=1}^{h_1-1} \Sigma_i^{-1} A_i \right)\Sigma^{-1}_{h_1} \Omega_{h_1,h_2} \Sigma^{-1}_{h_2} \left(\prod_{j=1}^{h_2-1} \Sigma_j^{-1} A_j \right)^{\top} \left(\int_{\mathcal{S} \times \mathcal{A}} \nabla_{\theta} f(\theta_1^*,\phi(s,a)) \pi(a \mid s) \xi(s) d a d s\right)^{\top}.
\end{align*}
To simplify expressions for $\Sigma_h$ and $A_h$ we define
\begin{equation*}
    F(\alpha,\beta) := \frac{1}{2}\nabla_{\alpha} \mathbb{E}_{\bar{\mu}} \left\{\left[ f(\alpha,\phi(s,a)) - r(s,a) + \int_{\mathcal{A}} f(\beta,\phi(s^{\prime},a^{\prime})) \pi(a^{\prime}\mid s^{\prime}) d a^{\prime}\right]^2\right\}
\end{equation*}
for $\alpha,\beta \in \Theta.$ We notice that
\begin{equation*}
    \Sigma_h = \left.\nabla_{\alpha}^{\top} F \right|_{\alpha = \theta_h^*, \beta = \theta_{h+1}^*} \text{ and } A_h = \left.\nabla_{\beta}^{\top} F \right|_{\alpha = \theta_h^*, \beta = \theta_{h+1}^*}.
\end{equation*}

Here we define another operator $\mathbb{T}$ to simplify the expression for $\sigma^2.$ Policy completeness under parameterized $\mathcal{F}$ can induce a mapping on $\Theta$, which is defined as
\begin{equation}\label{definition_T}
f_{\mathbb{T} \theta}(s, a)=\left(\mathcal{P} f_{\theta}\right)(s, a) ?
\end{equation}
That is, $\mathbb{T} \theta$ is the corresponding parameter of $\mathcal{P} f$. Note that $F(\theta_h^*,\theta_{h+1}^*) = 0$ and $F(\cdot,\cdot)$ is continuously differentiable in $\Theta \times \Theta.$ Since the Jacobian matrix of $Z(\theta)$ at $\theta^*$ is non-singular and is partitioned diagonal matrix with $\Sigma_h$ on its diagonal, we have $\Sigma_h$ is non-singular for $h \in [H].$ Therefore, we apply the theorem of implicit function and get that, for any $h \in [H],$ there exists unique vector-valued function $G_h$ defined on a neighbourhood of $\theta_{h+1}^*$ inside $\Theta$ such that $F(G_h(\theta_{h+1}^*),\theta_{h+1}^*) = 0.$ By definition of the operator $\mathbb{T},$ without loss of generality we let each $G_h$ be a confinement of $\mathbb{T}.$ That is, $\mathbb{T}$ is actually the operator defined on every neighbourhood of $\theta^*_h$ which satisfies \eqref{definition_T}. We denote the Jacobian matrix of $\mathbb{T}$ at $\theta$ as $D\mathbb{T}(\theta).$ From theorem of implicit functions we know $\mathbb{T}$ has continuous Jacobian matrix on neighborhood of any $\theta_h^*.$ If we use $D\mathbb{T}(\cdot)$ to denote its Jacobian matrix, then
\begin{equation*}
    D\mathbb{T}(\theta_{h+1}^*) = \Sigma_h^{-1} A_h.
\end{equation*}
Therefore,
\begin{align*}
    &\left(\int_{\mathcal{S} \times \mathcal{A}} \nabla_{\theta_1^*} f(\theta_1^*,\phi(s,a)) \pi(a \mid s) \xi(s) d a d s\right) \left(\prod_{i=1}^{h-1} \Sigma_i^{-1} A_i \right)\\
    = & \left(\int_{\mathcal{S} \times \mathcal{A}} \nabla_{\theta_1^*} f(\theta_1^*,\phi(s,a)) \pi(a \mid s) \xi(s) d a d s\right) \left(\prod_{i=1}^{h-1} D\mathbb{T}(\theta_{i+1})\right) \\
    = & \mathbb{E}\left[\nabla_{\theta_h^*} f\left(\mathbb{T}^{h-1}\left(\theta_{h}^*\right), \phi(s_1, a_1)\right) \bigg| s_1 \sim \xi(\cdot), a_1 \sim \pi(\cdot \mid s_1)\right].
\end{align*}
This expectation is actually dependent on transition probability and independent of policy, hence we can transform it to expectation over population generated by target policy $\pi.$
\begin{align}
    &\left(\int_{\mathcal{S} \times \mathcal{A}} \nabla_{\theta_1^*} f(\theta_1^*,\phi(s,a)) \pi(a \mid s) \xi(s) d a d s\right) \left(\prod_{i=1}^{h-1} \Sigma_i^{-1} A_i \right) \notag \\
     = & \mathbb{E}^{\pi} \left[\nabla_{\theta_h^*} f\left(\mathbb{T}^{h-1}\left(\theta_{h}^*\right), \phi(s_1, a_1)\right) \bigg| s_1 \sim \xi(\cdot), a_1 \sim \pi(\cdot \mid s_1)\right] \notag \\
    = & \mathbb{E}^{\pi}\left[\nabla_{\theta_h^*} \int_{\mathcal{S} \times \mathcal{A}} f(\mathbb{T}^{h-2}\left(\theta_{h}^*\right),\phi(s_2,a_2)) \pi(a_2 \mid s_2) p(s_2 \mid s_1,a_1) d a_2 d s_1 \bigg| s_1 \sim \xi(\cdot), a_1 \sim \pi(\cdot \mid s_1)\right] \notag \\
    = & \mathbb{E}^{\pi} \left[ \nabla_{\theta_h^*} f(\mathbb{\mathbb{T}}^{h-2}\left(\theta_{h}^*\right),\phi(s_2,a_2)) \bigg| s_1 \sim \xi(\cdot), a_1 \sim \pi(\cdot \mid s_1)\right] \notag \\
    = &\dots \notag \\
    = & \mathbb{E}^{\pi} \left[ \nabla_{\theta_h^*} f\left(\theta_{h}^*,\phi(s_h,a_h)\right) \bigg| s_1 \sim \xi(\cdot), a_1 \sim \pi(\cdot \mid s_1)\right] = \nu_h^{\top}. \label{derivation_nu}
\end{align}
The last equation is due to the definition of $\nu_h^{\top}$ in Theorem \ref{thm_normality}. Then we have
\begin{equation}
    \sigma^2 = \frac{1}{H} \sum_{h_1,h_2 = 1}^H  \nu_{h_1}^{\top} \Sigma^{-1}_{h_1} \Omega_{h_1,h_2} \Sigma^{-1}_{h_2} \nu_{h_2}.
\end{equation}
This finishes our proof for asymptotic normality.
\subsection{Proof of Corollary \ref{coro_prohorov}}
This corollary can be derived directly from the following lemma.
    \begin{lemma}[Prohorov's Theorem, Theorem 2.4 in \cite{van}]
       Let $X_n$ be random variables in $\mathbb{R}^k$ and $X_n \stackrel{d}{\longrightarrow} X$ for some $X,$ then $\left\{X_n : n \in \mathbb{N}\right\}$ is uniformly tight, i.e. for every $\delta > 0,$ there exists a constant $B,$ such that
       $$
       \sup_{n \in \mathbb{N}} \mathbb{P} \left(\left\Vert X_n \right\Vert_2 > B \right) < \delta.
       $$
    \end{lemma}
    
\subsection{Proof of Corollary \ref{coro_CI}}
We define $\Phi(t) := \mathbb{P}\left(N(0,k_0 \sigma^2) \leq t\right),$ where $\sigma^2$ is defined as \eqref{variance_expression} and $N(0,\sigma^2)$ represents a random variable which follows this distribution. From the asymptotic normality and distributional consistency, we have when $K \to \infty,$
\begin{equation*}
    \mathbb{P}_{\mathcal{D}} \left(\sqrt{K}\left(\widehat{v}_{\pi} - v_{\pi}\right) \leq t \right) \to \mathbb{P}\left(N(0,k_0\sigma^2) \leq \sqrt{k_0} t\right) = \Phi(\sqrt{k_0} t), \quad \mathbb{P}_{W^\circ \mid \mathcal{D}} \left(\sqrt{K}\left(\widehat{v}^\circ_{\pi} - \widehat{v}_{\pi}\right) \leq t\right) \to \Phi\left(t\right).
\end{equation*}
According to lemma 21.2 in \citep{van}, denote the cumulative distribution functions of $\sqrt{K}\left(\widehat{v}_{\pi}^\circ - \widehat{v}_{\pi}\right)$ as $\Phi_K,$ then $\Phi_K \stackrel{d}{\longrightarrow} \Phi$ implies $\Phi_K^{-1} \stackrel{d}{\longrightarrow} \Phi^{-1}.$ Therefore, $\sqrt{K} q_{\delta}^{\pi} = \Phi_K^{-1}\left(\delta\right) \to \Phi^{-1}\left(\delta\right).$ Therefore,
\begin{equation*}
    \mathbb{P}_{\mathcal{D}W^\circ}\left(\widehat{v}_{\pi} - v_{\pi} \geq \frac{1}{\sqrt{k_0}} q_{\frac{\delta}{2}}^{\pi}\right) = \mathbb{P}_{\mathcal{D}W^\circ}\left(\sqrt{K}\left(\widehat{v}_{\pi} - v_{\pi}\right) \geq \sqrt{\frac{K}{k_0}} q_{\frac{\delta}{2}}^{\pi}\right) \to \mathbb{P}_{\mathcal{D}W^\circ}\left(N\left(0,\sigma^2\right) \geq \frac{1}{\sqrt{k_0}}\Phi^{-1}\left(\frac{\delta}{2}\right)\right) = 1 - \frac{\delta}{2}.
\end{equation*}
We can bound $\mathbb{P}_{\mathcal{D}W^\circ} \left( \widehat{v}_{\pi} - v_{\pi} \leq - \frac{1}{\sqrt{k_0}} q_{1 - \frac{\delta}{2}}^{\pi}\right)$ similarly, and by an argument of union bound we can get the result for asymptotic confidence interval.

\section{Proof of Theorem \ref{thm_finite_sample_square}}
We are going to decompose FQE error $\widehat{v}_{\pi} - v_{\pi}$ into first order term and higher order term. Since we let $\lambda = 0,$ we do not have bias term containing $\lambda.$ Since we have
\begin{equation*}
    \widehat{v}_{\pi} - v_{\pi} = \int_{\mathcal{S} \times \mathcal{A}} \Bigg[f(\widehat{\theta}_1,\phi(s,a)) - f(\theta_1^*,\phi(s,a))\Bigg]\pi(a \mid s)\xi(s) d a d s,
\end{equation*}
the first order term will be
$$
\int_{\mathcal{S} \times \mathcal{A}} \left[\nabla_{\theta} f(\theta_1^*,\phi(s,a)) \cdot \left(\widehat{\theta}_1 - \theta_1^*\right)\right] \pi(a \mid s)\xi(s) d a d s.
$$
Next we expand $\widehat{\theta}_1 - \theta_1.$ From multivariate Taylor expansion we have 
\begin{equation}\label{finite_Taylor}
    \sqrt{K} \left[Z(\widehat{\theta}_K) - Z(\theta^*)\right] = \sqrt{K} H \Sigma_* (\widehat{\theta}_K - \theta^*) + \sqrt{K} R_K.
\end{equation}
where $R_K$ is Taylor remainder and $\Sigma_*$ is defined in \eqref{expression_Sigma}. Notice that $Z(\theta^*) = Z_K(\widehat{\theta}_K) = 0,$ we have
\begin{align}
    \sqrt{K}\left(\widehat{\theta}_K - \theta^*\right)
    &= \frac{\sqrt{K}}{H} \Sigma^{-1}_* \left(Z(\widehat{\theta}_K) - Z(\theta^*)\right) - \frac{\sqrt{K}}{H} \Sigma^{-1}_* R_K \notag \\
    &= -\frac{1}{\sqrt{K}H} \Sigma^{-1}_* \sum_{k=1}^K z\left(\theta^*,\boldsymbol{\tau}_k\right) + \frac{\Sigma^{-1}_*}{H} \left(\sqrt{K}\mathbb{E}_{\boldsymbol{\tau}} z(\widehat{\theta}_K, \boldsymbol{\tau}) + \frac{1}{\sqrt{K}} \sum_{k=1}^K z(\theta^*,\boldsymbol{\tau}_k) \right) - \frac{\sqrt{K}}{H} \Sigma^{-1}_* R_K. \label{decomposition}
\end{align}
Since the true value function is explicitly dependent only on $\theta_1^*,$ we take the first $d$ entries of this decomposition. We denote $R_{K,h}$ as the vector comprising $(h-1)d + i$-th to $hd$-th entry of $R_K,$ hence
\begin{align*}
    \widehat{\theta}_1 - \theta_1^* 
    & = -\frac{1}{KH} \sum_{h=1}^H \left(\prod_{i=1}^{h-1} \Sigma_{i}^{-1} A_i\right) \Sigma^{-1}_h \left(\sum_{k=1}^K z_{(h)}(\theta^*,\boldsymbol{\tau}_k)\right) \\
    & + \frac{1}{H}\sum_{h=1}^H \left(\prod_{i=1}^{h-1} \Sigma_{i}^{-1} A_i\right) \Sigma^{-1}_h \left(\mathbb{E}_{\boldsymbol{\tau}} z_{(h)}(\widehat{\theta}_K,\boldsymbol{\tau}) + \frac{1}{K} \sum_{k=1}^K z_{(h)}(\theta^*,\boldsymbol{\tau}_k)\right) - \frac{1}{H} \sum_{h=1}^H \left(\prod_{i=1}^{h-1} \Sigma_{i}^{-1} A_i\right) \Sigma^{-1}_h R_{K,h}.
\end{align*}
Therefore, the first order term of total error is
\begin{equation}
    I_1 := - \frac{1}{KH} \int_{\mathcal{S} \times \mathcal{A}} \left[\nabla_{\theta} f(\theta_1^*,\phi(s,a)) \cdot \sum_{h=1}^H \left(\prod_{i=1}^{h-1} \Sigma_{i}^{-1} A_i\right) \Sigma^{-1}_h \left(\sum_{k=1}^K z_{(h)}(\theta^*,\boldsymbol{\tau}_k)\right) \right] \pi(a \mid s)\xi(s) d a d s,
\end{equation}
and the remaining part will be the higher order term.
\begin{equation}
    I_2 := \widehat{v}_{\pi} - v_{\pi} - I_1.
\end{equation}

\subsection{First Order Term}\label{sec:pf_first_order}
The only difference between Variance aware error bound \eqref{finite_bound2} and Reward-free error bound \eqref{finite_bound1} is the contraction technique for the first order term. We deal with them separately.

\subsubsection{Variance Aware Error Bound}
We first bound $I_1$ using Bernstein's Inequality (lemma \ref{ine_bernstein}). From this upper bound we see the dependency of its dominant term on asymptotic variance $\sigma^2$ in Theorem \ref{thm_normality}. By the same argument of \eqref{derivation_nu}, we have
\begin{equation*}
    \left(\int_{\mathcal{S} \times \mathcal{A}} \nabla_{\theta_{1}^{*}} f\left(\theta_{1}^{*}, \phi(s, a)\right) \pi(a \mid s) \xi(s) d a d s\right)\left(\prod_{i=1}^{h-1} \Sigma_{i}^{-1} A_{i}\right) = \nu_h^{\top}.
\end{equation*}
We define
\begin{equation*}
    e_k := - \frac{1}{H} \int_{\mathcal{S} \times \mathcal{A}} \left[\nabla_{\theta} f(\theta_1,\phi(s,a)) \cdot \sum_{h=1}^H \left(\prod_{i=1}^{h-1} \Sigma_{i}^{-1} A_i\right) \Sigma^{-1}_h \cdot z_{(h)}(\theta^*,\boldsymbol{\tau}_k) \right] \pi(a \mid s)\xi(s) d a d s.
\end{equation*}
Then by the derivation in \eqref{derivation_nu}, we have
\begin{align*}
    e_k 
    &= - \frac{1}{H} \sum_{h=1}^H \nu_h^{\top} \Sigma_h^{-1} z_{(h)}(\theta^*,\boldsymbol{\tau}_k) \\
    &= - \frac{1}{H} \sum_{h=1}^H \sum_{j=1}^H \nu_h^{\top} \Sigma_h^{-1} \bigg[\nabla_{\theta}^{\top} f(\theta_h^*,\phi(s_j^k,a_j^k))\bigg] \varepsilon_{h,j}^k;
\end{align*}
The first order term will be independent sum of $e_k,$ 
$$
I_1 = \frac{1}{K} \sum_{k=1}^K e_k.
$$
From assumption \ref{assumption_data_coverage} and because of $\left|\varepsilon_{h,j}^k\right| \leq H - h + 1,$ we can bound $e_k$ by
\begin{align*}
    \left|e_k\right|
    & \leq \frac{1}{H} \sum_{h=1}^H \sum_{j=1}^H  \left\|\nu_h^{\top} \Sigma_h^{-\frac{1}{2}}\right\|_2 \left\|\Sigma_h^{-\frac{1}{2}} \cdot \nabla_{\theta}^{\top} f(\theta_h,\phi(s_j^k,a_j^k)) \right\|_2 \left|\varepsilon_{h,j}^k \right| \\
    & \leq \sqrt{C_2 d} \sum_{h=1}^H (H-h+1) \sqrt{\nu_h^{\top} \Sigma_h^{-1} \nu_h}.
\end{align*}
Then we calculate the variance for $e_k.$ By definition of $\theta_h^*,$ we have
\begin{equation*}
    \mathbb{E} \left[\bigg[\nabla_{\theta}^{\top} f(\theta_h^*,\phi(s_j^k,a_j^k))\bigg] \varepsilon_{h,j}^k\right] = 0,
\end{equation*}
hence $\mathbb{E}\left[e_k\right] = 0$ and $\operatorname{Var}\left[e_k\right] = \mathbb{E}\left[e_k^2\right]$ for $k \in [K].$ We have
\begin{align*}
    \mathbb{E}\left[e_k^2\right]
    & = \frac{1}{H^2} \sum_{h_1 = 1}^H \sum_{h_2 = 1}^H \sum_{j = 1}^H \sum_{l = 1}^H \left[\nu_{h_1}^{\top} \Sigma_{h_1}^{-1} \mathbb{E} \left[\bigg(\nabla_{\theta}^{\top} f(\theta_{h_1}^*,\phi(s_j^k,a_j^k))\bigg) \bigg(\nabla_{\theta} f(\theta_{h_2}^*,\phi(s_l^k,a_l^k))\bigg) \varepsilon_{h_1,j}^k\varepsilon_{h_2,l}^k\right] \Sigma_{h_2}^{-1} \nu_{h_2}\right].
\end{align*}
By conditional expectation, when $j < l,$ we have
\begin{align*}
    & \mathbb{E} \left[\bigg(\nabla_{\theta}^{\top} f(\theta_{h_1}^*,\phi(s_j^k,a_j^k))\bigg) \bigg(\nabla_{\theta} f(\theta_{h_2}^*,\phi(s_l^k,a_l^k))\bigg) \varepsilon_{h_1,j}^k\varepsilon_{h_2,l}^k\right] \\
    =& \mathbb{E} \left\{ \mathbb{E} \left[\bigg(\nabla_{\theta}^{\top} f(\theta_{h_1}^*,\phi(s_j^k,a_j^k))\bigg) \bigg(\nabla_{\theta} f(\theta_{h_2}^*,\phi(s_l^k,a_l^k))\bigg) \varepsilon_{h_1,j}^k\varepsilon_{h_2,l}^k \bigg| s_j^k,a_j^k\right]\right\} = 0,
\end{align*}
hence all cross term vanish and 
\begin{align*}
    \mathbb{E}\left[e_k^2\right]
    & = \frac{1}{H^2} \sum_{h_1 = 1}^H \sum_{h_2 = 1}^H \sum_{j = 1}^H \left[\nu_{h_1}^{\top} \Sigma_{h_1}^{-1} \mathbb{E} \left[\bigg(\nabla_{\theta}^{\top} f(\theta_{h_1}^*,\phi(s_j^k,a_j^k))\bigg) \bigg(\nabla_{\theta} f(\theta_{h_2}^*,\phi(s_j^k,a_j^k))\bigg) \varepsilon_{h_1,j}^k\varepsilon_{h_2,j}^k\right] \Sigma_{h_2}^{-1} \nu_{h_2}\right] \\
    & = \frac{1}{H} \sum_{h_1 = 1}^H \sum_{h_2 = 1}^H \sum_{j = 1}^H \left[\nu_{h_1}^{\top} \Sigma_{h_1}^{-1} \Omega_{h_1,h_2} \Sigma_{h_2}^{-1} \nu_{h_2}\right] = \sigma^2,
\end{align*}
where $\Omega_{h_1,h_2}$ and $\sigma^2$ are defined as in Theorem \ref{thm_normality}. We use lemma \ref{ine_bernstein}, then we have for any $\varepsilon > 0,$
\begin{equation*}
\mathbb{P}\left(\left\|\sum_{k=1}^K e_k\right\| \geq \varepsilon\right) \leq 2 \exp\left(- \frac{\varepsilon^2/2}{K \sigma^2 + \varepsilon \sqrt{C_2 d} \sum_{h=1}^H (H-h+1) \sqrt{\nu_h^{\top} \Sigma_h^{-1} \nu_h}/ 3 } \right)
\end{equation*}
We choose
$$
\varepsilon := \ln\left(\frac{2}{\delta}\right) \left[\sqrt{2K} \sigma + \frac{2}{3} \sqrt{C_2 d} \sum_{h=1}^H (H-h+1) \sqrt{\nu_h^{\top} \Sigma_h^{-1} \nu_h}\right],
$$
then RHS is bounded by $\delta > 0.$ Therefore, with probability at least $1 - \delta,$
$$
\left|I_1\right| \leq \ln\left(\frac{2}{\delta}\right) \left[\sqrt{\frac{2\sigma^2}{K}}+ \frac{2}{3K} \sqrt{C_2 d} \sum_{h=1}^H (H-h+1) \sqrt{\nu_h^{\top} \Sigma_h^{-1} \nu_h}\right].
$$
This gives one upper bound for the first order term. Notice that this bound depends on asymptotic variance $\sigma^2.$ When we apply this bound into final upper bound, we get variance-aware error bound \eqref{finite_bound2}.

\subsubsection{Reward-Free Error Bound}
Next, we are going to replace this dependence on $\sigma^2$ with a dependence on a reward-free $\chi^2$ divergence, with which we give an upper bound for the worst instance of reward. The main difference is that we use another decomposition of the first order term and use Freedman's Inequality as our central tool. We can simplify the expression for $I_1$ as
\begin{equation*}
    I_1 = -\frac{1}{KH} \sum_{k=1}^K \sum_{h=1}^{H} \sum_{j=1}^{H} \nu_{h}^{\top} \Sigma_{h}^{-1}\left[\nabla_{\theta}^{\top} f\left(\theta_{h}^*, \phi\left(s_{j}^{k}, a_{j}^{k}\right)\right)\right] \varepsilon_{h, j}^{k},
\end{equation*}
where
\begin{equation*}
    \varepsilon_{h,j}^k = f\left(\theta_h^*, \phi\left(s_{j}^k, a_{j}^k\right)\right) -  r\left(s_{j}^k, a_{j}^k\right) - \int_{\mathcal{A}} f(\theta_{h+1}^*,\phi(s_{j+1}^k,a^{\prime})) \pi(a^{\prime}\mid s_{j+1}^k) d a^{\prime}.
\end{equation*}
Below we can decompose $I_1$ into $N= KH$ items.We denote $\varepsilon_{h,n} = \varepsilon_{h,j}^k$ and $(s_n,a_n) = (s_j^k,a_j^k)$ if $n = (h-1)K + k.$ Then if we define
\begin{equation*}
    u_n := -\sum_{h=1}^H \nu_{h}^{\top} \Sigma_{h}^{-1}\left[\nabla_{\theta}^{\top} f\left(\theta_{h}^*, \phi\left(s_n, a_n\right)\right)\right] \varepsilon_{h, n},
\end{equation*}
then
\begin{equation*}
    I_1 = \frac{1}{N} \sum_{n=1}^N u_n.
\end{equation*}
Define $\mathcal{F}_n$ as the $\sigma$ field generated by $s_1,a_1,....,s_n,a_n,$ then $\left\{\mathcal{F}_n\right\}_{n=1}^N$ is a filtration. Since 
\begin{align*}
    & \mathbb{E}\left[\nabla_{\theta}^{\top} f\left(\theta_{h}^{*}, \phi\left(s_{n}, a_{n}\right)\right) \left[f\left(\theta_{h}^{*}, \phi\left(s_n, a_n\right)\right)-r\left(s_n, a_n\right)-\int_{\mathcal{A}} f\left(\theta_{h+1}^{*}, \phi\left(s_{n+1}, a^{\prime}\right)\right) \pi\left(a^{\prime} \mid s_{n+1}\right) d a^{\prime}\right] \bigg| s_n,a_n\right]\\
    =&  \nabla_{\theta}^{\top} \mathbb{E}\left[\left[f\left(\theta, \phi\left(s_n, a_n\right)\right)-r\left(s_n, a_n\right)-\int_{\mathcal{A}} f\left(\theta_{h+1}^{*}, \phi\left(s_{n+1}, a^{\prime}\right)\right) \pi\left(a^{\prime} \mid s_{n+1}\right) d a^{\prime}\right]^2 \bigg| s_n,a_n\right]\bigg|_{\theta = \theta_h^*} = 0,
\end{align*}
we have $\mathbb{E}\left[u_n \mid \mathcal{F}_n\right] = 0,$ and $\left\{u_n\right\}_{n=1}^N$ is a martingale difference sequence. We use Freedman's Inequality (lemma \ref{ine_freedman}) to analyze $I_1.$ Under data coverage assumption (Assumption \ref{assumption_data_coverage}), we have
\begin{align*}
    \left|u_n\right| 
    & \leq \sum_{h=1}^H \left\|\nu_h^{\top} \Sigma_h^{-\frac{1}{2}}\right\| \left\|\Sigma_h^{-\frac{1}{2}} \nabla_{\theta}^{\top} f\left(\theta_{h}^*, \phi\left(s_{j}^{k}, a_{j}^{k}\right)\right)\right\| \left|\varepsilon_{h,n}\right|\\
    & \leq \sqrt{C_{2} d} \sum_{h=1}^{H}(H-h+1) \sqrt{\nu_{h}^{\top} \Sigma_{h}^{-1} \nu_{h}}.
\end{align*}
Next we estimate the conditional variance $\operatorname{Var}\left[u_n\mid \mathcal{F}_n\right].$
\begin{align*}
    \operatorname{Var}\left[u_n\mid \mathcal{F}_n\right]
    &= \mathbb{E}\left[\left(\sum_{h=1}^H \nu_{h}^{\top} \Sigma_{h}^{-1}\left[\nabla_{\theta}^{\top} f\left(\theta_{h}^*, \phi\left(s_n, a_n\right)\right)\right] \varepsilon_{h, n}\right)^2 \bigg| \mathcal{F}_n\right] \\
    & \leq \left(\sum_{h=1}^H \frac{\sqrt{\nu_h^{\top}\Sigma_h^{-1} \nu_h}}{H-h+1} \operatorname{Var}\left(r\left(s_{n}, a_{n}\right)+\int_{\mathcal{A}} f\left(\theta_{h}^*, \phi\left(s_{n+1}, a^{\prime}\right)\right) \pi\left(a^{\prime} \mid s_{n+1}\right) d a^{\prime} \bigg| \mathcal{F}_{n}\right)\right)\\
    & \hspace{8ex} \cdot \left(\sum_{h=1}^H \frac{H-h+1}{\sqrt{\nu_h^{\top}\Sigma_h^{-1} \nu_h}} \left(\nu_{h}^{\top} \Sigma_{h}^{-1}\left[\nabla_{\theta}^{\top} f\left(\theta_{h}^*, \phi\left(s_n, a_n\right)\right)\right]\right)^2\right) \tag{Cauchy-Schwarz}
\end{align*}
Since $r\left(s_{n}, a_{n}\right)+\int_{\mathcal{A}} f\left(\theta_{h}^*, \phi\left(s_{n+1}, a^{\prime}\right) \right)\pi\left(a^{\prime} \mid s_{n+1}\right) d a^{\prime} \in [0,H-h+1],$ we have
$$
\operatorname{Var}\left(r\left(s_{n}, a_{n}\right)+\int_{\mathcal{A}} f\left(\theta_{h}^*, \phi\left(s_{n+1}, a^{\prime}\right)\right) \pi\left(a^{\prime} \mid s_{n+1}\right) d a^{\prime} \right) \leq \frac{1}{4}(H-h+1)^2,
$$
then
\begin{align*}
    \operatorname{Var}\left[u_n\mid \mathcal{F}_n\right]
    & \leq \frac{1}{4} \left(\sum_{h=1}^H (H-h+1)\sqrt{\nu_{h}^{\top} \Sigma_{h}^{-1}\nu_{h}}\right)\cdot \left(\sum_{h=1}^H \frac{H-h+1}{\sqrt{\nu_h^{\top}\Sigma_h^{-1} \nu_h}} \left(\nu_{h}^{\top} \Sigma_{h}^{-1}\left[\nabla_{\theta}^{\top} f\left(\theta_{h}^*, \phi\left(s_n, a_n\right)\right)\right]\right)^2\right),
\end{align*}
and
\begin{align*}
    & \sum_{n=1}^N \operatorname{Var}\left[u_n\mid \mathcal{F}_n\right] \\
    \leq &\frac{N}{4} \left(\sum_{h=1}^H (H-h+1)\sqrt{\nu_{h}^{\top} \Sigma_{h}^{-1}\nu_{h}}\right) \left(\sum_{h=1}^H \frac{H-h+1}{\sqrt{\nu_h^{\top}\Sigma_h^{-1} \nu_h}} 
    \nu_{h}^{\top} \Sigma_{h}^{-1} \left[\frac{1}{N}\sum_{n=1}^N \nabla_{\theta}^{\top} f\left(\theta_{h}^*, \phi\left(s_n, a_n\right)\right) \nabla_{\theta} f\left(\theta_{h}^*, \phi\left(s_n, a_n\right)\right)\right] \Sigma_{h}^{-1} \nu_{h}\right) \\
    \leq & \frac{N}{4} \left(\sum_{h=1}^H (H-h+1) \sqrt{\nu_{h}^{\top} \Sigma_{h}^{-1}\nu_{h}}\right) \left(\sum_{h=1}^H (H-h+1) \sqrt{\nu_h^{\top}\Sigma_h^{-1} \nu_h} \left\|\Sigma_{h}^{-\frac{1}{2}} \left[\frac{1}{N}\sum_{n=1}^N \nabla_{\theta}^{\top} f\left(\theta_{h}^*, \phi\left(s_n, a_n\right)\right) \nabla_{\theta} f\left(\theta_{h}^*, \phi\left(s_n, a_n\right)\right)\right] \Sigma_{h}^{-\frac{1}{2}} \right\| \right).
\end{align*}
where we use $\mu^{\top} \Sigma^{-1} X \Sigma^{-1} \mu \leq \mu^{\top} \Sigma^{-1} \mu \cdot\left\|\Sigma^{-1 / 2} X \Sigma^{-1 / 2}\right\|_{2} \text { for any } \mu \in \mathbb{R}^{d}, X \in \mathbb{R}^{d \times d}.$ We use a special case of lemma \ref{lemma_first_order} when $h_1 = h_2 = h$ and we have with probability at least $1-\frac{\delta}{2H},$
\begin{equation*}
    \left\|\Sigma_{h}^{-\frac{1}{2}} \left[\frac{1}{N}\sum_{n=1}^N \nabla_{\theta}^{\top} f\left(\theta_{h}^*, \phi\left(s_n, a_n\right)\right) \nabla_{\theta} f\left(\theta_{h}^*, \phi\left(s_n, a_n\right)\right)\right] \Sigma_{h}^{-\frac{1}{2}} \right\| \leq 1+\sqrt{\frac{2 C_{2} d}{K} \log \left(\frac{4 d H}{\delta}\right)}+\frac{2\left(C_{2} d+1\right)}{3 K} \log \left(\frac{4 d H}{\delta}\right).
\end{equation*}
By union bound, we have with probability at least $1 - \frac{\delta}{2},$ for all $h \in [H],$ the bounds above hold simultaneously and hence
\begin{equation*}
    \sum_{n=1}^N \operatorname{Var}\left[u_n\mid \mathcal{F}_n\right] \leq \frac{N}{4} \left(\sum_{h=1}^H (H-h+1) \sqrt{\nu_{h}^{\top} \Sigma_{h}^{-1}\nu_{h}}\right)^2 \left(1+\sqrt{\frac{2 C_{2} d}{K} \log \left(\frac{4 d H}{\delta}\right)}+\frac{2\left(C_{2} d+1\right)}{3 K} \log \left(\frac{4 d H}{\delta}\right)\right).
\end{equation*}
We take $\sigma^2_0$ equals the right hand side of inequality above, then
\begin{equation*}
    \mathbb{P}\left(\sum_{n=1}^{N} \operatorname{Var}\left[u_{n} \mid \mathcal{F}_{n}\right] \geq \sigma^{2}_0\right) \leq \frac{\delta}{2}.
\end{equation*}
Freedman's Inequality implies that for any $\varepsilon>0$, we have
$$
\mathbb{P}\left(\left|\sum_{n=1}^{N} u_{n}\right| \geq \varepsilon, \sum_{n=1}^{N} \operatorname{Var}\left[u_{n} \mid \mathcal{F}_{n}\right] \leq \sigma^{2}_0\right) \leq 2 \exp \left(-\frac{\varepsilon^{2} / 2}{\sigma^{2}_0+\varepsilon \sqrt{C_{2} d} \sum_{h=1}^{H}(H-h+1) \sqrt{\nu_{h}^{\top} \Sigma_{h}^{-1} \nu_{h} / 3}}\right) .
$$
We take
$$
\varepsilon:=\sqrt{2 \log \left(\frac{4}{\delta}\right)} \sigma_0+\log \left(\frac{4}{\delta}\right) \frac{2 \sqrt{C_{2} d} \sum_{h=1}^{H}(H-h+1) \sqrt{\nu_{h}^{\top} \Sigma_{h}^{-1} \nu_{h}}}{3},
$$
then
$$
\mathbb{P}\left(\left|\sum_{n=1}^{N} u_{n}\right| \geq \varepsilon\right) \leq \mathbb{P}\left(\sum_{n=1}^{N} \operatorname{Var}\left[u_{n} \mid \mathcal{F}_{n}\right] \geq \sigma^{2}_0\right)+\mathbb{P}\left(\left|\sum_{n=1}^{N} u_{n}\right| \geq \varepsilon, \sum_{n=1}^{N} \operatorname{Var}\left[u_{n} \mid \mathcal{F}_{n}\right] \leq \sigma^{2}_0\right) \leq \frac{\delta}{2}+\frac{\delta}{2}=\delta
$$
Eventually we use $\sqrt{1+x} \leq 1+ \frac{x}{2}$ for any $x \geq 0$ to get with probability at least $1-\delta,$
\begin{align*}
    \left|I_1\right| \leq \sqrt{\frac{\log(4/\delta)}{2KH}} \left(\sum_{h=1}^H (H-h+1) \sqrt{\nu_{h}^{\top} \Sigma_{h}^{-1}\nu_{h}}\right) + \Delta I_1,
\end{align*}
where
\begin{align*}
    \Delta I_1 = \left(\sum_{h=1}^H (H-h+1) \sqrt{\nu_{h}^{\top} \Sigma_{h}^{-1}\nu_{h}}\right) \left[\frac{7}{6K}\sqrt{\frac{C_2 d}{H}} \log\left(\frac{4d H}{\delta}\right) + \frac{C_2d+1}{3 \sqrt{2H} K^{\frac{3}{2}}} \log\left(\frac{4dH}{\delta}\right)^{\frac{3}{2}}\right]
\end{align*}
We define the function classes
\begin{equation}
    \mathcal{G}_h = \left\{\left(\nabla_{\theta} f\left(\theta_{h}^*, \phi(s, a)\right)\right)\cdot \mu: \mu \in \mathbb{R}^d\right\}
\end{equation}
Notice that by Cauchy-Schwarz Inequality, we have
\begin{equation*}
    \sqrt{\nu_h^{\top} \Sigma_h^{-1} \nu_h} = \sup_{\mu \in \mathbb{R}^d} \frac{\mu^{\top} \nu_h}{\sqrt{\mu^{\top} \Sigma_h \mu}}.
\end{equation*}
For $\mu \in \mathbb{R}^d,$ we take $g \in \mathcal{G}_h$ such that $g(s,a) = \left(\nabla_{\theta} f\left(\theta_{h}^*, \phi(s, a)\right)\right)\cdot \mu$ for any $(s,a) \in \mathcal{S} \times \mathcal{A}.$ Then
$$
\mu^{\top} \nu_h = \mathbb{E}^{\pi}\left[g(s_h,a_h) \mid s_1 \sim \xi(\cdot)\right].
$$
Additionally, we have
\begin{equation*}
    \mu^{\top} \Sigma_h \mu = \mathbb{E}_{\boldsymbol{\tau}} \left\{\frac{1}{H} \sum_{j=1}^H \left[\bigg(\nabla_{\theta} f\left(\theta_{h}^*, \phi(s_{j}, a_{j})\right)\bigg) \cdot \mu\right]^2 \right\} = \mathbb{E}_{\boldsymbol{\tau}} \left\{\frac{1}{H} \sum_{h=1}^H g^2\left(s_h,a_h\right) \right\}
\end{equation*}
Then,
$$
 \sqrt{\nu_h^{\top} \Sigma_h^{-1} \nu_h} = \sup_{\mu \in \mathbb{R}^d} \frac{\mu^{\top} \nu_h}{\sqrt{\mu^{\top} \Sigma_h \mu}} = \sup_{g \in \mathcal{G}_h} \frac{\mathbb{E}^{\pi}\left[g\left(s_{h}, a_{h}\right) \mid s_{1} \sim \xi(\cdot)\right]}{\sqrt{\mathbb{E}\left[\frac{1}{H} \sum_{h=1}^{H} g^{2}\left(s_{h}, a_{h}\right)\right]}}.
$$
We substitute this supremum into upper bound and this concludes the proof for the first order term.

\subsection{Higher Order Term}\label{sec:higher_order}
The higheer order term comprises three parts: one part comes from higher order term of $f(\widehat{\theta}_1,\phi) - f(\theta_1^*,\phi),$ the other two parts come from higher order term and Taylor remainder of $\widehat{\theta}_1 - \theta_1^*.$ We denote respectively as
\begin{align*}
    I_{21} &:= \int_{\mathcal{S} \times \mathcal{A}}\left[f\left(\widehat{\theta}_{1}, \phi(s, a)\right)-f\left(\theta_{1}^*, \phi(s, a)\right) - \nabla_{\theta} f\left(\theta_{1}^*, \phi(s, a)\right) \cdot\left(\widehat{\theta}_{1}-\theta_{1}\right)\right] \pi(a \mid s) \xi(s) d a d s;\\
    I_{22} &:= \frac{1}{H} \int_{\mathcal{S} \times \mathcal{A}}\left[\nabla_{\theta} f\left(\theta_{1}^*, \phi(s, a)\right) \sum_{h=1}^{H}\left(\prod_{i=1}^{h-1} \Sigma_{i}^{-1} A_{i}\right) \Sigma_{h}^{-1}\left(\mathbb{E}_{\tau} z_{(h)}\left(\widehat{\theta}_{K}, \boldsymbol{\tau}\right)+\frac{1}{K} \sum_{k=1}^{K} z_{(h)}\left(\theta^{*}, \boldsymbol{\tau}_{k}\right)\right)\right] \pi(a \mid s) \xi(s) d a d s;\\
    I_{23} &:= -\frac{1}{H} \int_{\mathcal{S} \times \mathcal{A}} \left[\nabla_{\theta} f\left(\theta_{1}^*, \phi(s, a)\right) \sum_{h=1}^{H}\left(\prod_{i=1}^{h-1} \Sigma_{i}^{-1} A_{i}\right) \Sigma_{h}^{-1} R_{K, h}\right]\pi(a \mid s) \xi(s) d a d s.
\end{align*}
We bound them separately. We keep the notation in \eqref{notation1} and use $\alpha = (\alpha_1,\alpha_2,...,\alpha_d)$ as multi-index. If $x = (x_1,x_2,...,x_d)$ is a vector of same dimension, we denote $x^{\alpha} := x_1^{\alpha_1} x_2^{\alpha_2} ... x_d^{\alpha_d}.$ Denote $\Delta \theta_1^* := \widehat{\theta}_1 - \theta_1^*,$ from Taylor's Theorem, we have
\begin{align*}
    f\left(\widehat{\theta}_{1}, \phi(s, a)\right)-f\left(\theta_{1}^*, \phi(s, a)\right)-\nabla_{\theta} f\left(\theta_{1}^*, \phi(s, a)\right) \cdot\left(\widehat{\theta}_{1}-\theta_{1}^*\right) = \sum_{\left|\alpha\right| = 2} \frac{\left(\Delta \theta_1^* \right)^{\alpha}}{2}\partial^{\alpha}f(\theta_1^* + c\Delta \theta_1^*)
\end{align*}
for some $c \in (0,1).$ $\sum_{\left|\alpha\right| = 2}$ denotes summation over all second order derivatives. We recall the definition of $\kappa_l,  l = 1,2,3,$
\begin{equation*}
    \kappa_{l}:=\sup _{\theta \in \Theta, \phi \in \Psi|\alpha|=l} \sup \left|\partial_{\theta}^{\alpha} f(\theta, \phi)\right|<\infty,
\end{equation*}
where $\alpha=\left(\alpha_{1}, \alpha_{2}, \ldots, \alpha_{d}\right)$ is a multi-index with order $|\alpha|=\alpha_{1}+\alpha_{2}+\cdots+\alpha_{n},$ and $\partial^{\alpha}_{\theta} f(\theta,\phi) = \partial_{\theta_1}^{\alpha_{1}} \partial_{\theta_2}^{\alpha_{2}} \cdots \partial_{\theta_d}^{\alpha_{d}} f=\frac{\partial^{|\alpha|} f}{\partial \theta_{1}^{\alpha_{1}} \partial \theta_{2}^{\alpha_{2}} \cdots \partial \theta_{d}^{\alpha_{d}}}.$s Hence we have
\begin{equation*}
    \left| I_{21} \right| \leq \frac{\kappa_2}{2} \int_{\mathcal{S} \times \mathcal{A}} \left|\sum_{\left|\alpha\right| = 2} \left(\Delta \theta_1^* \right)^{\alpha}\right| \pi(a \mid s) \xi(s) d a d s \leq \frac{\kappa_2}{2} \left\|\Delta \theta_1^*\right\|_1^2 \leq \frac{\kappa_2 d}{2} \left\|\Delta \theta_1^*\right\|_2^2
\end{equation*}
From Corrollary \ref{coro_prohorov}, we have for every fixed $K,$ with probability at least $1 - \delta,$ the event $\mathcal{E}_K$ happens, where
$$
\mathcal{E}_K := \left\{\left\|\widehat{\theta}_K - \theta^*\right\|_2 \leq \frac{B(\delta)}{\sqrt{K}}\right\}.
$$
Under $\mathcal{E}_K,$ we have $\left\|\Delta \theta_1^*\right\|_2^2 \leq \frac{B(\delta)^2}{K},$ hence
$$\left|I_{21}\right| \leq \frac{\kappa_2 d B(\delta)^2}{2 K}.$$
Next we bound $I_{22}.$ From \eqref{derivation_nu} and definition of $\widehat{\theta}_K,$ we have
\begin{align*}
    I_{22}
    & = \frac{1}{H} \sum_{h=1}^H \nu_h^{\top} \Sigma_h^{-1} \left(\mathbb{E}_{\boldsymbol{\tau}} z_{(h)}\left(\widehat{\theta}_{K}, \boldsymbol{\tau}\right) - \mathbb{E}_{\boldsymbol{\tau}} z_{(h)}\left(\theta^*, \boldsymbol{\tau}\right) +\frac{1}{K} \sum_{k=1}^{K} z_{(h)}\left(\theta^{*}, \boldsymbol{\tau}_{k}\right) - \frac{1}{K} \sum_{k=1}^{K} z_{(h)}\left(\widehat{\theta}_K, \boldsymbol{\tau}_{k}\right) \right).
\end{align*}
We use $\mathbb{G}_K := \sqrt{K}\left(\mathbb{P}_K - \mathbb{P}\right)$ to denote empirical measure and we have
\begin{align*}
    I_{22}
    & = \frac{1}{H\sqrt{K}} \mathbb{G}_K \left[\sum_{h=1}^H \nu_h^{\top} \Sigma_h^{-1} \left(z_{(h)}\left(\widehat{\theta}_{K}, \boldsymbol{\tau}\right) - z_{(h)}\left(\theta^*, \boldsymbol{\tau}\right)\right)\right].
\end{align*}
To bound this empirical process, we use lemma \ref{maxima_of_empirical_process}. We define the following function class.
\begin{equation}\label{definition_M}
    \mathcal{M} := \left\{m(\xi,\cdot) = C_3 \sum_{h=1}^H \nu_h^{\top} \Sigma_h^{-1} \left(z_{(h)}\left(\theta, \boldsymbol{\tau}\right) - z_{(h)}\left(\theta^*, \boldsymbol{\tau}\right)\right) + \frac{1}{2}\bigg| \left\|\theta - \theta^*\right\| \leq \frac{B(\delta)}{\sqrt{K}}\right\}
\end{equation}
Since lemma \ref{maxima_of_empirical_process} requires all function in a certain class take values in $[0,1],$ the constant $C_3$ and $\frac{1}{2}$ in the expression make each function in $\mathcal{M}$ satisfy this requirement and $\frac{1}{2}$ will not influence bracketing number of this function class. Each function in $\mathcal{M}$ is indexed by $\theta \in \Theta^H$ and is from $\left(\mathcal{S}\times \mathcal{A}\right)^H$ to $[0,1].$ We denote 
\begin{equation}\label{definition_B_0}
    B_0 := \sup_{h \in [H]} \sqrt{\nu_h^{\top}\Sigma_h^{-2}\nu_h}.
\end{equation}
Notice that here the $K$ is fixed, hence without loss of generosity we let $K \geq B(\delta)^2.$ Then for every $\theta^1,\theta^2 \in \Theta^H$ such that $\left\|\theta^i - \theta^*\right\| \leq  \frac{B(\delta)}{\sqrt{K}} \leq 1,$ we have
\begin{align*}
    &\left|\nu_h^{\top} \Sigma_h^{-1} \left(z_{(h)}\left(\theta^1, \boldsymbol{\tau}\right) - z_{(h)}\left(\theta^2, \boldsymbol{\tau}\right)\right)\right|\\
    \leq & \sum_{j=1}^H \left|\nu_h^{\top} \Sigma_h^{-1} \left(f(\theta_h^1,\phi_j) - f(\theta^2_h,\phi_j) + \int_{\mathcal{A}} f\left(\theta^2_{h+1}, \phi_{j+1}\right) \pi\left(a^{\prime}, s_{j+1}\right) d a^{\prime} - \int_{\mathcal{A}} f\left(\theta^1_{h+1}, \phi_{j+1}\right) \pi\left(a^{\prime}, s_{j+1}\right) d a^{\prime}\right) \cdot \nabla_{\theta}^{\top} f(\theta_h^1,\phi_j)\right| \\
    & + \sum_{j=1}^H \left|\nu_h^{\top} \Sigma_h^{-1} \left(f(\theta_h^2,\phi_j) -  r_j - \int_{\mathcal{A}} f\left(\theta_{h+1}^2, \phi_{j+1}\right) \pi\left(a^{\prime}, s_{j+1}\right) d a^{\prime}\right) \left(\nabla_{\theta}^{\top} f(\theta_h^1,\phi_j) - \nabla_{\theta}^{\top} f(\theta^2_h,\phi_j)\right)\right| \\
    \leq & \sum_{j=1}^H \sqrt{\nu_h^{\top} \Sigma_h^{-2} \nu_h} \sqrt{d} \kappa_1 \bigg[\left\|\theta_h^1 - \theta^2_h\right\| + \left\|\theta_{h+1}^1 - \theta^2_{h+1}\right\|\bigg] \left\|\nabla_{\theta}^{\top} f(\theta_h^1,\phi_j)\right\|\\
    & + \sum_{j=1}^H \sqrt{\nu_h^{\top} \Sigma_h^{-2} \nu_h} d \kappa_2 \left\|\theta^1_h - \theta^2_h\right\| \left|\left(f(\theta_h^2,\phi_j) -  r_j - \int_{\mathcal{A}} f\left(\theta_{h+1}^2, \phi_{j+1}\right) \pi\left(a^{\prime}, s_{j+1}\right) d a^{\prime}\right)\right| \\
    \leq & H d \kappa_1^2 B_0 \bigg[\left\|\theta_h^1 - \theta^2_h\right\| + \left\|\theta_{h+1}^1 - \theta^2_{h+1}\right\|\bigg] \\
    & + H d \kappa_2 B_0 \left\|\theta^1_h - \theta^2_h\right\| \left[\left(f(\theta_h^*,\phi_j) -  r_j - \int_{\mathcal{A}} f\left(\theta_{h+1}^*, \phi_{j+1}\right) \pi\left(a^{\prime}, s_{j+1}\right) d a^{\prime}\right) + \sqrt{d}\kappa_1 \bigg[\left\|\theta_h^* - \theta^2_h\right\| + \left\|\theta_{h+1}^* - \theta^2_{h+1}\right\|\bigg]\right].
\end{align*}
Therefore,
\begin{align*}
    & \left| \sum_{h=1}^H \nu_h^{\top} \Sigma_h^{-1} \left(z_{(h)}\left(\theta^1, \boldsymbol{\tau}\right) - z_{(h)}\left(\theta^2, \boldsymbol{\tau}\right)\right) \right| \\ 
    \leq & H^{\frac{3}{2}} d \kappa_1^2 B_0 \left\|\theta^1 - \theta^2\right\| + H d \kappa_2 B_0 \sum_{h=1}^H \left(H-h+1\right) \left\|\theta_h^1 - \theta_h^2\right\| + H d^{\frac{3}{2}} \kappa_1 \kappa_2 B_0 \sum_{h=1}^H \left\|\theta_h^1 - \theta_h^2\right\| \bigg[\left\|\theta_h^* - \theta^2_h\right\| + \left\|\theta_{h+1}^* - \theta^2_{h+1}\right\|\bigg] \\
    \leq & \bigg[H^{\frac{3}{2}} d \kappa_1^2 B_0 + H^{\frac{5}{2}} d \kappa_2 B_0 + 2 H d^{\frac{3}{2}} \kappa_1 \kappa_2 B_0 \left\|\theta^* - \theta^2\right\|\bigg] \left\|\theta^1 - \theta^2\right\| \\
    \leq & \bigg[H^{\frac{3}{2}} d \kappa_1^2 B_0 + H^{\frac{5}{2}} d \kappa_2 B_0 + 2 H d^{\frac{3}{2}} \kappa_1 \kappa_2 B_0\bigg] \left\|\theta^1 - \theta^2\right\|
\end{align*}
A special case to the bound above is when $\theta^1 = \theta$ and $\theta^2 = \theta^*,$ this implies for arbitrary trajectory $\boldsymbol{\tau}$ and $\theta$ such that $\left\|\theta - \theta^*\right\| \leq \frac{B(\delta)}{\sqrt{K}} \leq 1,$ we have
\begin{equation*}
    \left| \sum_{h=1}^H \nu_h^{\top} \Sigma_h^{-1} \left(z_{(h)}\left(\theta, \boldsymbol{\tau}\right) - z_{(h)}\left(\theta^*, \boldsymbol{\tau}\right)\right) \right| \leq  \bigg[H^{\frac{3}{2}} d \kappa_1^2 B_0 + H^{\frac{5}{2}} d \kappa_2 B_0 + 2 H d^{\frac{3}{2}} \kappa_1 \kappa_2 B_0\bigg] \frac{B(\delta)}{\sqrt{K}}.
\end{equation*}
Therefore, we take
\begin{equation*}
    C_3 := \bigg[H^{\frac{3}{2}} d \kappa_1^2 B_0 + H^{\frac{5}{2}} d \kappa_2 B_0 + 2 H d^{\frac{3}{2}} \kappa_1 \kappa_2 B_0\bigg]^{-1} \frac{\sqrt{K}}{2 B(\delta)},
\end{equation*}
then each function in $\mathcal{M}$ takes value in $[0,1].$ Notice that all functions in $\mathcal{M}$ are Lipschitz continuous, we can give an upper bound for its Lipschitz norm and similar to proof of Theorem \ref{thm_normality} we can bound the bracketing number. The definition of $C_3$ and the bound for Lipschitz norm of functions in $\mathcal{M}$ imply that the difference of any two functions in $\mathcal{M}$ can be bounded by
\begin{equation*}
    \left|m(\theta^1,\boldsymbol{\tau}) - m(\theta^2,\boldsymbol{\tau})\right| \leq \frac{\sqrt{K}}{2 B(\delta)} \left\|\theta^1-\theta^2\right\|.
\end{equation*}
Similar to the calculation in the proof of Theorem \ref{thm_normality}, we have
\begin{equation*}
    N_{[]}\left(\varepsilon,\mathcal{M},L_2(\mathbb{P})\right)
    \leq \left(\frac{2\sqrt{Hd} \frac{\sqrt{K}}{2 B(\delta)} \frac{B(\delta)}{\sqrt{K}}}{\varepsilon}\right)^{Hd} = \left(\frac{\sqrt{Hd}}{\varepsilon}\right)^{Hd}
\end{equation*}
We take $V = Hd$ and $U = \sqrt{Hd},$ then using lemma \ref{maxima_of_empirical_process}, we have
\begin{equation*}
    \mathbb{P}\left(\left\|\mathbb{G}_K\right\|_{\mathcal{M}} > t\right) \leq \left(\frac{Dt}{\sqrt{Hd}}\right)^{Hd} \exp(-2t^2) \leq \left(\frac{D}{\sqrt{Hd}}\right)^{Hd} \exp(-2t^2 + Hd t).
\end{equation*}
where $D > 0$ is a constant dependent on $U$ only. If we take
\begin{equation*}
    t = \ln\left(\frac{1}{\delta}\right) + Hd \ln\left(e + \frac{D}{\sqrt{Hd}}\right),
\end{equation*}
then conditional on $\mathcal{E}_K,$ with probability at least $1-\delta,$ we have
\begin{equation*}
    \left\|\mathbb{G}_K\right\|_{\mathcal{M}} \leq \ln\left(\frac{1}{\delta}\right) + Hd \ln\left(e + \frac{D}{\sqrt{Hd}}\right).
\end{equation*}
Therefore when $K \geq B(\delta)^2,$ under $\mathcal{E}_K,$ with probability at least $1-\delta,$
\begin{align*}
    \left|I_{22}\right|
    & \leq \frac{1}{C_3 H \sqrt{K}}\left\|\mathbb{G}_K\right\|_{\mathcal{M}} \\
    & \leq \frac{2 B(\delta)}{H K} \left[H^{\frac{3}{2}} d \kappa_{1}^{2} B_{0}+H^{\frac{5}{2}} d \kappa_{2} B_{0}+2 H d^{\frac{3}{2}} \kappa_{1} \kappa_{2} B_{0}\right] \left[\ln\left(\frac{1}{\delta}\right) + Hd \ln\left(e + \frac{D}{\sqrt{Hd}}\right)\right].
\end{align*}

Finally we give an high-probability upper bound for $I_{23}.$ We have
\begin{equation*}
    I_{23} = -\frac{1}{H} \sum_{h=1}^H \nu_h^{\top} \Sigma_{h}^{-1} R_{K, h}.
\end{equation*}
where $R_{K,h} = (R_{K,h}^1,R_{K,h}^2,...R_{K.h}^d)^{\top}$ is the Taylor remainder of $\mathbb{E} z_{(h)}(\theta,\boldsymbol{\tau})$ at $\theta^*.$ Note that $\mathbb{E} z_{(h)}(\theta,\boldsymbol{\tau})$ is only dependent on $\theta_h,\theta_{h+1}.$ If we denote $\Delta \theta_h := ((\widehat{\theta}_{K,h} - \theta^*_h)^{\top}, (\widehat{\theta}_{K,h+1} - \theta^*_{h+1})^{\top})^{\top},$ By Taylor Theorem, we have for some $c \in (0,1),$
\begin{equation*}
    R_{K,h}^i = \sum_{\left|\alpha\right| = 2} \frac{\left(\Delta \theta_h\right)^2}{\alpha} \mathbb{E} \left[\partial^{\alpha}_{\theta_h,\theta_{h+1}} z_{(h)}^i\left(\theta^* + c\left(\widehat{\theta}_K - \theta^*\right),\boldsymbol{\tau}\right)\right].
\end{equation*}
where the notation keeps the same as \eqref{notation1} and $\partial^{\alpha}_{\theta_h,\theta_{h+1}}$ means the $\alpha$ order derivatives with respect to $\theta_h,\theta_{h+1}$ only. Remember for $\theta = (\theta_1^1,\theta_1^2,...,\theta_1^d,\theta_2^1,...,\theta_H^d)^{\top}$ and $\theta_h = (\theta_h^1,...,\theta_h^d)^{\top}$ we have
$$
z_{(h)}^i(\theta, \boldsymbol{\tau})=\sum_{j=1}^{H}\left(f\left(\theta_{h}, \phi_{j}\right)-r_{j}-\int_{\mathcal{A}} f\left(\theta_{h+1}, \phi\left(s_{j+1}, a^{\prime}\right)\right) \pi\left(a^{\prime} \mid s_{j+1}\right) d a^{\prime}\right) \cdot \nabla_{\theta_h}^i f\left(\theta_{h}, \phi_{j}\right).
$$
We can compute the upper bound for partial derivatives of $z_{(h)}^i.$ We have under $\mathcal{E}_K,$ for any trajectory $\boldsymbol{\tau},$
$$
\left|\left(f\left(\theta_{h}, \phi_{n}\right)-r_{n}- \int_{\mathcal{A}} f\left(\theta_{h+1}, \phi\left(s_{n+1}, a^{\prime}\right)\right) \pi\left(a^{\prime} \mid s_{n+1}\right) d a^{\prime}\right)\right| \leq D_0 := (H-h+1) + 2\sqrt{d}\kappa_1 \frac{B(\delta)}{\sqrt{K}}.
$$
Therefore we have for $i, \in [d],$
\begin{align*}
    \left|\frac{\partial^2}{\partial \theta_h^i \partial \theta_h^j} z_{(h)}^i\left(\theta,\boldsymbol{\tau}\right)\right| \leq H \left(\kappa_3 D_0 + 3\kappa_1 \kappa_2 \right), \quad
    \left|\frac{\partial^2}{\partial \theta_h^i \partial \theta_{h+1}^j} z_{(h)}^i\left(\theta,\boldsymbol{\tau}\right)\right| \leq H \kappa_1 \kappa_2,\quad
    \left|\frac{\partial^2}{\partial \theta_{h+1}^i \partial \theta_{h+1}^j} z_{(h)}^i\left(\theta,\boldsymbol{\tau}\right)\right| \leq H \kappa_1 \kappa_2,
\end{align*}
Therefore, we have
\begin{align*}
    \left|R_{K,h}^i\right| 
    &\leq \frac{H(\kappa_3 D_0 + 3 \kappa_1 \kappa_2)}{2} \sum_{\left|\alpha\right| = 2} \left|\left(\Delta \theta_h\right)^\alpha\right| 
    \leq \frac{H(\kappa_3 D_0 + 3 \kappa_1 \kappa_2)}{2} \left[\left\|\widehat{\theta}_{K,h} - \theta_h^*\right\|_1 + \left\|\widehat{\theta}_{K,h+1} - \theta_{h+1}^*\right\|_1\right]^2 \\
    & \leq H^2(\kappa_3 D_0 + 3 \kappa_1 \kappa_2) \left[\left\|\widehat{\theta}_{K,h} - \theta_h^*\right\|_2^2 + \left\|\widehat{\theta}_{K,h+1} - \theta_{h+1}^*\right\|_2^2\right].
\end{align*}
Hence under $\mathcal{E}_K,$
\begin{equation*}
    \left\|R_{K,h}\right\|_2 \leq \left\|R_{K,h}\right\|_1 \leq H^2 d (\kappa_3 D_0 + 3 \kappa_1 \kappa_2) \frac{B(\delta)^2}{K}
\end{equation*}
Therefore, we have
\begin{equation*}
    \left|I_{23}\right| \leq B_0 H^3 d \left[\kappa_3 H + 3 \kappa_1 \kappa_2 + 2\sqrt{d} \kappa_1 \kappa_3 \frac{B(\delta)}{\sqrt{K}} \right]\frac{B(\delta)^2}{K}.
\end{equation*}

\subsection{Completion of the proof}
We combine the bound for first order and higher order term and formalize the final upper bound. We only prove for reward-free error bound \eqref{finite_bound1}. For variance-aware error bound, the proof is almost the same, except that we use the other bound for first order term. For now, we fix a $K$ such that
$$
    K \geq \max\left\{\frac{2(C_2 d + 1)^2}{C_2 d}\log\left(\frac{12dH}{\delta}\right),B(\delta/3)^2\right\},
$$
then with probability at least $1-\frac{\delta}{3},$ 
\begin{equation*}
    \left|I_1\right| \leq \sqrt{\frac{\log (12 / \delta)}{2 K H}}\left(\sum_{h=1}^{H}(H-h+1) \sqrt{\nu_{h}^{\top} \Sigma_{h}^{-1} \nu_{h}}\right) + \left(\sum_{h=1}^{H} (H-h+1)\sqrt{\nu_{h}^{\top} \Sigma_{h}^{-1} \nu_{h}}\right)\left[\frac{4}{3 K} \sqrt{C_2 d H} \log \left(\frac{12 d H}{\delta}\right)\right].
\end{equation*}
At the same time, with probability at least $1-\frac{\delta}{3},$
\begin{equation}\label{bound_higher_order1}
    \left\|\widehat{\theta}_{K}-\theta^{*}\right\|_{2} \leq \frac{B(\delta/3)}{\sqrt{K}}.
\end{equation}
Under \eqref{bound_higher_order1}, we have
\begin{align*}
    \left|I_{21}\right| 
    &\leq \frac{\kappa_2 d B(\delta/3)^2}{2K}; \\
    \left|I_{23}\right| 
    &\leq B_{0} H^{3} d\left[\kappa_{3} H+3 \kappa_{1} \kappa_{2}+2 \sqrt{d} \kappa_{1} \kappa_{3} \frac{B(\delta/3)}{\sqrt{K}}\right] \frac{B(\delta/3)^{2}}{K}.
\end{align*}
and with probability at least $1 - \frac{\delta}{3},$
\begin{align*}
    \left|I_{22}\right| 
    &\leq \frac{2 B(\delta)}{H K}\left[H^{\frac{3}{2}} d \kappa_{1}^{2} B_{0}+H^{\frac{5}{2}} d \kappa_{2} B_{0}+2 H d^{\frac{3}{2}} \kappa_{1} \kappa_{2} B_{0}\right]\left[\ln \left(\frac{3}{\delta}\right)+H d \ln \left(e+\frac{D}{\sqrt{H d}}\right)\right]
\end{align*}
By union bound, we have when $K\geq B(\delta/3)^2,$ with probability at least $1-\delta,$
\begin{align*}
    \left|\widehat{v}_{\pi} - v_{\pi}\right| 
    & \leq \sqrt{\frac{\log (12 / \delta)}{2 K H}}\left(\sum_{h=1}^{H}(H-h+1) \sqrt{\nu_{h}^{\top} \Sigma_{h}^{-1} \nu_{h}}\right) + \left(\sum_{h=1}^{H} (H-h+1) \sqrt{\nu_{h}^{\top} \Sigma_{h}^{-1} \nu_{h}}\right)\left[\frac{4}{3 K} \sqrt{C_{2} d H} \log \left(\frac{12 d H}{\delta}\right)\right] \\
    & + \frac{B(\delta/3)}{K} \left[H^{2\frac{1}{2}} d \kappa_{1}^{2} B_{0}+2H^{\frac{3}{2}} d \kappa_{2} B_{0}+4 d^{\frac{3}{2}} \kappa_{1} \kappa_{2} B_{0}\right] + \frac{B(\delta/3)^2}{K} \left[B_{0} H^{3} d\left[\kappa_{3} H+3 \kappa_{1} \kappa_{2}+2 \sqrt{d} \kappa_{1} \kappa_{3}\right]\right],
\end{align*}
which implies the upper bound in Theorem \ref{thm_finite_sample_square}. Here $D$ is a constant dependent only on $H$ and $d.$

\section{Proof and Discussion for Special Cases}
\subsection{Special Case: With Positivity Condition}\label{sec:positivity}
The only difference brought about by additional positivity condition will be on the bound for the first order term. Since
\begin{equation*}
    I_1 = -\frac{1}{KH} \sum_{k=1}^K \sum_{h=1}^{H} \sum_{j=1}^{H} \nu_{h}^{\top} \Sigma_{h}^{-1}\left[\nabla_{\theta}^{\top} f\left(\theta_{h}^*, \phi\left(s_{j}^{k}, a_{j}^{k}\right)\right)\right] \varepsilon_{h, j}^{k},
\end{equation*}
Similarly we denote $\varepsilon_{h,n} = \varepsilon_{h,j}^k$ and $(s_n,a_n) = (s_j^k,a_j^k)$ if $n = (j-1)K + k,$ and we define
\begin{equation*}
    u_n := -\sum_{h=1}^H \nu_{h}^{\top} \Sigma_{h}^{-1}\left[\nabla_{\theta}^{\top} f\left(\theta_{h}^*, \phi\left(s_n, a_n\right)\right)\right] \varepsilon_{h, n},
\end{equation*}
then
\begin{equation*}
    I_1 = \frac{1}{N} \sum_{n=1}^N u_n.
\end{equation*}
Define $\mathcal{F}_n$ is the $\sigma$ field generated by $s_1,a_1,....,s_n,a_n,$ then $\left\{\mathcal{F}_n\right\}_{n=1}^N$ is a filtration. Since $\mathbb{E}\left[u_n \mid \mathcal{F}_n\right] = 0,$ we have $\left\{u_n\right\}_{n=1}^N$ is a martingale difference sequence. We use the following \ref{ine_freedman} to analyze $I_1.$ Parallel to the previus analysis, it is easy to see that
\begin{equation*}
    \left|u_n\right| \leq \sqrt{C_{2} d} \sum_{h=1}^{H}(H-h+1) \sqrt{\nu_{h}^{\top} \Sigma_{h}^{-1} \nu_{h}}
\end{equation*}
The main difference with Theorem \ref{thm_finite_sample_square} is the estimate of variance $\operatorname{Var}\left[u_n\mid \mathcal{F}_n\right].$ We expand this conditional variance into
\begin{align*}
    \operatorname{Var}\left[u_n\mid \mathcal{F}_n\right]
    &= \mathbb{E}\left[\left(\sum_{h=1}^H \nu_{h}^{\top} \Sigma_{h}^{-1}\left[\nabla_{\theta}^{\top} f\left(\theta_{h}^*, \phi\left(s_n, a_n\right)\right)\right] \varepsilon_{h, n}\right)^2 \bigg| \mathcal{F}_n\right] \\
    &= \sum_{h_1=1}^H \sum_{h_2=1}^H \bigg(\nu_{h_1}^{\top} \Sigma_{h_1}^{-1}\nabla_{\theta}^{\top} f\left(\theta_{h_1}^*, \phi\left(s_n, a_n\right)\right)\bigg) \bigg(\nu_{h_2}^{\top} \Sigma_{h_2}^{-1}\nabla_{\theta}^{\top} f\left(\theta_{h_2}^*, \phi\left(s_n, a_n\right)\right)\bigg) \mathbb{E}\left[\varepsilon_{h_1, n} \varepsilon_{h_2, n} \bigg| \mathcal{F}_n\right].
\end{align*}
By definition of $\varepsilon_{h,n},$ we have
\begin{align*}
    \mathbb{E}\left[\varepsilon_{h_1, n} \varepsilon_{h_2, n} \bigg| \mathcal{F}_n\right]
    &= \operatorname{Cov}\left[r\left(s_n, a_n\right) + \int_{\mathcal{A}} f\left(\theta_{h_1+1}^*, \phi\left(s_{n+1}, a^{\prime}\right)\right) \pi\left(a^{\prime} \mid s_{n+1}\right) d a^{\prime},\right.\\
    &\hspace{8ex} \left.r\left(s_n, a_n\right) + \int_{\mathcal{A}} f\left(\theta_{h_2+1}^*, \phi\left(s_{n+1}, a^{\prime}\right)\right) \pi\left(a^{\prime} \mid s_{n+1}\right) d a^{\prime} \bigg| \mathcal{F}_n\right] \\
    & \leq \left[\operatorname{Var}\left(r\left(s_n, a_n\right) + \int_{\mathcal{A}} f\left(\theta_{h_1+1}^*, \phi\left(s_{n+1}, a^{\prime}\right)\right) \pi\left(a^{\prime} \mid s_{n+1}\right) d a^{\prime}\bigg| \mathcal{F}_n\right)\right.\\
    & \hspace{8ex} \left.\operatorname{Var}\left(r\left(s_n, a_n\right) + \int_{\mathcal{A}} f\left(\theta_{h_2+1}^*, \phi\left(s_{n+1}, a^{\prime}\right)\right) \pi\left(a^{\prime} \mid s_{n+1}\right) d a^{\prime}\bigg| \mathcal{F}_n\right)\right]^{\frac{1}{2}} \tag{Cauchy-Schwarz}\\
    &\leq \frac{1}{4} \left(H-h_1+1\right) \left(H-h_2+1\right).
\end{align*}
The last inequality comes from $r\left(s_n, a_n\right) + \int_{\mathcal{A}} f\left(\theta_{h+1}^*, \phi\left(s_{n+1}, a^{\prime}\right)\right) \pi\left(a^{\prime} \mid s_{n+1}\right) d a^{\prime} \in [0,H-h+1].$ Under the condition of $\nabla_{\theta} f(\theta_h^*,\phi(s,a)) \Sigma_h^{-1} \nabla_{\theta}^{\top} f(\theta_h^*,\phi(s^{\prime},a^{\prime})) \geq 0$ for any $(s,a),(s^{\prime},a^{\prime}) \in \mathcal{S} \times \mathcal{A},$ we have
\begin{align*}
    \operatorname{Var}\left[u_n\mid \mathcal{F}_n\right]
    & \leq \frac{1}{4} \sum_{h_1=1}^H \sum_{h_2=1}^H \bigg(\nu_{h_1}^{\top} \Sigma_{h_1}^{-1}\nabla_{\theta}^{\top} f\left(\theta_{h_1}^*, \phi\left(s_n, a_n\right)\right)\bigg) \bigg(\nu_{h_2}^{\top} \Sigma_{h_2}^{-1}\nabla_{\theta}^{\top} f\left(\theta_{h_2}^*, \phi\left(s_n, a_n\right)\right)\bigg) (H - h_1 + 1)(H - h_2 + 1) \\
    & = \left(\frac{1}{2} \sum_{h=1}^H (H-h+1)\nu_{h}^{\top} \Sigma_{h}^{-1}\nabla_{\theta}^{\top} f\left(\theta_{h}^*, \phi\left(s_n, a_n\right)\right)\right)^2.
\end{align*}
Therefore,
\begin{align}
    \sum_{n=1}^N \operatorname{Var}\left[u_n\mid \mathcal{F}_n\right]
    &\leq \frac{1}{4} \sum_{h_1=1}^H \sum_{h_2=1}^H (H-h_1+1)(H-h_2+1)\nu_{h_1}^{\top} \Sigma_{h_1}^{-1} \left(\sum_{n=1}^N \nabla_{\theta}^{\top}f\left(\theta_{h_1}^*, \phi\left(s_n, a_n\right)\right) \nabla_{\theta} f\left(\theta_{h_2}^*, \phi\left(s_n, a_n\right)\right) \right) \Sigma_{h_2}^{-1} \nu_{h_2} \label{positivity_variance}
\end{align}
We use lemma \ref{lemma_first_order} to bound
\begin{equation*}
    \left\|\Sigma_{h_1}^{-\frac{1}{2}} \left(\sum_{n=1}^N \nabla_{\theta}^{\top}f\left(\theta_{h_1}^*, \phi\left(s_n, a_n\right)\right) \nabla_{\theta} f\left(\theta_{h_2}^*, \phi\left(s_n, a_n\right)\right) \right) \Sigma_{h_2}^{-\frac{1}{2}}\right\|.
\end{equation*} 
\begin{lemma}\label{lemma_first_order}
   Under the assumption that for any $h \in [H],$
   $$
   \nabla_{\theta} f(\theta^*_h, \phi(s, a)) \Sigma_{h}^{-1} \nabla_{\theta}^{\top} f(\theta^*_h, \phi(s, a)) \leq C_{2} d,
   $$
   with probability at least $1 - \delta,$
   \begin{align*}
   &\left\|\Sigma_{h_1}^{-\frac{1}{2}} \left(\sum_{n=1}^N \nabla_{\theta}^{\top}f\left(\theta_{h_1}^*, \phi\left(s_n, a_n\right)\right) \nabla_{\theta} f\left(\theta_{h_2}^*, \phi\left(s_n, a_n\right)\right) \right) \Sigma_{h_2}^{-\frac{1}{2}}\right\| \\
   \leq& N\left(\sigma_{h_{1}, h_{2}}+\sqrt{\frac{2 C_{2} d}{K} \log \left(\frac{2 d }{\delta}\right)}+\frac{2\left(C_{2} d+\sigma_{h_{1}, h_{2}}\right)}{3 K} \log \left(\frac{2 d }{\delta}\right)\right),
   \end{align*}
   where $\sigma_{h_{1}, h_{2}}:=\left\|\Sigma_{h_{1}}^{-\frac{1}{2}} \Sigma_{h_{1}, h_{2}} \Sigma_{h_{2}}^{-\frac{1}{2}}\right\|_{2}$ and
   $$
   \Sigma_{h_1,h_2} := \mathbb{E}\left[\frac{1}{H}\sum_{j=1}^H \nabla_{\theta}^{\top}f\left(\theta_{h_1}^*, \phi\left(s_j, a_j\right)\right) \cdot \nabla_{\theta}f\left(\theta_{h_2}^*, \phi\left(s_j, a_j\right)\right)\right].
   $$
\end{lemma}
The proof of lemma \ref{lemma_first_order} is deferred to \ref{sec:lemma}. By this lemma we have with fixed $h_1,h_2$ and with probability at least $1 - \frac{\delta}{H^2},$
\begin{align}
    & \left\|\Sigma_{h_{1}}^{-\frac{1}{2}}\left(\sum_{n=1}^{N} \nabla_{\theta}^{\top} f\left(\theta_{h_{1}}, \phi\left(s_{n}, a_{n}\right)\right) \nabla_{\theta} f\left(\theta_{h_{2}}, \phi\left(s_{n}, a_{n}\right)\right)\right) \Sigma_{h_{2}}^{-\frac{1}{2}}\right\| \notag \\
    \leq&  N\left(\sigma_{h_1,h_2} + \sqrt{\frac{2C_2 d}{K}\log\left(\frac{2dH^2}{\delta}\right)} + \frac{2\left(C_2d+\sigma_{h_1,h_2}\right)}{3 K}\log\left(\frac{2dH^2}{\delta}\right)\right). \label{bound_finiteC1}
\end{align}
By union bound we have with probability at least $1 - \delta,$ \eqref{bound_finiteC1} holds for every $h_1,h_2 \in [H].$ This gives with probability at least $1 - \delta,$
\begin{align*}
    \sum_{n=1}^{N} \operatorname{Var}\left[u_{n} \mid \mathcal{F}_{n}\right]
    \leq & \frac{1}{4} \sum_{h_{1}=1}^{H}\sum_{h_{2}=1}^{H} \left(H-h_{1}+1\right)\left(H-h_{2}+1\right)\\
    \leq & \frac{KH}{4}  \sum_{h_{1}=1}^{H}\sum_{h_{2}=1}^{H} \left(H-h_{1}+1\right)\left(H-h_{2}+1\right) \sqrt{\nu_{h_1}^{\top} \Sigma_{h_1}^{-1} \nu_{h_1}} \sqrt{\nu_{h_2}^{\top} \Sigma_{h_2}^{-1} \nu_{h_2}} \\
    & \cdot \left(\sigma_{h_1,h_2} + \sqrt{\frac{2C_2 d}{K}\log\left(\frac{2dH^2}{\delta}\right)} + \frac{2\left(C_2d+\sigma_{h_1,h_2}\right)}{3 K}\log\left(\frac{2dH^2}{\delta}\right)\right).
\end{align*}
We take
\begin{align*}
    \sigma^2_0 &:= \frac{KH}{4}  \sum_{h_{1}=1}^{H}\sum_{h_{2}=1}^{H} \left(H-h_{1}+1\right)\left(H-h_{2}+1\right) \sqrt{\nu_{h_1}^{\top} \Sigma_{h_1}^{-1} \nu_{h_1}} \sqrt{\nu_{h_2}^{\top} \Sigma_{h_2}^{-1} \nu_{h_2}} \\
    & \cdot \left(\sigma_{h_1,h_2} + \sqrt{\frac{2C_2 d}{K}\log\left(\frac{4dH^2}{\delta}\right)} + \frac{2\left(C_2d+\sigma_{h_1,h_2}\right)}{3 K}\log\left(\frac{4dH^2}{\delta}\right)\right),
\end{align*}
then
\begin{equation*}
    \mathbb{P}\left(\sum_{n=1}^{N} \operatorname{Var}\left[u_{n} \mid \mathcal{F}_{n}\right] \geq \sigma^2_0 \right) \leq \frac{\delta}{2}.
\end{equation*}
Freedman's Inequality implies that for any $\varepsilon > 0,$ we have
\begin{equation*}
    \mathbb{P}\left(\left|\sum_{n=1}^N u_n\right| \geq \varepsilon, \sum_{n=1}^{N} \operatorname{Var}\left[u_{n} \mid \mathcal{F}_{n}\right] \leq \sigma^2_0\right) \leq 2\exp\left(-\frac{\varepsilon^2/2}{\sigma^2_0 + \varepsilon \sqrt{C_{2} d} \sum_{h=1}^{H}(H-h+1) \sqrt{\nu_{h}^{\top} \Sigma_{h}^{-1} \nu_{h}} / 3}\right).
\end{equation*}
We take
\begin{equation*}
    \varepsilon := \sqrt{2 \log\left(\frac{4}{\delta}\right)} \sigma_0 + \log\left(\frac{4}{\delta}\right)\frac{2 \sqrt{C_{2} d} \sum_{h=1}^{H}(H-h+1) \sqrt{\nu_{h}^{\top} \Sigma_{h}^{-1} \nu_{h}}}{3},
\end{equation*}
then
\begin{equation*}
    \mathbb{P}\left(\left|\sum_{n=1}^N u_n\right| \geq \varepsilon \right) \leq \mathbb{P}\left(\sum_{n=1}^{N} \operatorname{Var}\left[u_{n} \mid \mathcal{F}_{n}\right] \geq \sigma^2_0 \right) + \mathbb{P}\left(\left|\sum_{n=1}^N u_n\right| \geq \varepsilon, \sum_{n=1}^{N} \operatorname{Var}\left[u_{n} \mid \mathcal{F}_{n}\right] \leq \sigma^2_0\right) \leq \frac{\delta}{2} + \frac{\delta}{2} = \delta.
\end{equation*}
Notice that
\begin{equation*}
    \frac{\varepsilon}{N} = \sqrt{\frac{\log(4 / \delta)}{2HK}} \cdot \sqrt{\sum_{h_{1}=1}^{H} \sum_{h_{2}=1}^{H}\left(H-h_{1}+1\right)\left(H-h_{2}+1\right) \sqrt{\nu_{h_{1}}^{\top} \Sigma_{h_{1}}^{-1} \nu_{h_{1}}} \sqrt{\nu_{h_{2}}^{\top} \Sigma_{h_{2}}^{-1} \nu_{h_{2}}} \sigma_{h_1,h_2}} + O(\frac{1}{K}),
\end{equation*}
we insert this into the first order term and hence finish the proof.

\subsection{Special Case: Linear Function Class}

Before we provide the result in linear function case, we briefly describe the simplest case with linear function approximation $f(\theta,\phi) := \theta^{\top}\phi(s,a)$ \citep{bootstrap}. Assumption \ref{assumption_completeness} for policy completeness here is equivalent to that there exists $M \in \mathbb{R}^{d \times d}$ such that
$$
\phi(s, a)^{\top} M =\mathbb{E}\left[\left(\int_{\mathcal{A}} \phi(s^{\prime},a^{\prime}) \pi(a^{\prime}\mid s^{\prime})\right)^{\top} \bigg| s, a\right].
$$
This implies that for $h \in [H],$ we have $\theta_h^* = M \theta_{h+1}^* + \theta_r^*$ and $\theta_H^* = \theta_r^*$ for some $\theta_r^* \in \Theta.$ $\theta_r^*$ is the parameter of reward function $r(s,a).$ In linear setting, \eqref{FQE} has following explicit expression.
\begin{equation}\label{special_case1}
    \widehat{\theta}_h = \widehat{M} \widehat{\theta}_{h+1} + \widehat{R}.
\end{equation}
where
$$
\widehat{M}=\widehat{\Sigma}^{-1} \sum_{n=1}^{N} \phi\left(s_{n}, a_{n}\right) \left(\int_{\mathcal{A}} \phi(s_{n+1},a^{\prime}) \pi(a^{\prime}\mid s_{n+1})\right)^{\top}, \widehat{R}=\widehat{\Sigma}^{-1} \sum_{n=1}^{N} r_{n} \phi\left(s_{n}, a_{n}\right), \widehat{\Sigma}=\sum_{n=1}^{N} \phi\left(s_{n}, a_{n}\right) \phi\left(s_{n}, a_{n}\right)^{\top}+\lambda I_{d}
$$
\cite{bootstrap} showed that, the limit distribution of $\sqrt{N}\left(\widehat{v}_{\pi} - v_{\pi}\right)$ is $N(0,\sigma^2),$ where 
\begin{equation}\label{variance_linear_case}
    \sigma^{2}=\sum_{h_1=1}^{H} \sum_{h_2=1}^{H} \left(\nu_{h_1}\right)^{\top} \Sigma^{-1} \Omega_{h_1, h_2} \Sigma^{-1} \nu_{h_2},
\end{equation}
where
\begin{align}
    \nu_{h} = \mathbb{E}^{\pi}\left[\phi\left(s_{h}, a_{h}\right) \mid s_{1} \sim \xi(\cdot)\right]; & \quad \Sigma=\mathbb{E}\left[\frac{1}{H} \sum_{h=1}^{H} \phi\left(s_{h}, a_{h}\right) \phi\left(s_{h}, a_{h}\right)^{\top}\right]; \notag \\
    \Omega_{h_{1}, h_{2}}=\mathbb{E}\left[\frac{1}{H} \sum_{h^{\prime}=1}^{H} \phi\left(s_{h^{\prime}}, a_{h^{\prime}}\right) \phi\left(s_{h^{\prime}}, a_{h^{\prime}}\right)^{\top} \varepsilon_{h_{1}, h^{\prime}} \varepsilon_{h_{2}, h^{\prime}}\right]; & \quad \varepsilon_{h_{1}, h^{\prime}}=Q_{h_{1}}\left(s_{h^{\prime}}, a_{h^{\prime}}\right)-\left(r_{h^{\prime}}+V_{h_{1}+1}\left(s_{h^{\prime}+1}\right)\right). \label{definitions_linear_case}
\end{align}
Further, they showed that FQE with linear function approximation achieves Cramer Rao lower bound for variance, and bootstrapping error with linear function approximation has the same limit distribution as standard FQE error. Actually, this expression can be immediately derived using our results. In linear case, we can provide a sharper bound for the dominant term in $\widehat{v}_\pi - v_\pi,$ which comes from a different variance calculation in Freeman's Inequality. Here, our data coverage assumption will be $\phi^{\top}(s,a)\Sigma^{-1}\phi(s,a) \leq C_2 d$ and the positivity condition will be $\phi^{\top}(s,a) \Sigma^{-1} \phi(s^\prime,a^\prime) \geq 0.$ Similar to the derivation before, we have $I_{1}=\frac{1}{N} \sum_{n=1}^{N} u_{n},$ where $u_{n}:=-\sum_{h=1}^{H} \nu_{h}^{\top} \Sigma^{-1}\left[\phi\left(s_{n}, a_{n}\right)\right] \varepsilon_{h, n}.$ Define $\mathcal{F}_{n}$ is the $\sigma$ field generated by $s_{1}, a_{1}, \ldots, s_{n}, a_{n}$, then $\left\{u_{n}\right\}_{n=1}^{N}$ is a martingale difference sequence and $\left|u_{n}\right| \leq \sqrt{C_{2} d} \sum_{h=1}^{H}(H-h+1) \sqrt{\nu_{h}^{\top} \Sigma_{h}^{-1} \nu_{h}}.$ In this case, \eqref{positivity_variance} becomes
\begin{align*}
    \sum_{n=1}^{N} \operatorname{Var}\left[u_{n} \mid \mathcal{F}_{n}\right] 
    & \leq \frac{1}{4} \sum_{h_{1}=1}^{H} \sum_{h_{2}=1}^{H}\left(H-h_{1}+1\right)\left(H-h_{2}+1\right) \nu_{h_{1}}^{\top} \Sigma^{-1}\left(\sum_{n=1}^{N}  \phi\left(s_{n}, a_{n}\right) \phi^{\top}\left(s_{n}, a_{n}\right)\right) \Sigma^{-1} \nu_{h_{2}} \\
    & \leq \frac{1}{4}\left(\sum_{h=1}^{H}(H-h+1) \nu_{h}\right)^{\top} \Sigma^{-1}\left(\sum_{n=1}^{N} \phi\left(s_{n}, a_{n}\right) \phi\left(s_{n}, a_{n}\right)^{\top}\right) \Sigma^{-1}\left(\sum_{h=1}^{H}(H-h+1) \nu_{h}\right) \\
    &  \leq \frac{1}{4}\left(\sum_{h=1}^{H}(H-h+1) \nu_{h}\right)^{\top} \Sigma^{-1} \left(\sum_{h=1}^{H}(H-h+1) \nu_{h}\right) \left\| \Sigma^{-\frac{1}{2}}\left(\sum_{n=1}^{N} \phi\left(s_{n}, a_{n}\right) \phi\left(s_{n}, a_{n}\right)^{\top}\right) \Sigma^{-\frac{1}{2}} \right\|.
\end{align*}
When we take $f$ as linear, lemma \ref{lemma_first_order} shows that with probability at least $1-\delta,$ 
\begin{equation*}
    \left\| \Sigma^{-\frac{1}{2}}\left(\sum_{n=1}^{N} \phi\left(s_{n}, a_{n}\right) \phi\left(s_{n}, a_{n}\right)^{\top}\right) \Sigma^{-\frac{1}{2}} \right\| \leq N\left(1+\sqrt{\frac{2 C_{2} d}{K} \log \left(\frac{2 d}{\delta}\right)}+\frac{2\left(C_{2} d + 1\right)}{3 K} \log \left(\frac{2 d}{\delta}\right)\right),
\end{equation*}
We take
\begin{equation*}
    \sigma^2 := \frac{N}{4}\left(\sum_{h=1}^{H}(H-h+1) \nu_{h}\right)^{\top} \Sigma^{-1} \left(\sum_{h=1}^{H}(H-h+1) \nu_{h}\right) \left(1+\sqrt{\frac{2 C_{2} d}{K} \log \left(\frac{2 d}{\delta}\right)}+\frac{2\left(C_{2} d + 1\right)}{3 K} \log \left(\frac{2 d}{\delta}\right)\right),
\end{equation*}
and
$$
\varepsilon:=\sqrt{2 \log \left(\frac{4}{\delta}\right)} \sigma+\log \left(\frac{4}{\delta}\right) \frac{2 \sqrt{C_{2} d} \sum_{h=1}^{H}(H-h+1) \sqrt{\nu_{h}^{\top} \Sigma_{h}^{-1} \nu_{h}}}{3}.
$$
By Freeman's Inequality, for any $\varepsilon > 0,$
$$
\mathbb{P}\left(\left|\sum_{n=1}^{N} u_{n}\right| \geq \varepsilon, \sum_{n=1}^{N} \operatorname{Var}\left[u_{n} \mid \mathcal{F}_{n}\right] \leq \sigma^{2}\right) \leq 2 \exp \left(-\frac{\varepsilon^{2} / 2}{\sigma^{2}+\varepsilon \sqrt{C_{2} d} \sum_{h=1}^{H}(H-h+1) \sqrt{\nu_{h}^{\top} \Sigma_{h}^{-1} \nu_{h} / 3}}\right),
$$
then
$$
\mathbb{P}\left(\left|\sum_{n=1}^{N} u_{n}\right| \geq \varepsilon\right) \leq \mathbb{P}\left(\sum_{n=1}^{N} \operatorname{Var}\left[u_{n} \mid \mathcal{F}_{n}\right] \geq \sigma^{2}\right)+\mathbb{P}\left(\left|\sum_{n=1}^{N} u_{n}\right| \geq \varepsilon, \sum_{n=1}^{N} \operatorname{Var}\left[u_{n} \mid \mathcal{F}_{n}\right] \leq \sigma^{2}\right) \leq \frac{\delta}{2}+\frac{\delta}{2}=\delta.
$$
Hence the upper bound for the first order term can be improved to
\begin{equation*}
    \left|I_1\right| \leq \sqrt{\left(\sum_{h=1}^{H}(H-h+1) \nu_{h}\right)^{\top} \Sigma^{-1}\left(\sum_{h=1}^{H}(H-h+1) \nu_{h}\right)} \cdot \sqrt{\frac{\ln (4/ \delta)}{2 K H}} + O(\frac{1}{K}).
\end{equation*}
By Cauchy-Schwarz Inequality, we have
\begin{equation*}
    \sqrt{\left(\sum_{h=1}^{H}(H-h+1) \nu_{h}\right)^{\top} \Sigma^{-1}\left(\sum_{h=1}^{H}(H-h+1) \nu_{h}\right)} = \sup_{\mu \in \mathbb{R}^d} \frac{\mu^{\top} \left(\sum_{h=1}^{H}(H-h+1) \nu_{h}\right)}{\sqrt{\mu^{\top} \Sigma \mu}}.
\end{equation*}
For $\mu \in \mathbb{R}^d,$ we take $g \in \mathcal{G}$ such that $g(s,a) = \phi^{\top}(s,a) \mu$ for any $(s,a) \in \mathcal{S} \times \mathcal{A}.$ Then $\mu^{\top} \left(\sum_{h=1}^{H}(H-h+1) \nu_{h}\right) = \mathbb{E}^{\pi}\left[(H-h+1) g(s_h,a_h) \mid s_1 \sim \xi(\cdot)\right].$ Additionally, we have
\begin{equation*}
    \mu^{\top} \Sigma \mu = \mathbb{E}_{\boldsymbol{\tau}} \left\{\frac{1}{H} \sum_{j=1}^H \left[\phi^{\top}(s_{j}, a_{j}) \cdot \mu\right]^2 \right\} = \mathbb{E}_{\boldsymbol{\tau}} \left\{\frac{1}{H} \sum_{h=1}^H g^2\left(s_h,a_h\right) \right\}
\end{equation*}
Then,
\begin{align*}
     \sqrt{\left(\sum_{h=1}^{H}(H-h+1) \nu_{h}\right)^{\top} \Sigma^{-1}\left(\sum_{h=1}^{H}(H-h+1) \nu_{h}\right)} 
     &= \sup_{g \in \mathcal{G}} \frac{\mathbb{E}^{\pi}\left[(H-h+1) g\left(s_{h}, a_{h}\right) \mid s_{1} \sim \xi(\cdot)\right]}{\sqrt{\mathbb{E}\left[\frac{1}{H} \sum_{h=1}^{H} g^{2}\left(s_{h}, a_{h}\right)\right]}} \\
     &= \frac{H(H+1)}{2} \sqrt{1+\chi_{\mathcal{G}}^{2}(\tilde{\mu}, \bar{\mu})}.
\end{align*}
We replace $\delta$ with $\delta/3,$ bound the higher order term as section \ref{sec:higher_order}, and substitute this supremum into upper bound. This implies the result in linear function case.

\subsection{Special Case: Tabular Case}
This part is proven in \citep{duan2020minimax}, but for completeness, we briefly sketch the outline. In tabular case, the MDP we consider has finite state space and action space. We denote their cardinality as $\left|\mathcal{S}\right|$ and $\left|\mathcal{A}\right|.$ Intrinsically, we can always parametrize tabular MDP in a linear way, hence tabular case is a special case of linear case. Our feature map is an $\left|\mathcal{S}\right| \times \left|\mathcal{A}\right|$- dimensional indicator vector $\phi(s,a) = \boldsymbol{1}_{s,a},$ where the $(s,a)-$th entry is 1 and other entries are 0. Covariance martrix $\Sigma$ is diagonal with non-negative entries, hence positivity condition holds naturally. By the definition of $\nu_h$ and $\Sigma,$ we have $\nu_h((s,a)) = \mathbb{P}^{\pi} \left[s_h = s,a_h = a\mid s_1 \sim \xi(\cdot)\right]$ and $\Sigma((s,a),(s,a)) = \bar{\mu}(s,a).$ Further we have $\sum_{h=1}^{H}(H-h+1) \nu_{h}((s,a)) = \frac{1}{2} H(H+1) \tilde{\mu}(s,a),$ where $\tilde{\mu}$ is defined as \eqref{definition_tilde_mu}. Therefore, the coefficient in the dominant term will become
\begin{equation*}
    \sqrt{\left(\sum_{h=1}^{H}(H-h+1) \nu_{h}\right)^{\top} \Sigma^{-1}\left(\sum_{h=1}^{H}(H-h+1) \nu_{h}\right)} = \sum_{h=1}^H (H-h+1)\sqrt{\sum_{s\in \mathcal{S},a \in \mathcal{A}} \frac{\tilde{\mu}((s,a))^2}{\bar{\mu}(s,a)}} =  \frac{H (H+1)}{2} \sqrt{1 + \chi^2\left(\tilde{\mu},\bar{\mu}\right)}.
\end{equation*}
This matches the result optimal result in \citep{yin2020asymptotically}.

\subsection{Proof of Lemma \ref{lemma_first_order}}\label{sec:lemma}

We fix a pair of $h_1,h_2$ and we denote
\begin{equation*}
    \Sigma_{h_1,h_2} := \mathbb{E}\left[\frac{1}{H}\sum_{j=1}^H \nabla_{\theta}^{\top}f\left(\theta_{h_1}^*, \phi\left(s_j, a_j\right)\right) \cdot \nabla_{\theta}f\left(\theta_{h_2}^*, \phi\left(s_j, a_j\right)\right)\right],
\end{equation*}
and
\begin{equation*}
    X_k := \frac{1}{H} \sum_{h=1}^H \Sigma_{h_1}^{-\frac{1}{2}} \nabla_{\theta}^{\top}f\left(\theta_{h_1}^*, \phi\left(s_j^k, a_j^k\right)\right) \nabla_{\theta}f\left(\theta_{h_2}^*, \phi\left(s_j^k, a_j^k\right)\right) \Sigma_{h_2}^{-\frac{1}{2}},
\end{equation*}
then $X_1,X_2,..,X_K$ are independent with $\mathbb{E}\left[X_k\right] = \Sigma_{h_1}^{-\frac{1}{2}} \Sigma_{h_1,h_2} \Sigma_{h_2}^{-\frac{1}{2}}.$ Below we use matrix \ref{ine_matrix_bernstein} to bound $\frac{1}{K}\sum_{k=1}^K X_k.$ If we denote
\begin{equation*}
    \Phi_{h,k} := \left(\nabla_{\theta}^{\top}f\left(\theta_{h}^*, \phi\left(s_1^k, a_1^k\right)\right),\nabla_{\theta}^{\top}f\left(\theta_{h}^*, \phi\left(s_2^k, a_2^k\right)\right),...,\nabla_{\theta}^{\top}f\left(\theta_{h}^*, \phi\left(s_H^k, a_H^k\right)\right)\right) \in \mathbb{R}^{d \times H},
\end{equation*}
then $X_k = \frac{1}{H} \Sigma_{h_1}^{-\frac{1}{2}} \Phi_{h_1,k} \Phi_{h_2,k}^{\top} \Sigma_{h_2}^{-\frac{1}{2}}.$ For any vector $\mu \in \mathbb{R}^d,$ we have
\begin{align*}
    \mu^{\top}\mathbb{E}\left[X_k^2\right] \mu 
    &= \mathbb{E}\left[\left\|X_k\mu\right\|_2^2\right] = \frac{1}{H^2} \mathbb{E}\left[\left\|\Sigma_{h_1}^{-\frac{1}{2}} \Phi_{h_1,k} \Phi_{h_2,k}^{\top} \Sigma_{h_2}^{-\frac{1}{2}} \mu\right\|_2^2\right]\\
    &\leq \frac{1}{H^2} \mathbb{E}\left[\left\|\Sigma_{h_1}^{-\frac{1}{2}} \Phi_{h_1,k} \right\|_2^2 \left\|\Phi_{h_2,k}^{\top} \Sigma_{h_2}^{-\frac{1}{2}} \mu \right\|_2^2\right]
\end{align*}
From assumption \ref{assumption_data_coverage}, we have $\left|\left(\Phi_{h_1,k}^{\top} \Sigma_{h_1}^{-1} \Phi_{h_1,k}\right)_{ij}\right| \leq C_2 d$ for any $i,j \in [H],$ we have
\begin{equation*}
    \left\|\Sigma_{h_1}^{-\frac{1}{2}} \Phi_{h_1,k} \right\|_2^2 = \left\|\Phi_{h_1,k}^{\top} \Sigma_{h_1}^{-1} \Phi_{h_1,k}\right\|_2 \leq \left\|\Phi_{h_1,k}^{\top} \Sigma_{h_1}^{-1} \Phi_{h_1,k}\right\|_{\operatorname{F}} \leq C_2 d H.
\end{equation*}
Therefore,
\begin{equation*}
    \mu^{\top}\mathbb{E}\left[X_k^2\right] \mu 
    \leq \frac{C_2 d}{H} \mathbb{E}\left[\left\|\Phi_{h_2,k}^{\top} \Sigma_{h_2}^{-\frac{1}{2}} \mu \right\|_2^2\right] = \frac{C_2 d}{H} \mu^{\top} \mathbb{E} \left[\Sigma_{h_2}^{-\frac{1}{2}} \Phi_{h_2,k} \Phi_{h_2,k}^{\top} \Sigma_{h_2}^{-\frac{1}{2}}\right] \mu = C_2 d \left\|\mu\right\|_2^2.
\end{equation*}
because $\frac{1}{H}\mathbb{E} \left[\Sigma_{h_2}^{-\frac{1}{2}} \Phi_{h_2,k}^{\top} \Phi_{h_2,k} \Sigma_{h_2}^{-\frac{1}{2}}\right] = I_d$ by definition. Therefore,
\begin{equation*}
    \operatorname{Var}\left[X_k\right] \preceq \mathbb{E}\left[X_k^2\right] \preceq C_2d I_d.
\end{equation*}
and if we denote $\sigma_{h_1,h_2} := \left\|\Sigma_{h_1}^{-\frac{1}{2}} \Sigma_{h_1,h_2} \Sigma_{h_2}^{-\frac{1}{2}}\right\|_2,$ then
\begin{align*}
    \left\|X_k - \Sigma_{h_1}^{-\frac{1}{2}} \Sigma_{h_1,h_2} \Sigma_{h_2}^{-\frac{1}{2}}\right\|_2 \leq \left\|X_k\right\|_2 + \left\|\Sigma_{h_1}^{-\frac{1}{2}} \Sigma_{h_1,h_2} \Sigma_{h_2}^{-\frac{1}{2}}\right\|_2 \leq \frac{1}{H} \left\|\Sigma_{h_1}^{-\frac{1}{2}} \Phi_{h_1,k} \right\|_2^2 \left\|\Sigma_{h_2}^{-\frac{1}{2}} \Phi_{h_2,k} \right\|_2^2 + \sigma_{h_1,h_2}\leq C_2 d + \sigma_{h_1,h_2}.
\end{align*}

Therefore, by matrix Bernstein inequality, we have for $\varepsilon > 0,$
\begin{equation*}
    \mathbb{P}\left(\left\|\sum_{k=1}^K\left(X_k - \Sigma_{h_1}^{-\frac{1}{2}} \Sigma_{h_1,h_2} \Sigma_{h_2}^{-\frac{1}{2}} \right)\right\|_2 \geq \varepsilon\right) \leq 2d \exp\left(-\frac{\varepsilon^2/2}{C_2d K + \left(C_2d+\sigma_{h_1,h_2}\right)\varepsilon/3}\right).
\end{equation*}
Therefore, with probability at least $1 - \delta,$
\begin{equation*}
    \left\|\frac{1}{K}\sum_{k=1}^K\left(X_k - \Sigma_{h_1}^{-\frac{1}{2}} \Sigma_{h_1,h_2} \Sigma_{h_2}^{-\frac{1}{2}} \right)\right\|_2 \leq \sqrt{\frac{2C_2 d}{K}\log\left(\frac{2d}{\delta}\right)} + \frac{2\left(C_2d+\sigma_{h_1,h_2}\right)}{3 K}\log\left(\frac{2d}{\delta}\right).
\end{equation*}

\section{Proof of Information-Theoretic Lower Bound}
    We first compute the influence function of $v_{\pi}$ and show that the expectation of squared influence function meets the variance term in \eqref{variance_expression}. We denote 
    $$p_{\eta}(s^{\prime}\mid s,a) = p(s^{\prime} \mid s,a) + \eta \Delta p(s^{\prime},s,a).$$
    where $\Delta p$ is arbitrary probability shift. When $\eta = 0,$ this notation is same as our original transition probability $p_{0}(s^{\prime} \mid s,a) = p(s^{\prime} \mid s,a).$ Suppose $\Delta p$ satisfies $(\Delta p) \mathcal{F} \subset \mathcal{F}.$ We denote $Q_{h,\eta}(s,a)$ and $V_{h,\eta}(s)$ as the Q function and state value function with transition probability being $p_{\eta}(s^{\prime} \mid s,a),$ and $Q_{h,0}(s,a) = Q_h(s,a), V_{h,0}(s) = V_h(s).$ $\theta_h^*$ will be dependent on $\eta,$ hence we write $\theta_{h,\eta}^*$ to explicate this dependency. When $\eta = 0,$ we let $\theta_{h,0}^* = \theta_h^*.$ We emphasize again that $\mathbb{E}^{\pi}$ denotes expectation over population generated by target policy, and $\mathbb{E}$ or $\mathbb{E}_{\boldsymbol{\tau}}$ denotes that by behavior policy. Let $\Bar{\mu}$ be the occupancy measure of state action pair generated by behavior policy. We define the score function as 
    $$
    l_{\eta}(s^{\prime} \mid s,a) := \frac{\partial}{\partial \eta} \log p_{\eta}(s^{\prime} \mid s,a) \ \text{ and } \ 
    l_{\eta}(\boldsymbol{\tau}) := \sum_{h=1}^H l_{\eta}(s_{h+1} \mid s_h,a_h).    
    $$
    When $\eta$ vanishes, we have $l(s^{\prime} \mid s,a) := \left.\frac{\partial}{\partial \eta} \log p_{\eta}(s^{\prime} \mid s,a)\right|_{\eta = 0}$ and $l(\boldsymbol{\tau}) := \sum_{h=1}^H l(s_{h+1} \mid s_h,a_h).$ Our objective function is 
    $$v_{\pi,\eta} := \mathbb{E}^{\pi} \left[\left.\sum_{h=1}^H r(s_h,a_h) \right| s_1 \sim \xi(\cdot), p_{\eta}\right].$$
    We take its derivatives and then let $\eta = 0$ to compute the influence function.
    \begin{align*}
        \left.\frac{\partial}{\partial \eta}v_{\pi,\eta}\right|_{\eta = 0}
        &= \frac{\partial}{\partial \eta}\left.\left[\sum_{h=1}^H \int_{(\mathcal{S} \times \mathcal{A})^h} r(s_h,a_h) \xi(s_1) \prod_{h^{\prime}=1}^{h-1} p_{\eta}(s_{h^{\prime}+1} \mid s_{h^{\prime}},a_{h^{\prime}}) \prod_{h^{\prime} = 1}^h \pi(a_{h^{\prime}} \mid s_{h^{\prime}}) d \boldsymbol{\tau}_h \right] \right|_{\eta = 0}\\
        &= \left.\sum_{h=1}^H \int_{(\mathcal{S} \times \mathcal{A})^h} r(s_h,a_h) \xi(s_1) \sum_{h^{\prime} = 1}^{h-1} l_{\eta}(s_{h^{\prime}+1} \mid s_{h^{\prime}}, a_{h^{\prime}}) \prod_{h^{\prime}=1}^{h-1} p_{\eta}(s_{h^{\prime}+1} \mid s_{h^{\prime}},a_{h^{\prime}}) \prod_{h^{\prime} = 1}^h \pi(a_{h^{\prime}} \mid s_{h^{\prime}}) d \boldsymbol{\tau}_h  \right|_{\eta = 0}\\
        &= \left.\int_{(\mathcal{S} \times \mathcal{A})^H} \left[\sum_{h=1}^{H-1} l_{\eta}(s_{h+1} \mid s_{h}, a_{h}) \sum_{h^{\prime} = h+1}^H r(s_{h^{\prime}},a_{h^{\prime}})\right] \xi(s_1) \prod_{j=1}^{H} p_{\eta}(s_{j+1} \mid s_{j},a_{j}) \pi(a_j \mid s_j) d \boldsymbol{\tau}\right|_{\eta = 0}\\
        &= \mathbb{E}^{\pi} \left.\left[\left.\sum_{h=1}^{H-1} l_{\eta}(s_{h+1} \mid s_{h}, a_{h}) \sum_{h^{\prime} = h+1}^H r(s_{h^{\prime}},a_{h^{\prime}})\right| s_1 \sim \xi, p_{\eta}\right]\right|_{\eta = 0}\\
        & = \mathbb{E}^{\pi} \left.\left[\left.\sum_{h=1}^{H} \mathbb{E}^{\pi}\bigg[\left. l_{\eta}(s_{h+1} \mid s_{h}, a_{h}) V_{h+1,\eta}(s_{h+1})\right| s_h, a_h\bigg]\right| s_1 \sim \xi, p_{\eta}\right]\right|_{\eta = 0}.
    \end{align*}
    Since
    \begin{align*}
        \mathbb{E}^{\pi}\left[l_{\eta}(s^{\prime} \mid s,a) V_{h+1,\eta}(s^{\prime}) \mid s,a\right]
        &= \int_{\mathcal{S} \times \mathcal{A}} \frac{\partial}{\partial \eta} f(\theta_{h+1,\eta}^*,\phi(s,a)) p_{\eta}(s^{\prime}\mid s,a) \pi(a^{\prime} \mid s^{\prime}) d s^{\prime} d a^{\prime} = \frac{\partial}{\partial \eta}f(\theta_{h,\eta}^*,\phi(s,a)),
    \end{align*}
    we have
    \begin{align*}
        \left.\frac{\partial}{\partial \eta}v_{\pi,\eta}\right|_{\eta = 0}
        &= \mathbb{E}^{\pi} \left[\sum_{h=1}^{H} \left.\left.\frac{\partial}{\partial \eta}f(\theta_{h,\eta}^*,\phi(s_h,a_h))\right|_{\eta = 0} \right| s_1 \sim \xi, p \right]\\
        &= \mathbb{E}^{\pi} \left[\sum_{h=1}^{H} \left.\left.\bigg(\nabla_{\theta_h} f(\theta_{h}^*,\phi(s_h,a_h))\bigg) \Sigma_h^{-1} \Sigma_h \frac{\partial \theta_{h,\eta}}{\partial \eta}\right|_{\eta = 0} \right| s_1 \sim \xi, p \right].
    \end{align*}
    Notice that
    \begin{align*}
        \Sigma_h &= \frac{1}{H} \mathbb{E}_{\tau}\left\{\sum_{h^{\prime}=1}^{H}\bigg(\nabla_{\theta_h} f\left(\theta_{h}^*, \phi(s_{h^{\prime}}, a_{h^{\prime}})\right)\bigg)^{\top}\bigg(\nabla_{\theta_h} f\left(\theta_{h}^*, \phi(s_{h^{\prime}}, a_{h^{\prime}})\right)\bigg)\right\} \\
        &= \mathbb{E}_{(s,a) \sim \Bar{\mu}}\left[\bigg(\nabla_{\theta_h} f\left(\theta_{h}^*, \phi(s, a)\right)\bigg)^{\top} \bigg(\nabla_{\theta_h} f\left(\theta_{h}^*, \phi(s, a)\right)\bigg)\right],
    \end{align*}
    then
    \begin{align*}
        &\left.\frac{\partial}{\partial \eta}v_{\pi,\eta}\right|_{\eta = 0}\\
        =& \mathbb{E}^{\pi} \left[\sum_{h=1}^{H} \left.\left.\bigg(\nabla_{\theta_h} f(\theta_{h}^*,\phi(s_h,a_h))\bigg) \Sigma_h^{-1}  \mathbb{E}_{(s,a) \sim \Bar{\mu}}\left[\bigg(\nabla_{\theta_h} f\left(\theta_{h}^*, \phi(s, a)\right)\bigg)^{\top} \bigg(\nabla_{\theta_h} f\left(\theta_{h}^*, \phi(s, a)\right)\bigg) \frac{\partial \theta_{h,\eta}}{\partial \eta}\right|_{\eta = 0} \right]\right| s_1 \sim \xi, p \right]\\
        =& \mathbb{E}^{\pi} \left[\sum_{h=1}^{H} \left.\left.\bigg(\nabla_{\theta_h} f(\theta_{h}^*,\phi(s_h,a_h))\bigg) \Sigma_h^{-1}  \mathbb{E}_{(s,a) \sim \Bar{\mu}}\left[\bigg(\nabla_{\theta_h} f\left(\theta_{h}^*, \phi(s, a)\right)\bigg)^{\top} \frac{\partial}{\partial \eta} f(\theta_{h,\eta}^*,\phi(s,a))\right|_{\eta = 0} \right]\right| s_1 \sim \xi, p \right]\\
        =& \mathbb{E}_{(s,a) \sim \Bar{\mu}} \left\{ \sum_{h=1}^{H} \mathbb{E}^{\pi} \left[\nabla_{\theta_h}f(\theta_{h}^*,\phi(s_h,a_h))\bigg| s_1 \sim \xi, p\right] \Sigma_h^{-1} \bigg(\nabla_{\theta_h} f\left(\theta_{h}^*, \phi(s, a)\right)\bigg)^{\top} l(s^{\prime} \mid s,a) V_{h+1}(s^{\prime}) \right\}.
    \end{align*}
    We define
    \begin{equation*}
        w_h(s,a) := \mathbb{E}^{\pi}\bigg[\nabla_{\theta_h} f\left(\theta_{h}^*, \phi(s_h, a_h)\right) \mid s_{1} \sim \xi, p\bigg] \Sigma_{h}^{-1}\bigg(\nabla_{\theta_h} f\left(\theta_{h}^*, \phi(s, a)\right)\bigg)^{\top} \in \mathbb{R},
    \end{equation*}
    and use $\mathbb{E}\left[l(s^{\prime} | s,a) | s,a\right] = 0$ to get
    \begin{align*}
        \left.\frac{\partial}{\partial \eta}v_{\pi,\eta}\right|_{\eta = 0}
        &= \mathbb{E}_{(s,a) \sim \Bar{\mu}} \left\{ \sum_{h=1}^{H} w_h(s,a)  l(s^{\prime} \mid s,a) V_{h+1}(s^{\prime}) \right\}\\
        &= \mathbb{E}_{(s,a) \sim \Bar{\mu}} \left\{ \sum_{h=1}^{H} w_h(s,a)  l(s^{\prime} \mid s,a) \bigg[V_{h+1}(s^{\prime}) - \mathbb{E} \left[V_{h+1}(s^{\prime}) | s,a\right]\bigg]\right\}\\
        &= \frac{1}{H} \mathbb{E}\left\{\sum_{j=1}^H \sum_{h=1}^H w_h(s_j,a_j) l(s_{j+1}\mid s_j,a_j) \bigg[V_{h+1}(s_{j+1}) - \mathbb{E} \left[V_{h+1}(s_{j+1}) | s_j,a_j\right]\bigg]\right\}\\
        &= \frac{1}{H} \mathbb{E}\left\{l(\boldsymbol{\tau}) \sum_{j=1}^H \sum_{h=1}^H w_h(s_j,a_j) \bigg[V_{h+1}(s_{j+1}) - \mathbb{E} \left[V_{h+1}(s_{j+1}) | s_j,a_j\right]\bigg]\right\}.
    \end{align*}
    This give us the influence function of our objective function. If we denote
    \begin{equation*}
        q(s,a,s^{\prime}) := \sum_{h=1}^H w_h(s,a) \bigg[V_{h+1}(s^{\prime}) - \mathbb{E} \left[V_{h+1}(s^{\prime}) | s,a\right]\bigg],
    \end{equation*}
    then the influence function can be written as
    \begin{equation}\label{influence_function}
        \mathcal{I}_p(\boldsymbol{\tau}) := \frac{1}{H} \sum_{j=1}^H \sum_{h=1}^H w_h(s_j,a_j) \bigg[V_{h+1}(s_{j+1}) - \mathbb{E} \left[V_{h+1}(s_{j+1}) | s_j,a_j\right]\bigg] = \frac{1}{H} \sum_{h=1}^H q(s_h,a_h,s_{h+1}). 
    \end{equation}
    Next, we square the influence function and take expectation with it. Then we use $\mathbb{E}\left[q(s,a,s^{\prime}) \mid s,a\right] = 0$ to make cross terms vanish.
    \begin{align*}
        \mathbb{E}\left\{\mathcal{I}_p(\boldsymbol{\tau})^2\right\}
        &= \frac{1}{H^2} \mathbb{E}\left\{\ \sum_{h=1}^H q(s_h,a_h,s_{h+1}) \right\}^2\\
        &= \frac{1}{H^2} \sum_{h=1}^H  \mathbb{E}\left\{q(s_h,a_h,s_{h+1})^2 \right\}\\
        &= \frac{1}{H^2} \sum_{h=1}^H \sum_{h_1 = 1}^H \sum_{h_2=1}^H \mathbb{E} \bigg\{w_{h_1}(s_h,a_h) w_{h_2}(s_h,a_h) \varepsilon_{h_1,h} \varepsilon_{h_2,h}\bigg\}.
    \end{align*}
    where for $j,h \in [H],$
    \begin{equation*}
        \varepsilon_{j,h} := f\left(\theta_{j}^*, \phi_{h}\right)-r_{h}-\int_{\mathcal{A}} f\left(\theta_{j+1}^*, \phi_{h+1}\right) \pi\left(a_{h+1} \mid s_{h+1}\right) d a_{h+1} \in \mathbb{R}.
    \end{equation*}
    Then we have
    \begin{align*}
        &\mathbb{E}\left\{\mathcal{I}_p(\boldsymbol{\tau})^2\right\}\\
        =& \frac{1}{H^2} \sum_{h_1 = 1}^H \sum_{h_2 = 1}^H \mathbb{E}^{\pi}\bigg[\nabla_{\theta_{h_1}} f\left(\theta_{h_1}^*, \phi(s_{h_1}, a_{h_1})\right) \mid s_{1} \sim \xi, p\bigg] \Sigma_{h_1}^{-1} \cdot \mathbb{E}\left[\sum_{h=1}^H \bigg(\nabla_{\theta_{h_1}} f(\theta_{h_1},\phi(s_h,a_h))\bigg)^{\top} \right.\\
        & \left.\bigg(\nabla_{\theta_{h_2}} f(\theta_{h_2}^*,\phi(s_h,a_h))\bigg) \varepsilon_{h_1,h} \varepsilon_{h_2,h} \right] \Sigma_{h_2}^{-1} \mathbb{E}^{\pi}\bigg[\nabla_{\theta_{h_2}} f\left(\theta_{h_2}^*, \phi(s_{h_2}, a_{h_2})\right) \mid s_{1} \sim \xi, p\bigg]^{\top}\\
        =& \frac{1}{H}\sum_{h_1 = 1}^H \sum_{h_2 = 1}^H \nu_{h_1}^{\top} \Sigma_{h_1}^{-1} \Omega_{h_1,h_2} \Sigma_{h_2}^{-1} \nu_{h_2}
    \end{align*}
    where $\nu_h, \Sigma_h, \Omega_{h_1,h_2}$ are defined in Theorem \ref{thm_normality}. In conclusion, the expression above implies that
    \begin{equation*}
        \mathbb{E}\left\{\mathcal{I}_p(\boldsymbol{\tau})^2\right\} = \sigma^2.
    \end{equation*}
    where $\sigma^2$ is defined as \eqref{variance_expression}, and this proves the Cramer Rao lower bound for variance.

\section{Technical Lemmas for Contraction}
\begin{lemma}[Bernstein's Inequality]\label{ine_bernstein}
    Let $X_{1}, \ldots, X_{N}$ be independent mean-zero random variables such that $\left|X_{i}\right| \leq K$ all $i$. Then, for every $t \geq 0$, we have
    $$
    \mathbb{P}\left\{\left|\sum_{i=1}^{N} X_{i}\right| \geq t\right\} \leq 2 \exp \left(-\frac{t^{2} / 2}{\sigma^{2}+K t / 3}\right)
    $$
    Here $\sigma^{2}=\sum_{i=1}^{N} \mathbb{E} X_{i}^{2}$ is the variance of the sum.
\end{lemma}

\begin{lemma}[Freedman's Inequality]\label{ine_freedman}
    Consider a real-valued martingale $\left\{Y_{k}: k=0,1,2, \ldots\right\}$ with difference sequence $\left\{X_{k}: k=1,2,3, \ldots\right\}$. Assume that the difference sequence is uniformly bounded:
    $$
    X_{k} \leq R \quad \text { almost surely } \quad \text { for } k=1,2,3, \ldots
    $$
    Define the predictable quadratic variation process of the martingale:
    $$
    W_{k}:=\sum_{j=1}^{k} \mathbb{E}_{j-1}\left(X_{j}^{2}\right) \quad \text { for } k=1,2,3, \ldots
    $$
    Then, for all $t \geq 0$ and $\sigma^{2}>0$,
    $$
    \mathbb{P}\left\{\exists k \geq 0: Y_{k} \geq t \text { and } W_{k} \leq \sigma^{2}\right\} \leq \exp \left\{-\frac{t^{2} / 2}{\sigma^{2}+R t / 3}\right\}
    $$
\end{lemma}

\begin{lemma}[Matrix Bernstein inequality]\label{ine_matrix_bernstein}
 Let $X_{1}, \ldots, X_{N}$ be independent mean-zero $n \times n$ symmetric random matrices, such that $\left\|X_{i}\right\| \leq K$ almost surely for all $i$. Then, for every $t \geq 0$, we have
$$
\mathbb{P}\left\{\left\|\sum_{i=1}^{N} X_{i}\right\| \geq t\right\} \leq 2 n \exp \left(-\frac{t^{2} / 2}{\sigma^{2}+K t / 3}\right)
$$
Here $\sigma^{2}=\left\|\sum_{i=1}^{N} \mathbb{E} X_{i}^{2}\right\|$ is the norm of the matrix variance of the sum.
\end{lemma}

\begin{lemma}[Theorem 2.14.9 of \citep{empirical_process}]\label{maxima_of_empirical_process}
   Let $\mathcal{H}$ be a class of measurable functions $g: \mathcal{X} \mapsto[0,1]$ that satisfies 
   $$
   N_{[]}\left(\varepsilon, \mathcal{H}, L_{2}(\mathbb{P})\right) \leq\left(\frac{U}{\varepsilon}\right)^{V}, \quad \text { for every } 0<\varepsilon<U.
   $$
   We denote the empirical process: $$\mathbb{G}_n(\cdot)=\sqrt{n} (\mathbb{P}_n (\cdot) - \mathbb{P} (\cdot) ) $$
   For any function class $\mathcal{H},$ we define the supremum norm
   $$
   \left\Vert \mathbb{G}_n \right\Vert_{\mathcal{H}} := \sup_{g \in \mathcal{H}} \left\{\sqrt{n} (\mathbb{P}_n g - \mathbb{P} g)\right\}
   $$
   Then, for every $t>0,$
    $$
    \mathrm{P}\left(\left\|\mathbb{G}_{n}\right\|_{\mathcal{H}}>t\right) \leq\left(\frac{D t}{\sqrt{V}}\right)^{V} e^{-2 t^{2}}
    $$
    for a constant $D$ that depends on $U$ only.
\end{lemma}

\end{document}